\documentclass[11pt]{article}
\usepackage[utf8]{inputenc}
\usepackage[margin=1in]{geometry}
\usepackage{amsmath,amssymb}
\usepackage{graphicx}
\usepackage{booktabs,array}
\usepackage[font=small,labelfont=bf,skip=6pt]{caption}
\usepackage[numbers,sort&compress]{natbib}
\usepackage{authblk}
\usepackage{microtype}
\usepackage{xcolor}
\usepackage[hidelinks]{hyperref}
\usepackage{url}
\usepackage{xurl}
\usepackage{placeins}
\title{Benchmarking CNNs, Vision Transformers, and Foundation Models for Explainable Brain Tumor MRI Classification: Calibration, Decision Analysis, and External Validation}
\author[1,*]{Akwasi Asare}
\affil[1]{Department of Computer Science, Faculty of Computing and Information Systems, Ghana Communication Technology University, Accra, Ghana}
\affil[2]{Department of Research and Engineering, Centre for Medical and Agro Computing (C-MAC), Accra, Ghana}
\affil[*]{Corresponding author: \href{mailto:nasare34@yahoo.com}{nasare34@yahoo.com}}
\date{}

\begin{document}
\maketitle

\begin{abstract}
Deep learning models for brain tumor MRI classification are usually compared by accuracy alone, on public datasets whose independence is rarely checked. We benchmarked six architectures, two convolutional networks (EfficientNetV2-M, ConvNeXt-V2-Tiny), two vision transformers (Swin-V2-S, ViT-L/16) and two foundation models (DINOv2-L, MedSigLIP-448), for four-class classification of 7,216 public MRI images, assessing accuracy, calibration, explainability, ensembles and decision curves, with every fitted quantity estimated on validation data only. Pixel-verified duplicate screening showed that 36\% of the test images had a near-duplicate in the training data. Test accuracy ranged from 94.5\% to 95.6\%, with no pairwise difference after Holm correction, and fell to 90.3--91.7\% on the 773 test images free of duplicates and shared patients. Ensembles did not outperform the best single model. Attribution maps passed a model-randomisation test but localised expert-annotated tumors poorly. On 1,675 external BDNeuro-MRI images that did not duplicate the training data, accuracy ranged from 77.6\% (MedSigLIP-448) to 95.0\% (ViT-L/16), and 12 of 15 pairwise differences were significant. Internal accuracy on this public collection therefore did not distinguish the architectures, whereas the duplicate-filtered external test did.
\end{abstract}


\section{Introduction}

Brain tumors are among the most dangerous and life-threatening conditions affecting the central nervous system. They are characterised by the abnormal, uncontrolled proliferation of cells within or surrounding the brain, and their clinical consequences depend on the tumor's histological type, grade, size, and anatomical location~\cite{r1,r2}. Primary brain tumors arise directly from brain tissue or the meninges, while secondary or metastatic tumors originate elsewhere in the body and spread to the brain~\cite{r2,r3,r4}. The most common primary malignant variant, glioblastoma multiforme, carries a median survival of approximately 15 months even with aggressive multimodal therapy including surgery, radiotherapy, and temozolomide chemotherapy~\cite{r5}. Meningiomas, although predominantly benign, carry significant morbidity when they compress eloquent cortex or cranial nerves and can recur following incomplete resection~\cite{r6}. Pituitary adenomas disrupt hormonal homeostasis and can impair vision through compression of the optic chiasm~\cite{r7}. Globally, primary central nervous system tumors account for over 300,000 new diagnoses annually, with the true burden substantially higher when secondary metastatic deposits are included~\cite{r8,r9}.

The clinical pathway for brain tumor diagnosis is centred on magnetic resonance imaging, which offers superior soft-tissue contrast, multiplanar capability, and freedom from ionising radiation relative to computed tomography~\cite{r10}. T1-weighted contrast-enhanced sequences depict blood-brain barrier disruption and active tumor enhancement; T2-weighted and FLAIR sequences reveal perilesional oedema, infiltration margins, and non-enhancing tumor components; and diffusion-weighted imaging provides information on tissue cellularity~\cite{r11}. In practice, however, the manual interpretation of these multiparametric imaging datasets is time-consuming, subject to inter-reader variability, and dependent on specialist neuroradiological expertise that is unevenly distributed across healthcare systems worldwide~\cite{r12}. Diagnostic delays and inconsistencies carry direct clinical consequences: late identification of high-grade glioma, for example, reduces the window for effective intervention and worsens prognosis~\cite{r13,r14}. These pressures motivate research into automated, computer-aided classification systems that could provide consistent and reproducible first-pass categorisation of brain MRI findings, provided that their performance is established under rigorous evaluation.

Convolutional neural networks (CNNs) were among the first deep learning architectures to demonstrate clinically relevant performance on medical image classification tasks~\cite{r15,r16,r17,r18,r19}. Networks such as ResNet, DenseNet, and EfficientNet progressively expanded the accuracy frontier through innovations in residual connectivity, dense feature reuse, and compound scaling of network depth, width, and resolution~\cite{r15,r19,r20}. However, the inductive biases of convolution, particularly translational equivariance and local receptive fields, limit the ability of CNNs to model long-range spatial dependencies across an entire image. For brain tumor classification, where the relationship between a lesion and surrounding anatomical structures such as ventricles, sulci, and the midline can carry diagnostic significance, this is a meaningful constraint~\cite{r21}.

Vision transformers (ViTs), introduced by Dosovitskiy et al.~~\cite{r22}, addressed this limitation by treating image patches as sequential tokens processed through multi-head self-attention, enabling each token to attend to every other token in the image. This global receptive field allows transformers to capture contextual relationships that convolutions cannot model as naturally. Hierarchical variants such as Swin Transformer~\cite{r23} and Swin-V2~\cite{r24} restored the multi-scale feature pyramids valuable for transfer learning while preserving global attention through shifted-window mechanisms. ConvNeXt~\cite{r25} and ConvNeXt-V2~\cite{r26} further demonstrated that a rigorously modernized pure CNN could match transformer-level performance, establishing that the architectural principle and the training recipe are both important determinants of outcome.

The most recent development in this landscape is the emergence of foundation models pretrained on massive, diverse datasets and then adapted to specific tasks. DINOv2~\cite{r27}, trained through self-supervised self-distillation on 142 million images without any annotation, produces feature representations that transfer broadly across visual tasks. Medical domain-specific foundation models extend this paradigm: MedSigLIP~\cite{r28,r29}, the vision encoder released with the MedGemma models and pretrained on medical image--text pairs, is intended to encode biomedically grounded visual representations. Whether such pretraining translates into better brain tumor classification, and whether any advantage survives a change of data source, are empirical questions.

Published benchmarks on public brain tumor MRI collections routinely report accuracies of 95--99\%~\cite{r30,r31,r32,r33,r34,r35,r36,r37}, yet several features of the evaluation pipeline can inflate such figures. The most widely used four-class collection~\cite{r38} is a composite of three earlier datasets, Figshare~\cite{r39}, SARTAJ~\cite{r40} and Br35H~\cite{r41}, and is distributed with a predefined split whose independence has seldom been examined. Near-duplicate images and slices from the same patient can occur on both sides of a split; test sets are often reused to select models, weight ensembles or tune thresholds; and differences between architectures are rarely tested for statistical significance or corrected for multiple comparisons. Such leakage is a recognised source of irreproducible machine-learning results in science and medicine~\cite{r42}. Models can also exploit acquisition or source-specific cues rather than pathology~\cite{r43,r44}, which internal test sets drawn from the same sources cannot reveal. External validation addresses this only if the external data are themselves independent of the training data. Finally, explainability is usually presented as qualitative heatmaps, without testing whether attributions fall on the lesion.

This study benchmarks six architectures from three families, convolutional networks, vision transformers and foundation models, for explainable four-class brain tumor MRI classification, under a protocol designed to limit the sources of bias described above. EfficientNetV2-M, ConvNeXt-V2-Tiny, Swin-V2-S, ViT-L/16, DINOv2-L and MedSigLIP-448 are trained under an identical protocol and compared with bootstrap confidence intervals, exact McNemar tests with Holm correction, and resolution- and preprocessing-controlled ablations. Before training, every image is screened for exact and pixel-verified near-duplicates and attributed to its source collection, with patient identifiers recovered for the Figshare portion, and the validation split is grouped by patient and by duplicate. Temperature scaling, ensemble weights, stacking and decision thresholds are fitted on validation data only, and the predefined test set is evaluated once. Explanations from Grad-CAM, Grad-CAM++, EigenCAM, Integrated Gradients and LIME are evaluated against expert tumor masks, with a model-randomisation check and stability analyses, and the ensemble is explained directly. The study also reports calibration, ensemble diversity and shortcut tests, an external evaluation on a separately published dataset after removal of the images that duplicate the training data, and an exploratory decision analysis that describes the models' operating characteristics without claiming clinical utility.

\section{Related Work}

The application of deep learning to brain tumor MRI classification has evolved rapidly over the past decade, progressing from shallow CNNs to large-scale transformer and foundation model approaches. This section reviews the major threads of this literature and identifies the gaps that motivate the present study.

\subsection{From Handcrafted Features to Convolutional Neural Networks}

Early computer-aided diagnosis systems for brain tumor classification relied on handcrafted feature extraction followed by classical machine learning classifiers. Techniques based on texture descriptors, wavelet coefficients, and morphological features were fed to support vector machines, k-nearest-neighbour classifiers, and artificial neural networks, achieving moderate accuracy on small binary or ternary classification problems~\cite{r30,r45}. These pipelines required domain expertise to design features and were brittle in the face of imaging variability. The introduction of CNNs transformed this landscape by learning discriminative features directly from image data without manual engineering. Pereira et al.~~\cite{r46} were among the first to demonstrate that CNNs with small 3x3 kernels could segment glioma in MRI with clinically useful precision. Subsequent adoption of deeper architectures, especially ResNet~\cite{r47} and DenseNet~\cite{r48}, provided training stability through residual connections and feature reuse respectively, and several groups reported accuracy improvements when applying these architectures to multi-class brain tumor datasets.

Younis et al.~~\cite{r31} combined a CNN and VGG-16 in an ensemble for binary brain tumor detection on 253 MRI images, reporting accuracies of 96\% (CNN), 98.5\% (VGG-16) and 98.14\% (ensemble); the frequently quoted figure of 98.41\% is the ensemble's precision, not its accuracy. The small dataset required extensive augmentation, and the binary task differs from four-class classification.

In a recent study addressing the need for accurate computer-aided detection, Lall et al.~~\cite{r32} compared the efficacy of a custom CNN, VGG-19, and ResNet-50 for classifying brain tumors into glioma, meningioma, and pituitary types using MRI scans. Their findings indicated that ResNet-50 was the most effective architecture, achieving an accuracy of 97.91\% compared to VGG-19 (95.83\%) and the custom CNN (88.6\%), largely due to its residual connections mitigating vanishing gradient issues. Ali et al.~~\cite{r33} investigated the efficacy of Convolutional Neural Networks for multi-class brain tumor classification, comparing a classic CNN architecture with ResNet50. Their results demonstrated that ResNet50 significantly outperformed the classic model, achieving an accuracy of 99.88\% compared to 94.55\%, highlighting the capability of deeper architectures in capturing complex features. Atmakuri et al.~~\cite{r34} evaluated the effectiveness of deep learning for predicting brain tumor phases by comparing a ResNet-50 model against standard CNN and ANN architectures. Using a dataset from Kaggle, their ResNet-50 approach achieved a maximum accuracy of 95\%, significantly outperforming the CNN (89\%) and ANN (65\%) testing baselines. The authors attributed this superior performance to ResNet's well-structured architecture and its capacity to utilize deeper training layers efficiently.

Abdusalomov et al.~~\cite{r30} took an object-detection perspective, fine-tuning YOLOv7 augmented with CBAM attention, SPPF+ multi-scale pooling, and BiFPN feature fusion on a large, augmented dataset, achieving 99.5\% detection accuracy. The detection framing differs from classification: bounding box localisation and class prediction are interleaved, the dataset was augmented to 51,448 images, and the task was three-class rather than four-class. Direct comparison with classification-oriented studies is therefore problematic. More generally, studies differ in dataset version, class definitions, split construction and the use of a test partition separate from the validation data, which limits the comparability of reported accuracies.

\subsection{Vision Transformers for Medical Image Classification}

Transformers process image patches as sequence tokens through self-attention, enabling each token to attend to every other token in the sequence regardless of spatial distance. Vision Transformers (ViTs)~\cite{r22} have emerged as a powerful alternative to conventional convolutional neural networks for medical image analysis by leveraging self-attention mechanisms to capture long-range spatial relationships within images. Building upon the original ViT architecture, variants such as Swin Transformer and Swin-V2~\cite{r24} introduced hierarchical feature representations and computationally efficient attention mechanisms, making them well-suited for high-resolution medical imaging tasks. At the same time, modernized convolutional architectures including ConvNeXt~\cite{r25} and ConvNeXt-V2~\cite{r26} have demonstrated that carefully redesigned CNNs can achieve performance comparable to transformer-based models. These advances have driven extensive research into transformer and hybrid architectures for brain MRI analysis, particularly for brain tumor classification and diagnosis.

For instance, recent literature highlights both the potential and the practical challenges of applying Vision Transformers to brain tumor classification. Several studies have reported exceptional predictive performance using hierarchical architectures; for instance, Shahzad et al.~~\cite{r35} achieved 98\% accuracy with a Tumor-Swin-Transformer on a multi-source dataset, while Asiri et al.~~\cite{r36} recently utilized a Swin Transformer to classify a four-class MRI dataset of 2,870 images, achieving 97\% accuracy and outperforming standard CNN and ViT baselines. Adding to these architectural optimizations, Alam et al.~~\cite{r37} enhanced the Swin Transformer V2 with a Dual-Branch Down-sampling module and an Inverted Residual Feed-Forward Network, achieving 98.97\% accuracy on 7,023 grayscale MRI scans while improving computational efficiency. Conversely, Raviprakash et al.~~\cite{r49} illustrated the complexities of raw multimodal clinical data, reporting a modest 63.10\% classification accuracy and a 0.76 segmentation Dice score using Swin Transformer V2, underscoring the necessity of rigorous tuning to manage clinical class imbalances.

Many of these evaluations examine a single architecture family, present explainability qualitatively (for example Grad-CAM only, or none)~\cite{r50}, and report point estimates without significance testing or calibration. The contrast between accuracies of 97--99\% on curated public collections~\cite{r35,r36,r37} and 63\% on raw multimodal clinical data~\cite{r49} also illustrates how strongly reported performance depends on the data and evaluation protocol~\cite{r51,r52,r53}. The present study therefore compares architecture families under one protocol and asks how far the comparison depends on the independence of the test data.

\subsection{Foundation Models and Self-Supervised Learning in Medical Imaging}

Recent advances in foundation models shift computer vision from task specific supervised training toward adapting large scale pretrained models. Self-supervised architectures like DINOv2~\cite{r27} learn rich representations without manual annotations, while vision language models capture semantic features through image and text alignment. MedSigLIP~\cite{r28} extends this to the medical domain using paired radiology images and clinical reports, offering strong generalization for disease classification and diagnostic support. Recent applications highlight the transfer learning capabilities of self-supervised vision transformers. Missaoui et al.~~\cite{r54} evaluated the DINO models as frozen feature extractors for medical image classification. Bypassing end to end fine tuning reduced computational overhead while achieving 98.17\% accuracy on a diverse dataset by combining transformer embeddings with a Support Vector Machine. Extending beyond static classification, Hussien et al.~~\cite{r55} used DINOv2 embeddings for semantic search in medical databases to retrieve clinically related cases. They also resolved the opacity of self-supervised representations by integrating ViT CX, a causal explanation technique that generated clinically actionable heatmaps and accurately localized targeted pathologies with 99\% accuracy.

Foundational models pretrained on natural images are inherently optimized for single-modality inputs, limiting their application to multiparametric clinical scans. To overcome this, Scholz et al.~~\cite{r56} proposed MM DINOv2, adapting the architecture through multimodal patch embeddings. By implementing full modality masking during semi-supervised training, the network learned robust cross-modality relationships, rendering it resilient to missing imaging sequences during complex diagnostic classification. Building on this trajectory, the present study examines whether a medical vision foundation model, MedSigLIP-448, pretrained on medical images paired with text, offers advantages over general-purpose foundation models such as DINOv2 and over CNNs and transformers trained on natural images. It examines this both on the internal test set and on duplicate-filtered external data.

\subsection{Explainability, Calibration, and Clinical Utility}

The deployment of deep learning in clinical settings depends not only on accuracy but on three additional properties that the existing brain tumor classification literature largely neglects. The first is explainability: the ability to identify which image regions drove a model's prediction. Regulatory guidance for AI in medical devices increasingly requires that model outputs be interpretable to clinicians, and unexplained black-box predictions are a known barrier to radiologist acceptance~\cite{r57}. Grad-CAM~\cite{r50} and Grad-CAM++~\cite{r58} have been the most widely used gradient-based saliency methods in medical imaging, generating class-discriminative visualisations from convolutional feature maps. For instance, Srinivas and Parvathi~\cite{r59} utilized Grad-CAM and LIME to validate CNN decision-making in brain tumor classification; the best-performing model, a 4-conv-1-dense-1-dropout CNN, achieved a classification accuracy of 95.86\%, outperforming deeper architectures and transfer learning approaches. While Grad-CAM provided global interpretability through class-specific heatmaps, LIME offered local, superpixel-level explanations via image perturbation. This dual-XAI approach successfully exposed irrelevant background activations in complex transfer learning architectures, allowing the authors to iteratively optimize a highly transparent and lightweight CNN. LIME~\cite{r60} provides a model-agnostic alternative through local perturbation-based surrogate modelling. Rasool et al.~~\cite{r61} achieved 99\% accuracy in binary tumor classification using CNN-TumorNet, coupling LIME with superpixel segmentation to foster clinical trust by visually isolating predictive morphological features. Integrated Gradients~\cite{r62} offers axiomatic attribution that satisfies completeness and sensitivity properties. EigenCAM~\cite{r63} adapts to transformer architectures by extracting the principal components of reshaped feature tensors.

A visually plausible heatmap is not evidence that a model relies on the lesion. Attribution maps can be insensitive to the model's parameters~\cite{r64} and should be validated quantitatively, for example by their overlap with expert lesion masks~\cite{r65}. The second property is calibration. A well-calibrated classifier assigns predicted probabilities that correspond to empirical frequencies, whereas deep networks trained with softmax cross-entropy are often miscalibrated~\cite{r66}. Expected calibration error~\cite{r66} and the Brier score are standard measures but are rarely reported in brain tumor deep learning studies. The third property is clinical utility. Decision curve analysis~\cite{r67} quantifies the net benefit of acting on a model's predictions, relative to treating all or no patients, across a range of threshold probabilities. It does so under explicit assumptions about the decision and its consequences, but a retrospective decision curve does not show that model-guided decisions improve patient outcomes; that requires a prospective clinical evaluation.

\subsection{Data Leakage, Duplicates, and Shortcut Learning in Public MRI Benchmarks}

Leakage between training and evaluation data is among the most common causes of overoptimistic results in machine-learning-based science~\cite{r42}. In image classification, leakage takes three main forms: identical or near-identical images on both sides of a split; different images of the same patient (for MRI, adjacent slices of one examination) split across partitions; and the use of test data for model selection, ensemble fitting or threshold tuning. A related problem is shortcut learning~\cite{r43}. Models can exploit features that separate the classes in a particular dataset but are unrelated to the disease, such as scanner- or hospital-specific appearance~\cite{r44}. Composite public collections are particularly exposed because each class may be dominated by a different source. In the collection used here, almost all no-tumor images come from one source. Perceptual hashing~\cite{r68} is a common screen for near-duplicates. But brain MRI slices share anatomy and dark backgrounds, so hash similarity alone also matches different images, and candidate pairs need pixel-level verification. These issues bear directly on how benchmark results on public brain tumor collections should be interpreted.

\subsection{Positioning of This Study}

Table~\ref{tab:1} summarises representative studies. It records the task, data and reported result of each study as described by its authors. It deliberately does not score studies on the presence or absence of particular analyses, because the cited works used different datasets, class definitions and validation protocols, and their accuracies are not directly comparable with each other or with the present results. The present study does not propose a new architecture or claim state-of-the-art accuracy. Its contribution is methodological: a reproducible, leakage-aware comparison of architecture families, and an assessment of how much benchmark conclusions depend on the independence of the evaluation data.

\begin{table}[!htbp]
\centering
\small
\setlength{\tabcolsep}{3.0pt}
\renewcommand{\arraystretch}{1.12}
\caption{Representative brain tumor MRI classification studies as described by their authors. The studies used different datasets, class definitions and validation protocols, so the reported results are not directly comparable.}
\label{tab:1}
\begin{tabular}{>{\raggedright\arraybackslash}p{0.134\linewidth}>{\raggedright\arraybackslash}p{0.268\linewidth}>{\raggedright\arraybackslash}p{0.168\linewidth}>{\raggedright\arraybackslash}p{0.210\linewidth}>{\raggedright\arraybackslash}p{0.153\linewidth}}
\toprule
\nobreak\hspace{0pt}\textbf{Study} & \nobreak\hspace{0pt}\textbf{Architecture(s)} & \nobreak\hspace{0pt}\textbf{Task} & \nobreak\hspace{0pt}\textbf{Data as described by the authors} & \nobreak\hspace{0pt}\textbf{Reported result} \\
\midrule
\nobreak\hspace{0pt}Lall et al.~~\cite{r32} & \nobreak\hspace{0pt}ResNet-50, VGG-19, custom CNN & \nobreak\hspace{0pt}3-class classification & \nobreak\hspace{0pt}Brain MRI (glioma, meningioma, pituitary) & \nobreak\hspace{0pt}Accuracy 97.91\% (ResNet-50) \\
\nobreak\hspace{0pt}Younis et al.~~\cite{r31} & \nobreak\hspace{0pt}CNN, VGG-16, ensemble & \nobreak\hspace{0pt}Binary detection & \nobreak\hspace{0pt}253 images (155 with tumor) & \nobreak\hspace{0pt}Accuracy 98.14\% (ensemble); precision 98.41\% \\
\nobreak\hspace{0pt}Ali et al.~~\cite{r33} & \nobreak\hspace{0pt}ResNet50, classic CNN & \nobreak\hspace{0pt}Multi-class classification & \nobreak\hspace{0pt}Brain MRI & \nobreak\hspace{0pt}Accuracy 99.88\% (ResNet50) \\
\nobreak\hspace{0pt}Abdusalomov et al.~~\cite{r30} & \nobreak\hspace{0pt}YOLOv7 with CBAM, SPPF+, BiFPN & \nobreak\hspace{0pt}3-class detection & \nobreak\hspace{0pt}Augmented to 51,448 images & \nobreak\hspace{0pt}Detection accuracy 99.5\% \\
\nobreak\hspace{0pt}Atmakuri et al.~~\cite{r34} & \nobreak\hspace{0pt}ResNet-50, CNN, ANN & \nobreak\hspace{0pt}Tumor phase prediction & \nobreak\hspace{0pt}Kaggle dataset & \nobreak\hspace{0pt}Accuracy about 95\% \\
\nobreak\hspace{0pt}Shahzad et al.~~\cite{r35} & \nobreak\hspace{0pt}Tumor-Swin-Transformer & \nobreak\hspace{0pt}4-class classification & \nobreak\hspace{0pt}Multi-source dataset & \nobreak\hspace{0pt}Accuracy 98\% \\
\nobreak\hspace{0pt}Asiri et al.~~\cite{r36} & \nobreak\hspace{0pt}Swin Transformer & \nobreak\hspace{0pt}4-class classification & \nobreak\hspace{0pt}2,870 images & \nobreak\hspace{0pt}Accuracy 97\% \\
\nobreak\hspace{0pt}Alam et al.~~\cite{r37} & \nobreak\hspace{0pt}Improved Swin Transformer V2 & \nobreak\hspace{0pt}4-class classification & \nobreak\hspace{0pt}7,023 images (Kaggle composite) & \nobreak\hspace{0pt}Accuracy 98.97\% \\
\nobreak\hspace{0pt}Raviprakash et al.~~\cite{r49} & \nobreak\hspace{0pt}Swin Transformer V2 & \nobreak\hspace{0pt}Classification and segmentation & \nobreak\hspace{0pt}Raw multimodal clinical MRI & \nobreak\hspace{0pt}Accuracy 63.10\%; Dice 0.76 \\
\nobreak\hspace{0pt}Srinivas \& Parvathi~\cite{r59} & \nobreak\hspace{0pt}Custom CNN (Grad-CAM, LIME) & \nobreak\hspace{0pt}4-class classification & \nobreak\hspace{0pt}Brain MRI & \nobreak\hspace{0pt}Accuracy 95.86\% \\
\nobreak\hspace{0pt}Rasool et al.~~\cite{r61} & \nobreak\hspace{0pt}CNN-TumorNet (LIME) & \nobreak\hspace{0pt}Binary classification & \nobreak\hspace{0pt}Brain MRI & \nobreak\hspace{0pt}Accuracy 99\% \\
\nobreak\hspace{0pt}This study & \nobreak\hspace{0pt}Six architectures (CNNs, ViTs, foundation models) & \nobreak\hspace{0pt}4-class classification & \nobreak\hspace{0pt}Kaggle composite, current release (7,216 images); duplicate-filtered BDNeuro-MRI (1,675 images) & \nobreak\hspace{0pt}Internal accuracy 94.5--95.6\% (clean subset 90.3--91.7\%); external 77.6--95.0\% \\
\bottomrule
\end{tabular}
\end{table}

\section{Materials and Methods}

\subsection{Dataset}

\begin{figure}[!htbp]
\centering
\includegraphics[width=0.85\linewidth,height=0.82\textheight,keepaspectratio]{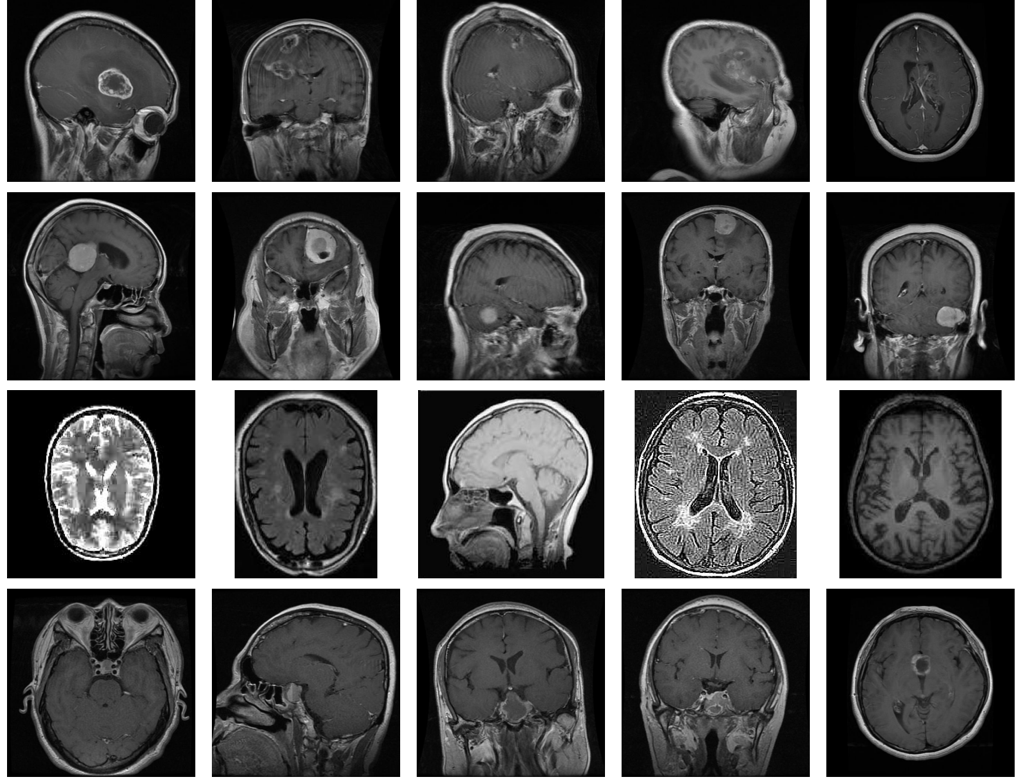}
\caption{Representative images from each class (rows from top: glioma, meningioma, no tumor, pituitary tumor). The collection mixes imaging planes and MRI contrasts.}
\label{fig:1}
\end{figure}

\begin{figure}[!htbp]
\centering
\includegraphics[width=0.88\linewidth,height=0.82\textheight,keepaspectratio]{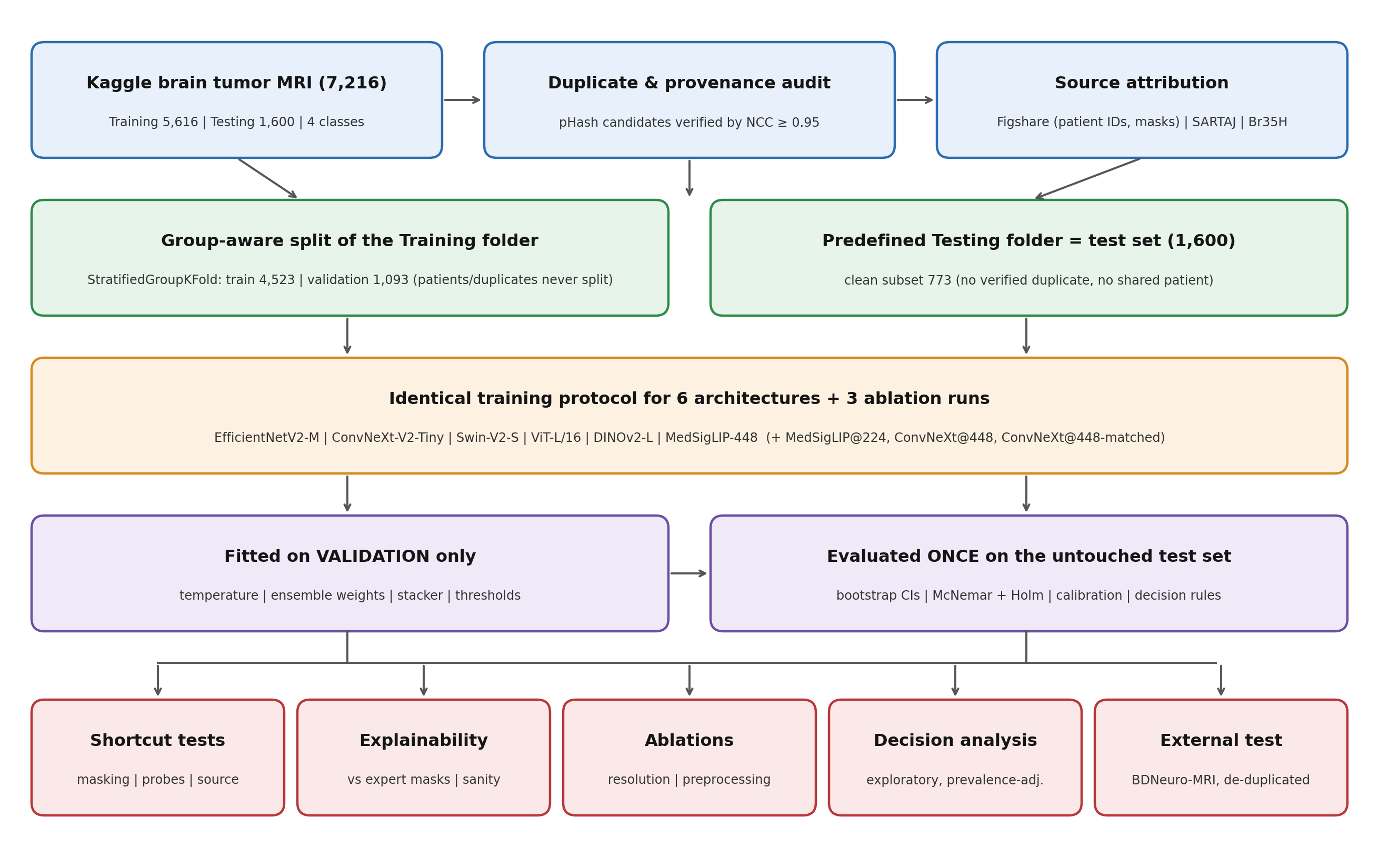}
\caption{Overview of the leakage-aware pipeline. Every fitted quantity (temperature, ensemble weights, stacker and thresholds) is estimated on the validation set, and the predefined test set is evaluated once; shortcut tests, quantitative explainability, ablations, an exploratory decision analysis and duplicate-filtered external validation follow.}
\label{fig:2}
\end{figure}

We used the current release of the public Brain Tumor MRI Dataset distributed on Kaggle~\cite{r38} (accessed in 2026), copied to Google Drive and read directly by the analysis notebook. The release contains 7,216 two-dimensional MRI images in four classes: glioma, meningioma, pituitary tumor and no tumor. It is distributed in two folders defined by the dataset curator. The Training folder holds 5,616 images (1,400 per tumor class and 1,416 no-tumor images), and the Testing folder holds 1,600 images (400 per class) (Table~\ref{tab:2}). The images are not homogeneous. 98.3\% of them store a single grayscale image replicated across three RGB channels, and the remainder are stored as single-channel, RGBA or palette images. Resolution varies, with 512 $\times$ 512 pixels for 5,014 images, 225 $\times$ 225 for 345 and 630 $\times$ 630 for 90, followed by a long tail. The slices include axial, coronal and sagittal planes and several MRI contrasts (Figure~\ref{fig:1}). They are therefore not limited to T1-weighted contrast-enhanced axial images. The collection is a composite of three earlier public datasets: Figshare~\cite{r39}, SARTAJ~\cite{r40} and Br35H~\cite{r41}. Section 3.2 describes how each image was attributed to its source, and Figure~\ref{fig:2} gives an overview of the complete pipeline.

\begin{table}[!htbp]
\centering
\footnotesize
\setlength{\tabcolsep}{3.0pt}
\renewcommand{\arraystretch}{1.12}
\caption{Class distribution of the Kaggle folders and of the partitions used in this study. The training/validation split is drawn only from the Training folder and is grouped by Figshare patient and verified near-duplicate; the Testing folder is the test set.}
\label{tab:2}
\begin{tabular}{>{\raggedright\arraybackslash}p{0.211\linewidth}>{\raggedright\arraybackslash}p{0.146\linewidth}>{\raggedright\arraybackslash}p{0.146\linewidth}>{\raggedright\arraybackslash}p{0.167\linewidth}>{\raggedright\arraybackslash}p{0.136\linewidth}>{\raggedright\arraybackslash}p{0.115\linewidth}}
\toprule
\nobreak\hspace{0pt}\textbf{Class} & \nobreak\hspace{0pt}\textbf{Kaggle Training folder} & \nobreak\hspace{0pt}\textbf{Training (this study)} & \nobreak\hspace{0pt}\textbf{Validation (this study)} & \nobreak\hspace{0pt}\textbf{Kaggle Testing folder = test set} & \nobreak\hspace{0pt}\textbf{Total} \\
\midrule
\nobreak\hspace{0pt}Glioma & \nobreak\hspace{0pt}1,400 & \nobreak\hspace{0pt}1,129 & \nobreak\hspace{0pt}271 & \nobreak\hspace{0pt}400 & \nobreak\hspace{0pt}1,800 \\
\nobreak\hspace{0pt}Meningioma & \nobreak\hspace{0pt}1,400 & \nobreak\hspace{0pt}1,151 & \nobreak\hspace{0pt}249 & \nobreak\hspace{0pt}400 & \nobreak\hspace{0pt}1,800 \\
\nobreak\hspace{0pt}No tumor & \nobreak\hspace{0pt}1,416 & \nobreak\hspace{0pt}1,124 & \nobreak\hspace{0pt}292 & \nobreak\hspace{0pt}400 & \nobreak\hspace{0pt}1,816 \\
\nobreak\hspace{0pt}Pituitary & \nobreak\hspace{0pt}1,400 & \nobreak\hspace{0pt}1,119 & \nobreak\hspace{0pt}281 & \nobreak\hspace{0pt}400 & \nobreak\hspace{0pt}1,800 \\
\nobreak\hspace{0pt}\textbf{Total} & \nobreak\hspace{0pt}\textbf{5,616} & \nobreak\hspace{0pt}\textbf{4,523} & \nobreak\hspace{0pt}\textbf{1,093} & \nobreak\hspace{0pt}\textbf{1,600} & \nobreak\hspace{0pt}\textbf{7,216} \\
\bottomrule
\end{tabular}
\end{table}

\subsection{Source Attribution and Patient Identifiers}

The three source datasets were downloaded from their original repositories. Every Kaggle image was compared with every source image using a 64-bit perceptual hash (pHash)~\cite{r68}. An image was attributed to the source holding its nearest hash, provided the Hamming distance was at most 8 bits; otherwise it was labelled ``unmatched''. The Figshare dataset~\cite{r39} contains 3,064 T1-weighted contrast-enhanced slices from 233 patients, each stored with the patient identifier and an expert tumor mask. For Kaggle images attributed to Figshare, the patient identifier was transferred. The expert mask was transferred only when the two images had identical dimensions and their pixel intensities were correlated (normalised cross-correlation above 0.9), which guarantees that the mask is aligned with the Kaggle image. Table~\ref{tab:3} gives the source composition of each partition.

\begin{table}[!htbp]
\centering
\footnotesize
\setlength{\tabcolsep}{3.0pt}
\renewcommand{\arraystretch}{1.12}
\caption{Source dataset of each image by partition and class, from perceptual-hash matching to the original Figshare, SARTAJ and Br35H datasets (``unmatched'': no source image within 8 bits).}
\label{tab:3}
\begin{tabular}{>{\raggedright\arraybackslash}p{0.178\linewidth}>{\raggedright\arraybackslash}p{0.178\linewidth}>{\raggedright\arraybackslash}p{0.115\linewidth}>{\raggedright\arraybackslash}p{0.119\linewidth}>{\raggedright\arraybackslash}p{0.091\linewidth}>{\raggedright\arraybackslash}p{0.126\linewidth}>{\raggedright\arraybackslash}p{0.099\linewidth}}
\toprule
\nobreak\hspace{0pt}\textbf{Partition} & \nobreak\hspace{0pt}\textbf{Class} & \nobreak\hspace{0pt}\textbf{Figshare} & \nobreak\hspace{0pt}\textbf{SARTAJ} & \nobreak\hspace{0pt}\textbf{Br35H} & \nobreak\hspace{0pt}\textbf{Unmatched} & \nobreak\hspace{0pt}\textbf{Total} \\
\midrule
\nobreak\hspace{0pt}Training & \nobreak\hspace{0pt}Glioma & \nobreak\hspace{0pt}371 & \nobreak\hspace{0pt}67 & \nobreak\hspace{0pt}14 & \nobreak\hspace{0pt}677 & \nobreak\hspace{0pt}1,129 \\
 & \nobreak\hspace{0pt}Meningioma & \nobreak\hspace{0pt}602 & \nobreak\hspace{0pt}144 & \nobreak\hspace{0pt}52 & \nobreak\hspace{0pt}353 & \nobreak\hspace{0pt}1,151 \\
 & \nobreak\hspace{0pt}No tumor & \nobreak\hspace{0pt}0 & \nobreak\hspace{0pt}150 & \nobreak\hspace{0pt}974 & \nobreak\hspace{0pt}0 & \nobreak\hspace{0pt}1,124 \\
 & \nobreak\hspace{0pt}Pituitary & \nobreak\hspace{0pt}533 & \nobreak\hspace{0pt}151 & \nobreak\hspace{0pt}8 & \nobreak\hspace{0pt}427 & \nobreak\hspace{0pt}1,119 \\
\nobreak\hspace{0pt}Validation & \nobreak\hspace{0pt}Glioma & \nobreak\hspace{0pt}82 & \nobreak\hspace{0pt}8 & \nobreak\hspace{0pt}4 & \nobreak\hspace{0pt}177 & \nobreak\hspace{0pt}271 \\
 & \nobreak\hspace{0pt}Meningioma & \nobreak\hspace{0pt}102 & \nobreak\hspace{0pt}30 & \nobreak\hspace{0pt}8 & \nobreak\hspace{0pt}109 & \nobreak\hspace{0pt}249 \\
 & \nobreak\hspace{0pt}No tumor & \nobreak\hspace{0pt}0 & \nobreak\hspace{0pt}37 & \nobreak\hspace{0pt}255 & \nobreak\hspace{0pt}0 & \nobreak\hspace{0pt}292 \\
 & \nobreak\hspace{0pt}Pituitary & \nobreak\hspace{0pt}125 & \nobreak\hspace{0pt}45 & \nobreak\hspace{0pt}1 & \nobreak\hspace{0pt}110 & \nobreak\hspace{0pt}281 \\
\nobreak\hspace{0pt}Test & \nobreak\hspace{0pt}Glioma & \nobreak\hspace{0pt}127 & \nobreak\hspace{0pt}57 & \nobreak\hspace{0pt}41 & \nobreak\hspace{0pt}175 & \nobreak\hspace{0pt}400 \\
 & \nobreak\hspace{0pt}Meningioma & \nobreak\hspace{0pt}90 & \nobreak\hspace{0pt}85 & \nobreak\hspace{0pt}49 & \nobreak\hspace{0pt}176 & \nobreak\hspace{0pt}400 \\
 & \nobreak\hspace{0pt}No tumor & \nobreak\hspace{0pt}0 & \nobreak\hspace{0pt}9 & \nobreak\hspace{0pt}391 & \nobreak\hspace{0pt}0 & \nobreak\hspace{0pt}400 \\
 & \nobreak\hspace{0pt}Pituitary & \nobreak\hspace{0pt}88 & \nobreak\hspace{0pt}21 & \nobreak\hspace{0pt}6 & \nobreak\hspace{0pt}285 & \nobreak\hspace{0pt}400 \\
\bottomrule
\end{tabular}
\end{table}

\subsection{Duplicate Screening}

Exact duplicates were identified by hashing the decoded grayscale pixels. Near-duplicates were screened in two stages, because brain MRI slices share anatomy and dark backgrounds, so hash similarity alone also flags distinct images. First, all pairs of images with a pHash Hamming distance of at most 6 bits were taken as candidates. Second, a candidate pair was accepted as a near-duplicate only if the normalised cross-correlation of its 56 $\times$ 56 grayscale thumbnails was at least 0.95:

\begin{equation}
\mathrm{NCC}(a,b) = \frac{1}{d}\sum_{i = 1}^{d}\frac{\left( a_{i} - \bar{a} \right)\left( b_{i} - \bar{b} \right)}{\sigma_{a}\,\sigma_{b}}
\end{equation}

where $a$ and $b$ are the flattened thumbnails ($d = 3,136$ pixels). Accepted pairs were merged into duplicate groups (connected components). The clean test subset comprises the test images that have no verified near-duplicate among the Training-folder images and no Figshare patient in common with them. The results of the audit are summarised in Table~\ref{tab:4} and described in Section 4.1.

\begin{table}[!htbp]
\centering
\small
\setlength{\tabcolsep}{3.0pt}
\renewcommand{\arraystretch}{1.12}
\caption{Data audit: duplicates, patient overlap and the clean test subset.}
\label{tab:4}
\begin{tabular}{>{\raggedright\arraybackslash}p{0.533\linewidth}>{\raggedright\arraybackslash}p{0.437\linewidth}}
\toprule
\nobreak\hspace{0pt}\textbf{Quantity} & \nobreak\hspace{0pt}\textbf{Value} \\
\midrule
\nobreak\hspace{0pt}Images in the Kaggle composite & \nobreak\hspace{0pt}7,216 (Training folder 5,616; Testing folder 1,600) \\
\nobreak\hspace{0pt}Exact-duplicate groups (images) & \nobreak\hspace{0pt}288 (626) \\
\nobreak\hspace{0pt}Candidate near-duplicate pairs (pHash Hamming distance $\leq$ 6) & \nobreak\hspace{0pt}8,922 (median NCC 0.978; 5th percentile 0.705) \\
\nobreak\hspace{0pt}Verified near-duplicate pairs (NCC $\geq$ 0.95) & \nobreak\hspace{0pt}5,786 \\
\nobreak\hspace{0pt}of which cross the Training/Testing folders & \nobreak\hspace{0pt}2,077 \\
\nobreak\hspace{0pt}of which have conflicting class labels & \nobreak\hspace{0pt}10 \\
\nobreak\hspace{0pt}Verified duplicate groups with more than one image (largest) & \nobreak\hspace{0pt}688 (28 images) \\
\nobreak\hspace{0pt}Test images with a verified near-duplicate in the Training folder & \nobreak\hspace{0pt}569 (35.6\%) \\
\nobreak\hspace{0pt}Figshare patients in Training / Testing / both & \nobreak\hspace{0pt}215 / 146 / 141 \\
\nobreak\hspace{0pt}Figshare-matched test images from patients also in Training & \nobreak\hspace{0pt}296 of 305 (97.0\%) \\
\nobreak\hspace{0pt}Clean test subset (no verified duplicate, no shared patient) & \nobreak\hspace{0pt}773 (48.3\%) \\
\nobreak\hspace{0pt}Test images with pixel-aligned expert tumor masks & \nobreak\hspace{0pt}165 (6 in the clean subset) \\
\nobreak\hspace{0pt}Groups used for the training/validation split (largest) & \nobreak\hspace{0pt}2,904 (64 images) \\
\bottomrule
\end{tabular}
\end{table}

\subsection{Data Splitting}

The predefined Testing folder (1,600 images) was used unchanged as the test set. This preserves comparability with other studies of the same collection, while its overlap with the training data is quantified (Section 4.1) and the clean subset is reported alongside it. The Training folder was split into training and validation partitions with a stratified, group-aware five-fold split (StratifiedGroupKFold, seed 42). The groups merge verified near-duplicate components with Figshare patient identifiers, so that no duplicate group and no patient appears in both partitions. The fold whose size was closest to 20\% was used for validation (1,093 images, 19.5\%; training 4,523 images; Table~\ref{tab:2}). Two checks were verified programmatically: no Testing-folder image appears in the training or validation partitions, and no group spans the two. The split files, with the group and source of every image, are available from the corresponding author on reasonable request.

\subsection{Preprocessing and Augmentation}

Each image was read as a single grayscale channel and resized with bilinear interpolation to the input resolution of each model (Table~\ref{tab:5}). The same grayscale channel was then replicated three times to match the three-channel input expected by the pretrained backbones. Images were normalised with the ImageNet mean and standard deviation, or with mean = standard deviation = 0.5 for MedSigLIP, following its native preprocessing. Training images were augmented independently per image with the following transformations: horizontal flip (probability 0.5), vertical flip (probability 0.2), rotation of up to $\pm$15$^{\circ}$, translation of up to $\pm$5\% of the image size, and brightness and contrast scaling by factors drawn from 0.8--1.2. Saturation and hue jitter were not used because they are meaningless for grayscale images. Validation and test images were only resized and normalised. For the shortcut analyses (Section 3.11), an approximate brain mask was computed for every image: Gaussian smoothing, Otsu thresholding, retention of the largest connected component, hole filling and a three-pixel dilation.

\subsection{Architectures and Training}

Six architectures span three model families (Table~\ref{tab:5}). The two CNNs are EfficientNetV2-M~\cite{r69,r70,r71}, which combines Fused-MBConv blocks in its early stages with MBConv blocks and squeeze-and-excitation in its later stages, and ConvNeXt-V2-Tiny~\cite{r26,r72,r73}, a modernised convolutional network with 7 $\times$ 7 depthwise convolutions, layer normalisation and global response normalisation, pretrained with a fully convolutional masked autoencoder. The two vision transformers are Swin-V2-S~\cite{r24,r37,r74}, which applies self-attention within shifted local windows in a four-stage hierarchy, and ViT-L/16~\cite{r22,r53,r75}, a 24-layer transformer over 16 $\times$ 16 patches pretrained on ImageNet-21K. The two foundation models are DINOv2-L~\cite{r27}, a ViT-Large trained by self-supervised self-distillation, whose classifier receives the class token concatenated with the mean patch token, and MedSigLIP-448~\cite{r28,r29,r76,r77}, the 448-pixel vision encoder of a SigLIP model~\cite{r78}, which replaces the softmax contrastive loss of CLIP~\cite{r79} with a pairwise sigmoid loss and was released with MedGemma after pretraining on medical image--text pairs.

Each backbone was loaded with its published weights and given a new four-way linear classification head. All layers were fine-tuned end to end.

All models were trained with the same protocol. Training used label-smoothed cross-entropy ($\varepsilon$ = 0.1)~\cite{r80}, cosine learning-rate annealing over 50 epochs and gradient-norm clipping at 1.0. Early stopping monitored the validation loss (patience 5 epochs, minimum improvement 10$^{-}$$^{4}$), and the checkpoint with the highest validation accuracy was retained. Optimisers, learning rates and batch sizes are given in Table~\ref{tab:5}. Training used bfloat16 mixed precision with a fixed seed (42) on a single NVIDIA A100-SXM4-80GB GPU (Google Colab), with PyTorch, timm and Hugging Face Transformers. Three additional runs serve as ablations (Section 3.10). The first is MedSigLIP at 224 pixels, with interpolated positional embeddings. The second is ConvNeXt-V2-Tiny at 448 pixels. The third is ConvNeXt-V2-Tiny at 448 pixels with MedSigLIP's batch size (8) and normalisation (the ``matched'' run).

\begin{table}[!htbp]
\centering
\scriptsize
\setlength{\tabcolsep}{3.0pt}
\renewcommand{\arraystretch}{1.12}
\caption{Architectures and training configuration. All runs used label smoothing 0.1, cosine annealing over 50 epochs, gradient clipping at 1.0, early stopping on validation loss (patience 5) and the checkpoint with the highest validation accuracy. Rows in italics are ablation runs.}
\label{tab:5}
\begin{tabular}{>{\raggedright\arraybackslash}p{0.188\linewidth}>{\raggedright\arraybackslash}p{0.208\linewidth}>{\raggedright\arraybackslash}p{0.070\linewidth}>{\raggedright\arraybackslash}p{0.059\linewidth}>{\raggedright\arraybackslash}p{0.141\linewidth}>{\raggedright\arraybackslash}p{0.115\linewidth}>{\raggedright\arraybackslash}p{0.058\linewidth}>{\raggedright\arraybackslash}p{0.057\linewidth}}
\toprule
\nobreak\hspace{0pt}\textbf{Model} & \nobreak\hspace{0pt}\textbf{Family / pretraining} & \nobreak\hspace{0pt}\textbf{Params (M)} & \nobreak\hspace{0pt}\textbf{Input (px)} & \nobreak\hspace{0pt}\textbf{Normalisation} & \nobreak\hspace{0pt}\textbf{Optimiser, LR, WD} & \nobreak\hspace{0pt}\textbf{Batch} & \nobreak\hspace{0pt}\textbf{Best / stop epoch} \\
\midrule
\nobreak\hspace{0pt}EfficientNetV2-M & \nobreak\hspace{0pt}CNN, ImageNet-1K & \nobreak\hspace{0pt}52.86 & \nobreak\hspace{0pt}224 & \nobreak\hspace{0pt}ImageNet & \nobreak\hspace{0pt}Adam, 1e-4, 1e-5 & \nobreak\hspace{0pt}32 & \nobreak\hspace{0pt}17 / 18 \\
\nobreak\hspace{0pt}ConvNeXt-V2-Tiny & \nobreak\hspace{0pt}CNN, FCMAE + IN-22K/1K & \nobreak\hspace{0pt}27.87 & \nobreak\hspace{0pt}224 & \nobreak\hspace{0pt}ImageNet & \nobreak\hspace{0pt}AdamW, 5e-5, 0.01 & \nobreak\hspace{0pt}32 & \nobreak\hspace{0pt}5 / 11 \\
\nobreak\hspace{0pt}Swin-V2-S & \nobreak\hspace{0pt}Hierarchical ViT, IN-1K & \nobreak\hspace{0pt}48.96 & \nobreak\hspace{0pt}256 & \nobreak\hspace{0pt}ImageNet & \nobreak\hspace{0pt}AdamW, 5e-5, 0.01 & \nobreak\hspace{0pt}32 & \nobreak\hspace{0pt}6 / 14 \\
\nobreak\hspace{0pt}ViT-L/16 & \nobreak\hspace{0pt}ViT, IN-21K & \nobreak\hspace{0pt}303.31 & \nobreak\hspace{0pt}224 & \nobreak\hspace{0pt}ImageNet & \nobreak\hspace{0pt}AdamW, 3e-5, 0.01 & \nobreak\hspace{0pt}32 & \nobreak\hspace{0pt}8 / 18 \\
\nobreak\hspace{0pt}DINOv2-L & \nobreak\hspace{0pt}ViT, self-supervised (LVD-142M) & \nobreak\hspace{0pt}304.38 & \nobreak\hspace{0pt}224 & \nobreak\hspace{0pt}ImageNet & \nobreak\hspace{0pt}AdamW, 2e-5, 0.01 & \nobreak\hspace{0pt}32 & \nobreak\hspace{0pt}6 / 11 \\
\nobreak\hspace{0pt}MedSigLIP-448 & \nobreak\hspace{0pt}ViT, medical image-text (SigLIP) & \nobreak\hspace{0pt}428.57 & \nobreak\hspace{0pt}448 & \nobreak\hspace{0pt}mean = SD = 0.5 & \nobreak\hspace{0pt}AdamW, 2e-5, 0.01 & \nobreak\hspace{0pt}8 & \nobreak\hspace{0pt}22 / 27 \\
\nobreak\hspace{0pt}\textit{MedSigLIP (224 px)} & \nobreak\hspace{0pt}\textit{resolution ablation} & \nobreak\hspace{0pt}428.57 & \nobreak\hspace{0pt}224 & \nobreak\hspace{0pt}mean = SD = 0.5 & \nobreak\hspace{0pt}AdamW, 2e-5, 0.01 & \nobreak\hspace{0pt}8 & \nobreak\hspace{0pt}8 / 8 \\
\nobreak\hspace{0pt}\textit{ConvNeXt-V2-Tiny (448 px)} & \nobreak\hspace{0pt}\textit{resolution ablation} & \nobreak\hspace{0pt}27.87 & \nobreak\hspace{0pt}448 & \nobreak\hspace{0pt}ImageNet & \nobreak\hspace{0pt}AdamW, 5e-5, 0.01 & \nobreak\hspace{0pt}32 & \nobreak\hspace{0pt}10 / 15 \\
\nobreak\hspace{0pt}\textit{ConvNeXt-V2-Tiny (448 px, matched)} & \nobreak\hspace{0pt}\textit{preprocessing-matched ablation} & \nobreak\hspace{0pt}27.87 & \nobreak\hspace{0pt}448 & \nobreak\hspace{0pt}mean = SD = 0.5 & \nobreak\hspace{0pt}AdamW, 5e-5, 0.01 & \nobreak\hspace{0pt}8 & \nobreak\hspace{0pt}8 / 13 \\
\bottomrule
\end{tabular}
\end{table}

\subsection{Evaluation Metrics and Statistical Analysis}

All results refer to the 1,600-image test set unless stated otherwise. The metrics are accuracy, balanced accuracy, macro-averaged precision, recall and F1, Matthews correlation coefficient (MCC), Cohen's kappa~\cite{r81}, macro-averaged one-versus-rest area under the ROC curve (AUC)~\cite{r82,r83}, log-loss, the multiclass Brier score and the expected calibration error (ECE). Per class, we report sensitivity, specificity, precision, F1, one-versus-rest AUC and average precision. In terms of true and false positives and negatives (TP, TN, FP, FN):

\begin{equation}
\mathrm{Accuracy} = \frac{TP + TN}{TP + TN + FP + FN},\quad\quad F_{1} = \frac{2\, \mathrm{Precision} \times \mathrm{Recall}}{\mathrm{Precision} + \mathrm{Recall}}
\end{equation}

\begin{equation}
\mathrm{MCC} = \frac{TP \times TN - FP \times FN}{\sqrt{(TP + FP)(TP + FN)(TN + FP)(TN + FN)}}
\end{equation}

The multiclass MCC was computed from the full confusion matrix~\cite{r84,r85}. Uncertainty was quantified by non-parametric bootstrap resampling of test images (2,000 resamples, percentile 95\% intervals)~\cite{r86}. Accuracy differences between models were given paired-bootstrap intervals. Models were compared pairwise with the exact (binomial) McNemar test on the discordant predictions; the continuity-corrected statistic

\begin{equation}
\chi^{2} = \frac{\left( |b - c| - 1 \right)^{2}}{b + c}
\end{equation}

is also reported, where $b$ and $c$ count the images classified correctly by only one of the two models~\cite{r87}. The 15 pairwise p-values among the six primary models were adjusted with the Holm procedure~\cite{r88}, and Cochran's Q test~\cite{r89} was applied as an omnibus test across the six models.

\subsection{Calibration}

For each model, a single temperature $T$ was fitted on the validation logits $z$ by minimising the negative log-likelihood, and calibrated probabilities $\widehat{p} = \mathrm{softmax}(z/T)$ were used in all subsequent analyses~\cite{r66}. Temperature scaling does not change argmax predictions. Calibration on the test set is reported before and after scaling as ECE with 15 equal-width bins,

\begin{equation}
\mathrm{ECE} = \sum_{m = 1}^{M}\frac{\left| B_{m} \right|}{n}\left| \mathrm{acc}\left( B_{m} \right) - \mathrm{conf}\left( B_{m} \right) \right|
\end{equation}

together with ECE using 15 equal-mass bins, class-wise ECE, the Brier score and NLL, each with a bootstrap 95\% interval.

\subsection{Ensembles and Ensemble Diversity}

Four ensembles were constructed from the calibrated probabilities of the six primary models, and every ensemble parameter was fitted on the validation set. The simple average is the mean of the six probability vectors and fits nothing. The weighted average uses non-negative weights summing to one, chosen to minimise the validation NLL. The majority vote uses hard votes, with ties broken by the mean probability; its vote fractions serve as scores for the AUC, so its AUC and log-loss are computed from its own outputs rather than from the averaged probabilities. Stacking is a multinomial logistic regression on the validation log-probabilities, with its regularisation chosen by five-fold cross-validation within the validation set.

All four ensembles, and the single model with the best validation accuracy, were evaluated once on the same 1,600 test images and compared with the best single model on test using McNemar tests. To assess whether the architectures make complementary errors, we report pairwise disagreement, double-fault rates, Yule's Q, prediction kappa, the distribution of the number of models correct per image, and the oracle accuracy (the share of images that at least one model classifies correctly).

A sensitivity analysis also fitted the weighted average, the stacker and the decision thresholds by five-fold cross-fitting within the test set. Each test image was predicted with parameters fitted on the other four folds, and the folds were group-aware with respect to Figshare patients and verified duplicates. This estimate never fits a parameter on the image it predicts, but it does use test-distribution data for fitting, so it is reported as a sensitivity analysis only.

\subsection{Decision Rule, Thresholds, and Ablations}

The networks output four logits. The primary prediction is the argmax of the calibrated probabilities, which does not require any probability to exceed 0.5. One-versus-rest thresholds $t_{k}$ maximising Youden's J~\cite{r90} were fitted on the validation set and applied once to the test set. Two explicit rules convert the four thresholds into a single prediction. The \textit{scaled argmax} rule predicts $\arg\max_{k}{\widehat{p}}_{k}/t_{k}$ and always yields exactly one class. The one-flag rule predicts the single class that exceeds its threshold when exactly one does, and otherwise falls back to the argmax. The ablations compare each pair of runs that differ in a single factor: MedSigLIP at 448 and 224 pixels, ConvNeXt-V2-Tiny at 224 and 448 pixels, and MedSigLIP-448 against ConvNeXt-V2-Tiny at 448 pixels with matched batch size and normalisation, which isolates architecture and pretraining.

\subsection{Shortcut Tests and Feature-Space Analysis}

Four analyses examined whether performance depends on non-pathological cues. First, each fine-tuned model was evaluated on brain-only test images, with the background outside the brain mask set to zero, and on background-only images, with the brain region set to zero. Second, logistic-regression probes were trained on the frozen, non-fine-tuned features of ConvNeXt-V2-Tiny, DINOv2-L and MedSigLIP-448 for full, brain-only and background-only training images, and were evaluated on the corresponding test images. Third, for tumor-class images of known Figshare or SARTAJ origin in the validation and test sets (n = 860), the source dataset was predicted from the fine-tuned embeddings by logistic regression with five-fold cross-validation and compared with the majority-class baseline. Fourth, t-SNE (perplexity 30, PCA initialisation) and UMAP (30 neighbours, minimum distance 0.1, seed 42) projections were computed from the z-scored penultimate embeddings of all 1,600 test images (400 per class) and coloured by class and by source; the coordinates, labels, sample identifiers, parameters and embeddings are available from the corresponding author on reasonable request.

\subsection{Explainability and Its Quantitative Evaluation}

\textbf{Methods and target layers: }Grad-CAM~\cite{r50}, Grad-CAM++~\cite{r58} and EigenCAM~\cite{r63} were computed for all six models at documented target layers: the last convolutional stage of EfficientNetV2-M, the last stage of ConvNeXt-V2-Tiny, and the last stage of Swin-V2-S after permuting its channels-last output to channels-first. For ViT-L/16 the target was the token output of the penultimate block with the class token removed (14 $\times$ 14 grid), because the classification head reads only the class token and the patch tokens of the final layer therefore receive no gradient. For DINOv2-L it was the final layer-normalised tokens (16 $\times$ 16 grid), and for MedSigLIP-448 the post-layer-normalised tokens (32 $\times$ 32 grid).

Integrated Gradients~\cite{r62} used 32 steps from a black baseline, with absolute attributions summed over channels; LIME~\cite{r60} used 300 perturbations and kept positive superpixel weights; and SHAP GradientExplainer used 50 training images as background. Maps were computed for the true class, min--max normalised per image and resized bilinearly to 224 $\times$ 224 pixels for evaluation. Figures show sample identifiers, the true class, and the predicted class and probability.

\textbf{Localisation against expert masks:} Of the test images, 165 are pixel-aligned with a Figshare original and therefore have an expert tumor mask; 6 of these belong to the clean subset. We report three measures. The pointing-game hit rate~\cite{r65} is the share of images whose attribution maximum falls inside the mask. The energy-in-mask fraction is

\begin{equation}
E_{\mathrm{mask}} = \frac{\sum_{x}A(x)\, M(x)}{\sum_{x}A(x)}
\end{equation}

where $A$ is the non-negative attribution map and $M$ the binary mask; its chance level equals the mask's area fraction. The third measure is the intersection-over-union of the top-20\% attribution region with the mask. Grad-CAM, Grad-CAM++ and EigenCAM were evaluated on all 165 images; Integrated Gradients and LIME, which are more costly, on 40 and 20 of them.

\textbf{Sanity checks, stability and ensemble explanation: }Following Adebayo et al.~\cite{r64}, Grad-CAM maps were recomputed for 20 images after cascading re-initialisation of the classifier head and then of progressively deeper blocks, and their Spearman correlations with the original maps were computed; maps that do not change under randomisation are uninformative. Stability was assessed by comparing maps after adding Gaussian noise ($\sigma$ = 0.02 on the [0, 1] intensity scale) and after a horizontal flip (mapped back), using Spearman correlation and the structural similarity index. Because explanations of individual base models do not explain an ensemble, the weighted-average ensemble was explained directly by occlusion sensitivity: each of 196 patches (a 14 $\times$ 14 grid of 32-pixel patches on the 448-pixel image) was blanked in turn, the occluded image was passed through all six models, and the drop in the ensemble's probability for the true class was recorded (40 images). For the same 40 images, we also measured the share of Grad-CAM energy inside the brain mask, within the outer 10\% border of the image and inside the tumor mask.

\textbf{Blinded expert rating:} A blinded rating package was prepared for neuroradiologists: randomised Grad-CAM overlays, with model identities concealed, and a rating form.

\subsection{Decision-Analytic Evaluation (Exploratory)}

We evaluated one explicit, hypothetical decision: whether an MRI study should be prioritised for expedited neuroradiology reading because the model indicates a tumor of any of the three types. The model's tumor probability is 1 $-$ p(no tumor). The comparators are prioritising no studies (the standard reading order) and prioritising all studies. A false positive is an unnecessary prioritisation, which delays other studies; a false negative leaves a tumor study in the standard queue. The threshold probability $p_{t}$ expresses the exchange rate between these errors: choosing $p_{t}$ means accepting $\left( 1 - p_{t} \right)/p_{t}$ unnecessary prioritisations in order to prioritise one tumor. Because an unnecessary prioritisation is inexpensive relative to a delayed tumor report, the relevant range is 5--30\%. Net benefit~\cite{r67} is

\begin{equation}
\mathrm{NB}\left( p_{t} \right) = \frac{TP}{n} - \frac{FP}{n} \cdot \frac{p_{t}}{1 - p_{t}}
\end{equation}

The test set is 75\% tumor by construction, so results were re-weighted to illustrative tumor prevalences of 10\% and 25\%, and the observed prevalence is also reported. For each scenario, the model probabilities were corrected for the change in class priors~\cite{r91}:

\begin{equation}
{\widehat{p}}_{k}^{\,'} = \frac{{\widehat{p}}_{k}\, q_{k}/r_{k}}{\sum_{j}{\widehat{p}}_{j}\, q_{j}/r_{j}}
\end{equation}

where $r$ are the training class priors and $q$ the scenario priors, with tumor prevalence divided equally among the three tumor classes. Confidence intervals were obtained by bootstrap resampling stratified by outcome (1,000 resamples). The decision curves are reported together with sensitivity, specificity, missed tumors per 1,000 studies and unnecessary prioritisations per 1,000 studies at each threshold. Three further analyses describe diagnostic accuracy: likelihood ratios and diagnostic odds ratios with a Haldane--Anscombe correction and 95\% confidence intervals; precision--recall curves; and selective prediction, in which the least confident images are referred and accuracy on the retained images is reported. This analysis does not establish clinical utility. It is retrospective and image-level and involves no patient outcomes, reader comparison or workflow measurement.

\subsection{External Validation}

BDNeuro-MRI~\cite{r92} is a separately published four-class dataset (glioma, meningioma, pituitary tumor and no tumor) of 5,941 T1-weighted contrast-enhanced images. It is described as originating from Epic \& CSCR Hospital, Bangladesh, and is distributed as train, validation and test folders (4,160, 892 and 889 images). None of these images was used in model development, so all were treated as external data. Every image was screened against the full Kaggle composite with the same two-stage criterion as in Section 3.3, and against the other external images for exact repeats. The primary external set excludes verified near-duplicates of Kaggle images and repeated images. A strict set additionally excludes any image that is a hash candidate (Hamming distance $\leq$ 6) of a Kaggle image. The trained models, temperatures, ensemble weights and stacker were all taken unchanged from the internal analysis, and the external images received the same preprocessing as the test set. Because the external classes are imbalanced, balanced accuracy is reported alongside accuracy, and pairwise differences were tested with Holm-corrected McNemar tests.

\subsection{Implementation and Reproducibility}

The complete pipeline is a single Google Colab notebook. It covers data audit, training, evaluation, explainability, decision analysis and external validation, and it saves every intermediate result so that interrupted runs resume without retraining. The notebook, split files, duplicate pairs, source attributions, per-image annotations, prediction files, result tables and figures are available from the corresponding author on reasonable request.

\section{Results}

\subsection{Data Audit: Duplicates, Sources, and Patient Overlap}

\begin{figure}[!htbp]
\centering
\includegraphics[width=0.75\linewidth,height=0.82\textheight,keepaspectratio]{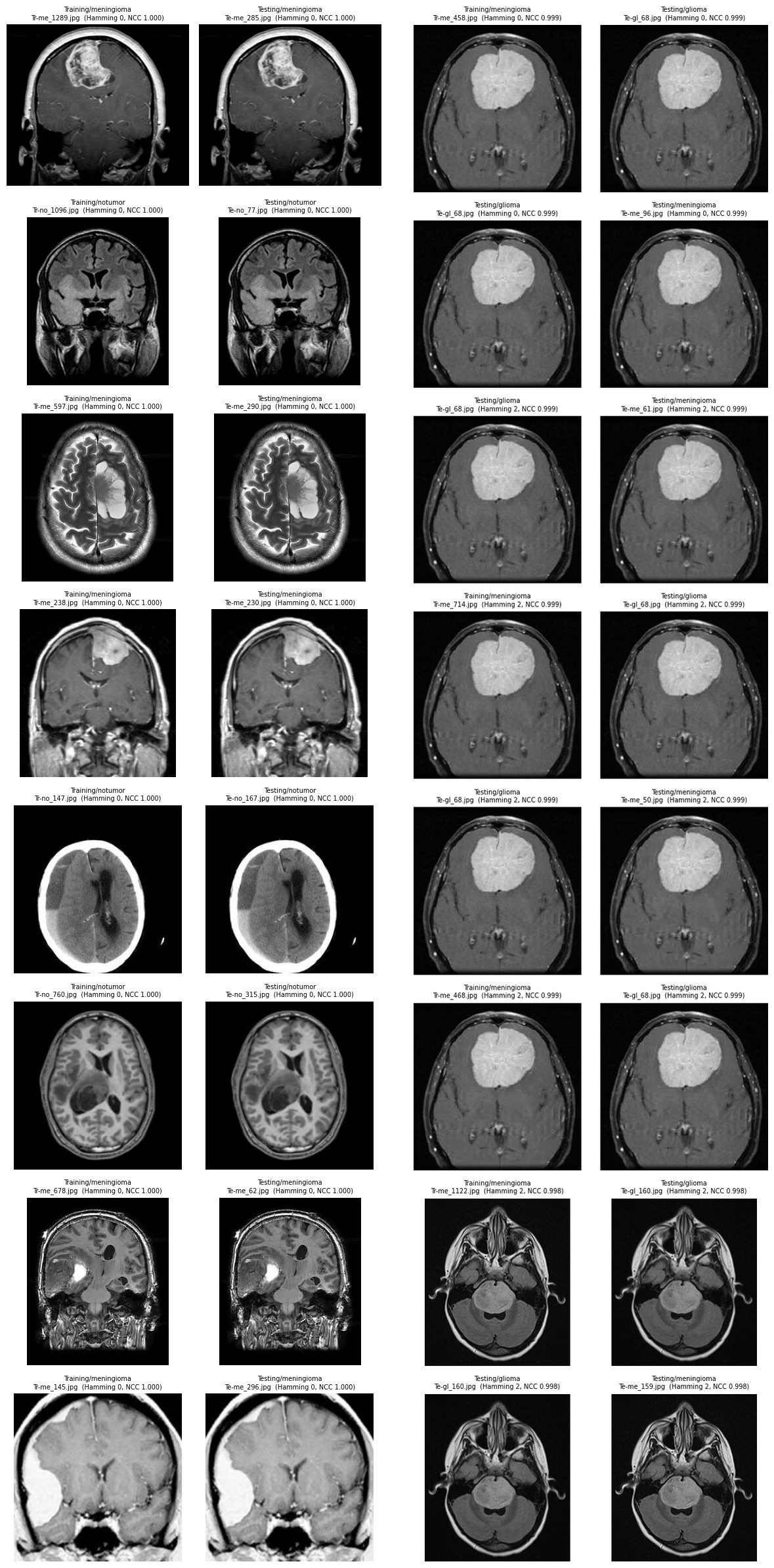}
\caption{(a) The closest pixel-verified near-duplicate pairs that cross the Kaggle Training and Testing folders; (b) verified near-duplicate pairs with conflicting class labels. Each panel shows the file name, folder, class, pHash Hamming distance and pixel correlation (NCC).}
\label{fig:3}
\end{figure}

The data audit is summarised in Table~\ref{tab:4}. The collection contains 288 groups of pixel-identical images, comprising 626 images. Perceptual hashing flagged 8,922 candidate near-duplicate pairs, whose pixel correlation varied widely (median NCC 0.978; 5th percentile 0.705), confirming that hash similarity alone also matches distinct slices. Of these, 5,786 pairs passed pixel verification, and 2,077 link a Training-folder image to a Testing-folder image. Ten verified pairs carry different class labels, which indicates label conflicts in the collection. Verified duplicates form 688 multi-image groups, the largest containing 28 images, and Figure~\ref{fig:3} shows the closest cross-split and cross-class pairs. In all, 569 of the 1,600 test images (35.6\%) have a verified near-duplicate in the Training folder.

The source composition differs sharply between classes (Table~\ref{tab:3}). Almost all no-tumor images come from Br35H (1,229 of 1,416 in the Training folder and 391 of 400 in the test set), whereas the tumor classes come mainly from Figshare and from images that matched no source, so class and source are partly confounded. Of the test images, 305 were matched to Figshare. They belong to 146 patients, 141 of whom also contribute images to the Training folder, so 296 of these 305 test images come from patients seen in training. Removing the test images with a verified duplicate or a shared patient leaves 773 clean test images (48.3\%).

The duplicate-grouping procedure itself matters. In a preliminary analysis that grouped images by hash similarity alone, the unverified matches chained 1,500 training images into a single group, leaving only 180 images for validation. With pixel-verified grouping, the 5,616 Training-folder images form 2,904 groups (largest 64 images), and the validation fold contains 1,093 images (19.5\%).

\subsection{Training Behaviour}

\begin{figure}[!htbp]
\centering
\includegraphics[width=0.69\linewidth,height=0.82\textheight,keepaspectratio]{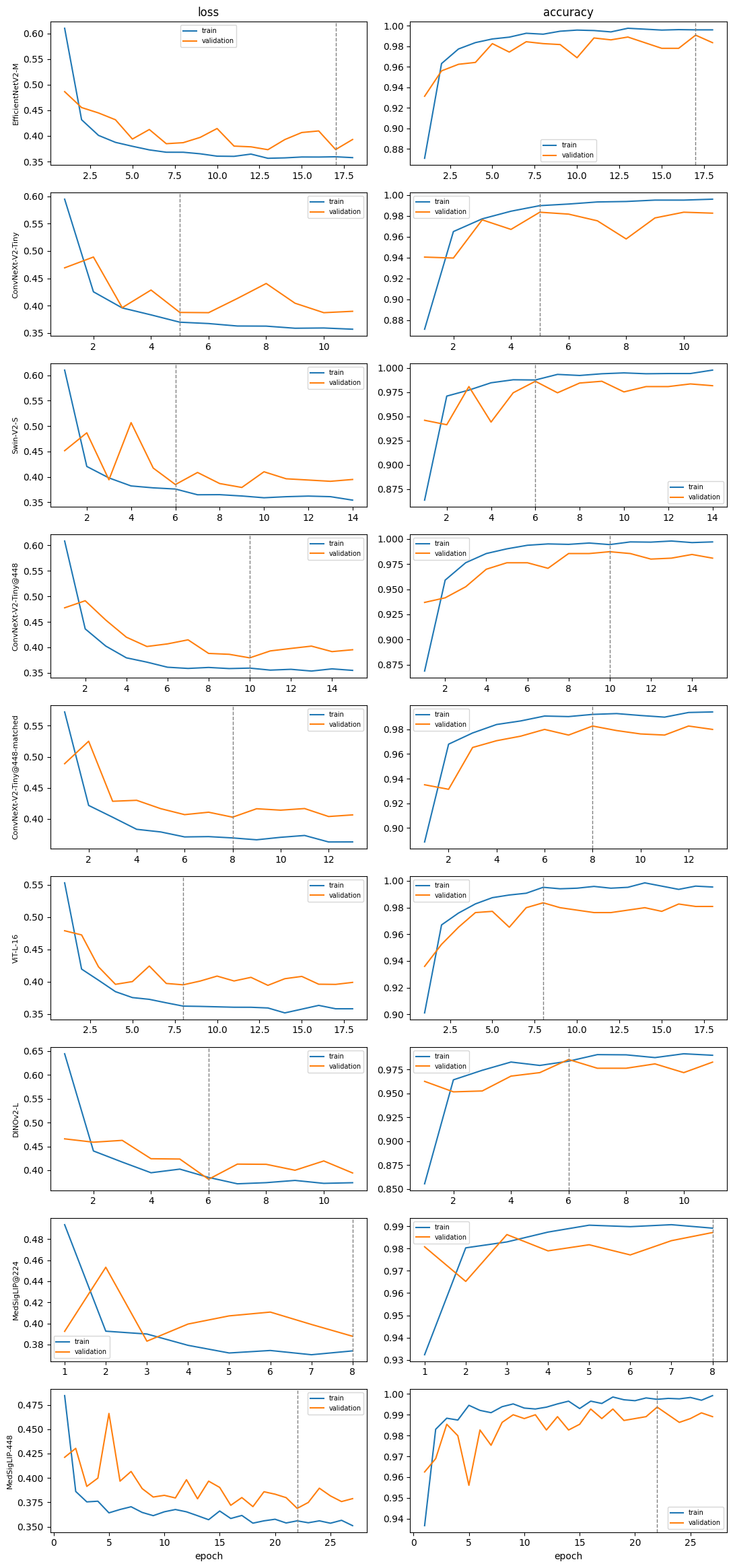}
\caption{Training and validation loss (left) and accuracy (right) for all nine runs. The dashed line marks the epoch of the retained checkpoint (highest validation accuracy).}
\label{fig:4}
\end{figure}

Training curves are shown in Figure~\ref{fig:4}. All nine runs converged and stopped early, after between 8 and 27 epochs, with best validation accuracies of 98.3--99.4\% (Table~\ref{tab:5}). Validation accuracy was 2.8--4.1 percentage points higher than test accuracy for every run (Table~\ref{tab:6}). This gap is consistent with the audit: the validation images come from the same folder and sources as the training images, whereas the test folder contains a different mix of sources (Table~\ref{tab:3}).

\subsection{Internal Test Performance}

Test accuracies of the six primary models fell within one percentage point of each other (Table~\ref{tab:6}). They ranged from 94.50\% (Swin-V2-S) to 95.56\% (EfficientNetV2-M), with overlapping bootstrap intervals, macro AUCs of 0.980--0.986 and MCCs of 0.928--0.942. The best single run overall was the ConvNeXt-V2-Tiny ablation at 448 pixels (95.88\%). On the 773 clean test images, accuracy fell by 3.5--4.2 percentage points, to 90.30--91.72\%, for every primary model.

\begin{figure}[!htbp]
\centering
\includegraphics[width=0.88\linewidth,height=0.82\textheight,keepaspectratio]{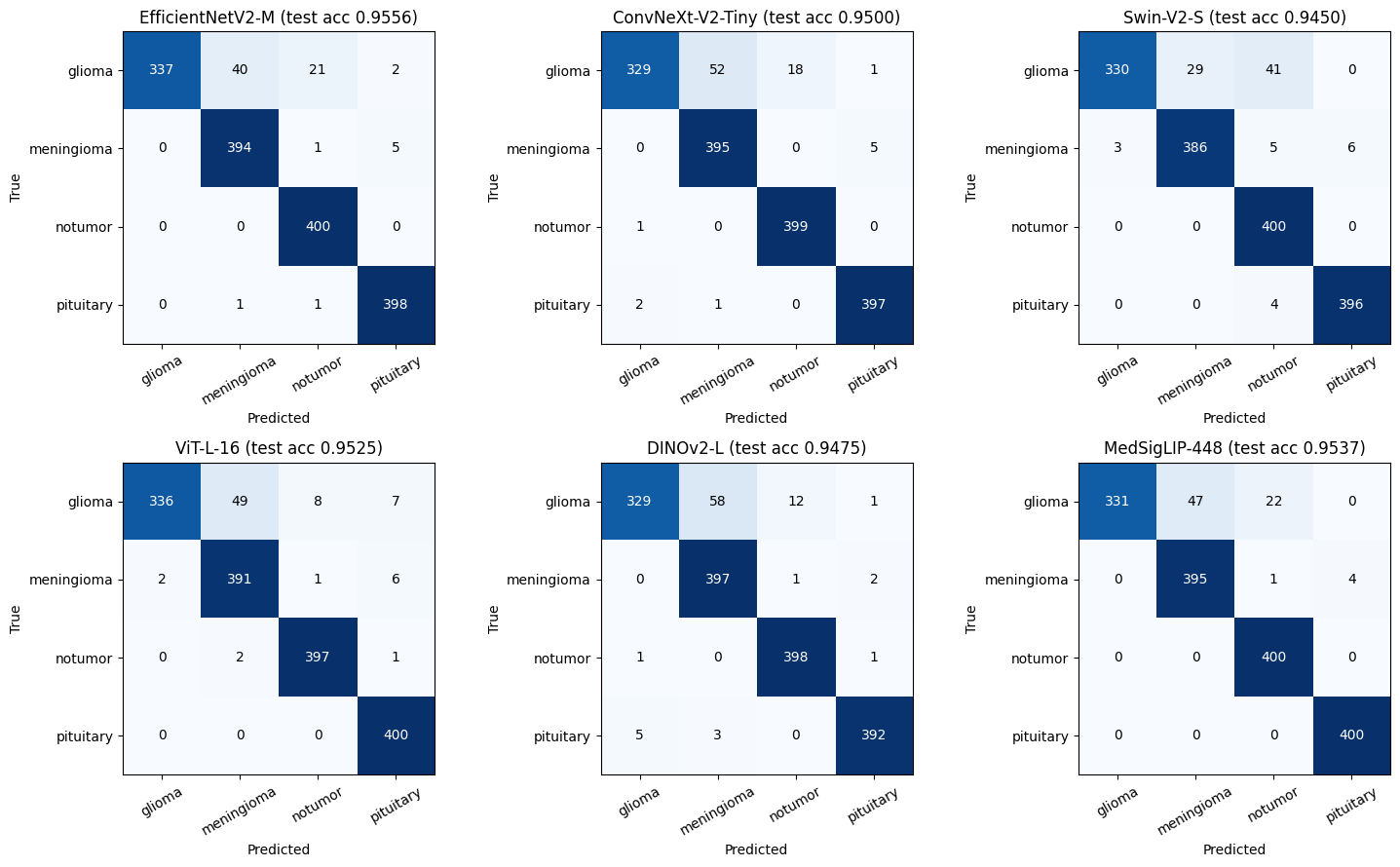}
\caption{Test-set confusion matrices of the six primary models (counts, n = 1,600).}
\label{fig:5}
\end{figure}

\begin{figure}[!htbp]
\centering
\includegraphics[width=0.88\linewidth,height=0.82\textheight,keepaspectratio]{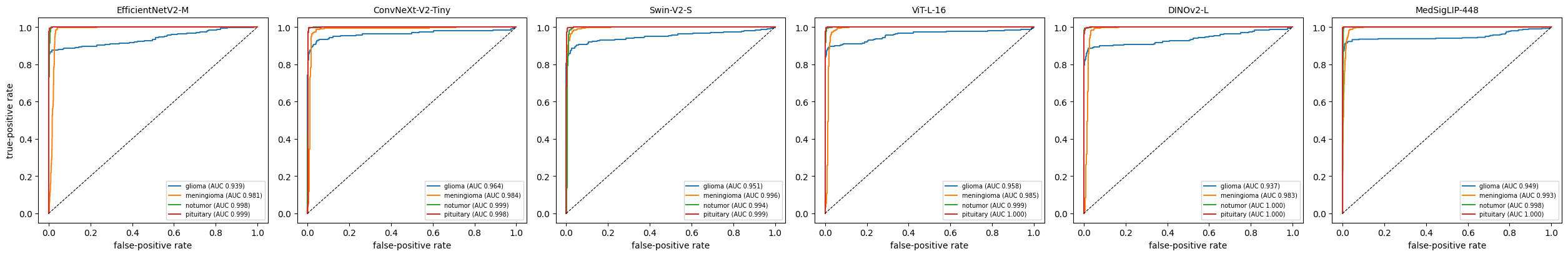}
\caption{One-versus-rest ROC curves per class on the test set, with class-wise AUC.}
\label{fig:6}
\end{figure}

Glioma was the weakest class for every model (Table~\ref{tab:7}). Sensitivity was 0.823--0.843, and glioma accounted for 63--71 of each model's 71--88 errors (Figure~\ref{fig:5}). Gliomas were misclassified mainly as meningioma (29--58 images) and, for some models, as no tumor (8--41 images). The no-tumor and pituitary classes were recognised almost perfectly. The one-versus-rest ROC curves are shown in Figure~\ref{fig:6}.

\begin{table}[!htbp]
\centering
\footnotesize
\setlength{\tabcolsep}{3.0pt}
\renewcommand{\arraystretch}{1.12}
\caption{Validation accuracy, test performance (n = 1,600) with 95\% bootstrap confidence intervals, and accuracy on the clean test subset. Probabilities are temperature-scaled; argmax predictions are unaffected by scaling.}
\label{tab:6}
\begin{tabular}{>{\raggedright\arraybackslash}p{0.216\linewidth}>{\raggedright\arraybackslash}p{0.120\linewidth}>{\raggedright\arraybackslash}p{0.111\linewidth}>{\raggedright\arraybackslash}p{0.122\linewidth}>{\raggedright\arraybackslash}p{0.109\linewidth}>{\raggedright\arraybackslash}p{0.109\linewidth}>{\raggedright\arraybackslash}p{0.120\linewidth}}
\toprule
\nobreak\hspace{0pt}\textbf{Model} & \nobreak\hspace{0pt}\textbf{Validation accuracy} & \nobreak\hspace{0pt}\textbf{Test accuracy [95\% CI]} & \nobreak\hspace{0pt}\textbf{Macro-F1 [95\% CI]} & \nobreak\hspace{0pt}\textbf{Macro AUC [95\% CI]} & \nobreak\hspace{0pt}\textbf{MCC [95\% CI]} & \nobreak\hspace{0pt}\textbf{Clean-test accuracy (n = 773)} \\
\midrule
\nobreak\hspace{0pt}EfficientNetV2-M & \nobreak\hspace{0pt}0.9909 & \nobreak\hspace{0pt}0.9556 [0.9450, 0.9656] & \nobreak\hspace{0pt}0.9548 [0.9441, 0.9648] & \nobreak\hspace{0pt}0.9795 [0.9734, 0.9850] & \nobreak\hspace{0pt}0.9422 [0.9289, 0.9550] & \nobreak\hspace{0pt}0.9133 \\
\nobreak\hspace{0pt}ConvNeXt-V2-Tiny & \nobreak\hspace{0pt}0.9835 & \nobreak\hspace{0pt}0.9500 [0.9394, 0.9600] & \nobreak\hspace{0pt}0.9490 [0.9382, 0.9591] & \nobreak\hspace{0pt}0.9862 [0.9806, 0.9910] & \nobreak\hspace{0pt}0.9351 [0.9215, 0.9478] & \nobreak\hspace{0pt}0.9082 \\
\nobreak\hspace{0pt}Swin-V2-S & \nobreak\hspace{0pt}0.9863 & \nobreak\hspace{0pt}0.9450 [0.9331, 0.9556] & \nobreak\hspace{0pt}0.9441 [0.9323, 0.9547] & \nobreak\hspace{0pt}0.9850 [0.9798, 0.9897] & \nobreak\hspace{0pt}0.9284 [0.9136, 0.9419] & \nobreak\hspace{0pt}0.9030 \\
\nobreak\hspace{0pt}ConvNeXt-V2-Tiny (448 px) & \nobreak\hspace{0pt}0.9872 & \nobreak\hspace{0pt}0.9587 [0.9481, 0.9675] & \nobreak\hspace{0pt}0.9580 [0.9474, 0.9672] & \nobreak\hspace{0pt}0.9810 [0.9747, 0.9870] & \nobreak\hspace{0pt}0.9462 [0.9328, 0.9576] & \nobreak\hspace{0pt}0.9237 \\
\nobreak\hspace{0pt}ConvNeXt-V2-Tiny (448 px, matched) & \nobreak\hspace{0pt}0.9826 & \nobreak\hspace{0pt}0.9419 [0.9306, 0.9525] & \nobreak\hspace{0pt}0.9402 [0.9286, 0.9513] & \nobreak\hspace{0pt}0.9735 [0.9665, 0.9804] & \nobreak\hspace{0pt}0.9250 [0.9111, 0.9385] & \nobreak\hspace{0pt}0.8939 \\
\nobreak\hspace{0pt}ViT-L/16 & \nobreak\hspace{0pt}0.9835 & \nobreak\hspace{0pt}0.9525 [0.9419, 0.9625] & \nobreak\hspace{0pt}0.9518 [0.9411, 0.9616] & \nobreak\hspace{0pt}0.9853 [0.9802, 0.9900] & \nobreak\hspace{0pt}0.9381 [0.9245, 0.9508] & \nobreak\hspace{0pt}0.9172 \\
\nobreak\hspace{0pt}DINOv2-L & \nobreak\hspace{0pt}0.9854 & \nobreak\hspace{0pt}0.9475 [0.9363, 0.9581] & \nobreak\hspace{0pt}0.9468 [0.9355, 0.9573] & \nobreak\hspace{0pt}0.9797 [0.9731, 0.9858] & \nobreak\hspace{0pt}0.9319 [0.9180, 0.9452] & \nobreak\hspace{0pt}0.9082 \\
\nobreak\hspace{0pt}MedSigLIP (224 px) & \nobreak\hspace{0pt}0.9872 & \nobreak\hspace{0pt}0.9531 [0.9425, 0.9631] & \nobreak\hspace{0pt}0.9524 [0.9420, 0.9626] & \nobreak\hspace{0pt}0.9848 [0.9791, 0.9900] & \nobreak\hspace{0pt}0.9388 [0.9254, 0.9517] & \nobreak\hspace{0pt}0.9146 \\
\nobreak\hspace{0pt}MedSigLIP-448 & \nobreak\hspace{0pt}0.9936 & \nobreak\hspace{0pt}0.9538 [0.9431, 0.9637] & \nobreak\hspace{0pt}0.9527 [0.9418, 0.9627] & \nobreak\hspace{0pt}0.9853 [0.9798, 0.9905] & \nobreak\hspace{0pt}0.9401 [0.9268, 0.9527] & \nobreak\hspace{0pt}0.9120 \\
\bottomrule
\end{tabular}
\end{table}

\begin{table}[!htbp]
\centering
\small
\setlength{\tabcolsep}{3.0pt}
\renewcommand{\arraystretch}{1.12}
\caption{Per-class test performance, shown as F1 / sensitivity / one-versus-rest AUC.}
\label{tab:7}
\begin{tabular}{>{\raggedright\arraybackslash}p{0.282\linewidth}>{\raggedright\arraybackslash}p{0.147\linewidth}>{\raggedright\arraybackslash}p{0.193\linewidth}>{\raggedright\arraybackslash}p{0.130\linewidth}>{\raggedright\arraybackslash}p{0.181\linewidth}}
\toprule
\nobreak\hspace{0pt}\textbf{Model} & \nobreak\hspace{0pt}\textbf{Glioma} & \nobreak\hspace{0pt}\textbf{Meningioma} & \nobreak\hspace{0pt}\textbf{No tumor} & \nobreak\hspace{0pt}\textbf{Pituitary} \\
\midrule
\nobreak\hspace{0pt}EfficientNetV2-M & \nobreak\hspace{0pt}0.914 / 0.843 / 0.939 & \nobreak\hspace{0pt}0.944 / 0.985 / 0.982 & \nobreak\hspace{0pt}0.972 / 1.000 / 0.998 & \nobreak\hspace{0pt}0.989 / 0.995 / 0.999 \\
\nobreak\hspace{0pt}ConvNeXt-V2-Tiny & \nobreak\hspace{0pt}0.899 / 0.823 / 0.964 & \nobreak\hspace{0pt}0.932 / 0.988 / 0.984 & \nobreak\hspace{0pt}0.977 / 0.998 / 0.999 & \nobreak\hspace{0pt}0.989 / 0.993 / 0.998 \\
\nobreak\hspace{0pt}Swin-V2-S & \nobreak\hspace{0pt}0.900 / 0.825 / 0.951 & \nobreak\hspace{0pt}0.947 / 0.965 / 0.996 & \nobreak\hspace{0pt}0.941 / 1.000 / 0.994 & \nobreak\hspace{0pt}0.988 / 0.990 / 0.999 \\
\nobreak\hspace{0pt}ViT-L/16 & \nobreak\hspace{0pt}0.911 / 0.840 / 0.958 & \nobreak\hspace{0pt}0.929 / 0.978 / 0.985 & \nobreak\hspace{0pt}0.985 / 0.993 / 0.999 & \nobreak\hspace{0pt}0.983 / 1.000 / 1.000 \\
\nobreak\hspace{0pt}DINOv2-L & \nobreak\hspace{0pt}0.895 / 0.823 / 0.937 & \nobreak\hspace{0pt}0.925 / 0.993 / 0.983 & \nobreak\hspace{0pt}0.982 / 0.995 / 1.000 & \nobreak\hspace{0pt}0.985 / 0.980 / 1.000 \\
\nobreak\hspace{0pt}MedSigLIP-448 & \nobreak\hspace{0pt}0.906 / 0.828 / 0.949 & \nobreak\hspace{0pt}0.938 / 0.988 / 0.993 & \nobreak\hspace{0pt}0.972 / 1.000 / 0.999 & \nobreak\hspace{0pt}0.995 / 1.000 / 1.000 \\
\nobreak\hspace{0pt}Weighted-average ensemble & \nobreak\hspace{0pt}0.909 / -- / -- & \nobreak\hspace{0pt}0.939 / -- / -- & \nobreak\hspace{0pt}0.973 / -- / -- & \nobreak\hspace{0pt}0.995 / -- / -- \\
\bottomrule
\end{tabular}
\end{table}

\subsection{Performance by Source Dataset}

Accuracy depended more strongly on the source of an image than on the architecture (Table~\ref{tab:8}). Every model reached 98.4--99.7\% on Figshare-derived test images and 98.3--99.4\% on unmatched images, but only 91.4--92.8\% on Br35H-derived images and 77.9--86.6\% on SARTAJ-derived images. The spread between sources (up to 22 percentage points) is an order of magnitude larger than the spread between architectures. The low accuracy on SARTAJ images is consistent with label-quality concerns about that collection raised by the curator of the composite~\cite{r38}.

\begin{table}[!htbp]
\centering
\footnotesize
\setlength{\tabcolsep}{3.0pt}
\renewcommand{\arraystretch}{1.12}
\caption{Test accuracy by source dataset of the image, and on the clean test subset.}
\label{tab:8}
\begin{tabular}{>{\raggedright\arraybackslash}p{0.344\linewidth}>{\raggedright\arraybackslash}p{0.122\linewidth}>{\raggedright\arraybackslash}p{0.126\linewidth}>{\raggedright\arraybackslash}p{0.098\linewidth}>{\raggedright\arraybackslash}p{0.132\linewidth}>{\raggedright\arraybackslash}p{0.097\linewidth}}
\toprule
\nobreak\hspace{0pt}\textbf{Model} & \nobreak\hspace{0pt}\textbf{Figshare (n = 305)} & \nobreak\hspace{0pt}\textbf{SARTAJ (n = 172)} & \nobreak\hspace{0pt}\textbf{Br35H (n = 487)} & \nobreak\hspace{0pt}\textbf{Unmatched (n = 636)} & \nobreak\hspace{0pt}\textbf{Clean subset (n = 773)} \\
\midrule
\nobreak\hspace{0pt}EfficientNetV2-M & \nobreak\hspace{0pt}0.9967 & \nobreak\hspace{0pt}0.8547 & \nobreak\hspace{0pt}0.9220 & \nobreak\hspace{0pt}0.9890 & \nobreak\hspace{0pt}0.9133 \\
\nobreak\hspace{0pt}ConvNeXt-V2-Tiny & \nobreak\hspace{0pt}0.9869 & \nobreak\hspace{0pt}0.8140 & \nobreak\hspace{0pt}0.9261 & \nobreak\hspace{0pt}0.9874 & \nobreak\hspace{0pt}0.9082 \\
\nobreak\hspace{0pt}Swin-V2-S & \nobreak\hspace{0pt}0.9869 & \nobreak\hspace{0pt}0.8081 & \nobreak\hspace{0pt}0.9179 & \nobreak\hspace{0pt}0.9827 & \nobreak\hspace{0pt}0.9030 \\
\nobreak\hspace{0pt}ViT-L/16 & \nobreak\hspace{0pt}0.9934 & \nobreak\hspace{0pt}0.8023 & \nobreak\hspace{0pt}0.9261 & \nobreak\hspace{0pt}0.9937 & \nobreak\hspace{0pt}0.9172 \\
\nobreak\hspace{0pt}DINOv2-L & \nobreak\hspace{0pt}0.9836 & \nobreak\hspace{0pt}0.8023 & \nobreak\hspace{0pt}0.9281 & \nobreak\hspace{0pt}0.9843 & \nobreak\hspace{0pt}0.9082 \\
\nobreak\hspace{0pt}MedSigLIP-448 & \nobreak\hspace{0pt}0.9967 & \nobreak\hspace{0pt}0.8256 & \nobreak\hspace{0pt}0.9220 & \nobreak\hspace{0pt}0.9921 & \nobreak\hspace{0pt}0.9120 \\
\nobreak\hspace{0pt}MedSigLIP (224 px) & \nobreak\hspace{0pt}0.9967 & \nobreak\hspace{0pt}0.8256 & \nobreak\hspace{0pt}0.9240 & \nobreak\hspace{0pt}0.9890 & \nobreak\hspace{0pt}0.9146 \\
\nobreak\hspace{0pt}ConvNeXt-V2-Tiny (448 px) & \nobreak\hspace{0pt}0.9967 & \nobreak\hspace{0pt}0.8663 & \nobreak\hspace{0pt}0.9281 & \nobreak\hspace{0pt}0.9890 & \nobreak\hspace{0pt}0.9237 \\
\nobreak\hspace{0pt}ConvNeXt-V2-Tiny (448 px, matched) & \nobreak\hspace{0pt}0.9902 & \nobreak\hspace{0pt}0.7791 & \nobreak\hspace{0pt}0.9138 & \nobreak\hspace{0pt}0.9843 & \nobreak\hspace{0pt}0.8939 \\
\bottomrule
\end{tabular}
\end{table}

\subsection{Pairwise Statistical Comparison}

Without correction, three of the 15 pairwise McNemar tests had p \textless{} 0.05 (Table~\ref{tab:9}): EfficientNetV2-M versus Swin-V2-S (p = 0.006), Swin-V2-S versus MedSigLIP-448 (p = 0.016) and EfficientNetV2-M versus DINOv2-L (p = 0.041). None survived Holm correction; the smallest adjusted p-value was 0.090. Cochran's Q indicated some heterogeneity across the six models (Q = 14.06, p = 0.015), but no individual pair could be distinguished.

\begin{table}[!htbp]
\centering
\footnotesize
\setlength{\tabcolsep}{3.0pt}
\renewcommand{\arraystretch}{1.12}
\caption{Pairwise comparison of the six primary models on the test set: exact McNemar tests with Holm correction, and paired-bootstrap 95\% confidence intervals for the accuracy difference (first model minus second).}
\label{tab:9}
\begin{tabular}{>{\raggedright\arraybackslash}p{0.245\linewidth}>{\raggedright\arraybackslash}p{0.118\linewidth}>{\raggedright\arraybackslash}p{0.196\linewidth}>{\raggedright\arraybackslash}p{0.126\linewidth}>{\raggedright\arraybackslash}p{0.079\linewidth}>{\raggedright\arraybackslash}p{0.156\linewidth}}
\toprule
\nobreak\hspace{0pt}\textbf{Comparison} & \nobreak\hspace{0pt}\textbf{Accuracy difference (pp)} & \nobreak\hspace{0pt}\textbf{95\% paired-bootstrap CI (pp)} & \nobreak\hspace{0pt}\textbf{Discordant b / c} & \nobreak\hspace{0pt}\textbf{Exact p} & \nobreak\hspace{0pt}\textbf{Holm-adjusted p} \\
\midrule
\nobreak\hspace{0pt}EfficientNetV2-M vs ConvNeXt-V2-Tiny & \nobreak\hspace{0pt}+0.56 & \nobreak\hspace{0pt}[-0.06, +1.25] & \nobreak\hspace{0pt}19 / 10 & \nobreak\hspace{0pt}0.1360 & \nobreak\hspace{0pt}1.0000 \\
\nobreak\hspace{0pt}EfficientNetV2-M vs Swin-V2-S & \nobreak\hspace{0pt}+1.06 & \nobreak\hspace{0pt}[+0.31, +1.81] & \nobreak\hspace{0pt}26 / 9 & \nobreak\hspace{0pt}0.0060 & \nobreak\hspace{0pt}0.0898 \\
\nobreak\hspace{0pt}EfficientNetV2-M vs ViT-L/16 & \nobreak\hspace{0pt}+0.31 & \nobreak\hspace{0pt}[-0.31, +0.94] & \nobreak\hspace{0pt}16 / 11 & \nobreak\hspace{0pt}0.4421 & \nobreak\hspace{0pt}1.0000 \\
\nobreak\hspace{0pt}EfficientNetV2-M vs DINOv2-L & \nobreak\hspace{0pt}+0.81 & \nobreak\hspace{0pt}[+0.06, +1.56] & \nobreak\hspace{0pt}24 / 11 & \nobreak\hspace{0pt}0.0410 & \nobreak\hspace{0pt}0.5325 \\
\nobreak\hspace{0pt}EfficientNetV2-M vs MedSigLIP-448 & \nobreak\hspace{0pt}+0.19 & \nobreak\hspace{0pt}[-0.37, +0.75] & \nobreak\hspace{0pt}12 / 9 & \nobreak\hspace{0pt}0.6636 & \nobreak\hspace{0pt}1.0000 \\
\nobreak\hspace{0pt}ConvNeXt-V2-Tiny vs Swin-V2-S & \nobreak\hspace{0pt}+0.50 & \nobreak\hspace{0pt}[-0.19, +1.25] & \nobreak\hspace{0pt}21 / 13 & \nobreak\hspace{0pt}0.2295 & \nobreak\hspace{0pt}1.0000 \\
\nobreak\hspace{0pt}ConvNeXt-V2-Tiny vs ViT-L/16 & \nobreak\hspace{0pt}-0.25 & \nobreak\hspace{0pt}[-0.87, +0.37] & \nobreak\hspace{0pt}9 / 13 & \nobreak\hspace{0pt}0.5235 & \nobreak\hspace{0pt}1.0000 \\
\nobreak\hspace{0pt}ConvNeXt-V2-Tiny vs DINOv2-L & \nobreak\hspace{0pt}+0.25 & \nobreak\hspace{0pt}[-0.38, +0.87] & \nobreak\hspace{0pt}15 / 11 & \nobreak\hspace{0pt}0.5572 & \nobreak\hspace{0pt}1.0000 \\
\nobreak\hspace{0pt}ConvNeXt-V2-Tiny vs MedSigLIP-448 & \nobreak\hspace{0pt}-0.38 & \nobreak\hspace{0pt}[-0.87, +0.12] & \nobreak\hspace{0pt}5 / 11 & \nobreak\hspace{0pt}0.2101 & \nobreak\hspace{0pt}1.0000 \\
\nobreak\hspace{0pt}Swin-V2-S vs ViT-L/16 & \nobreak\hspace{0pt}-0.75 & \nobreak\hspace{0pt}[-1.50, -0.06] & \nobreak\hspace{0pt}11 / 23 & \nobreak\hspace{0pt}0.0576 & \nobreak\hspace{0pt}0.6914 \\
\nobreak\hspace{0pt}Swin-V2-S vs DINOv2-L & \nobreak\hspace{0pt}-0.25 & \nobreak\hspace{0pt}[-1.06, +0.56] & \nobreak\hspace{0pt}21 / 25 & \nobreak\hspace{0pt}0.6587 & \nobreak\hspace{0pt}1.0000 \\
\nobreak\hspace{0pt}Swin-V2-S vs MedSigLIP-448 & \nobreak\hspace{0pt}-0.88 & \nobreak\hspace{0pt}[-1.56, -0.25] & \nobreak\hspace{0pt}8 / 22 & \nobreak\hspace{0pt}0.0161 & \nobreak\hspace{0pt}0.2257 \\
\nobreak\hspace{0pt}ViT-L/16 vs DINOv2-L & \nobreak\hspace{0pt}+0.50 & \nobreak\hspace{0pt}[-0.19, +1.19] & \nobreak\hspace{0pt}19 / 11 & \nobreak\hspace{0pt}0.2005 & \nobreak\hspace{0pt}1.0000 \\
\nobreak\hspace{0pt}ViT-L/16 vs MedSigLIP-448 & \nobreak\hspace{0pt}-0.12 & \nobreak\hspace{0pt}[-0.69, +0.44] & \nobreak\hspace{0pt}9 / 11 & \nobreak\hspace{0pt}0.8238 & \nobreak\hspace{0pt}1.0000 \\
\nobreak\hspace{0pt}DINOv2-L vs MedSigLIP-448 & \nobreak\hspace{0pt}-0.62 & \nobreak\hspace{0pt}[-1.31, +0.12] & \nobreak\hspace{0pt}11 / 21 & \nobreak\hspace{0pt}0.1102 & \nobreak\hspace{0pt}1.0000 \\
\bottomrule
\end{tabular}
\end{table}

\subsection{Calibration}

\begin{figure}[!htbp]
\centering
\includegraphics[width=0.88\linewidth,height=0.82\textheight,keepaspectratio]{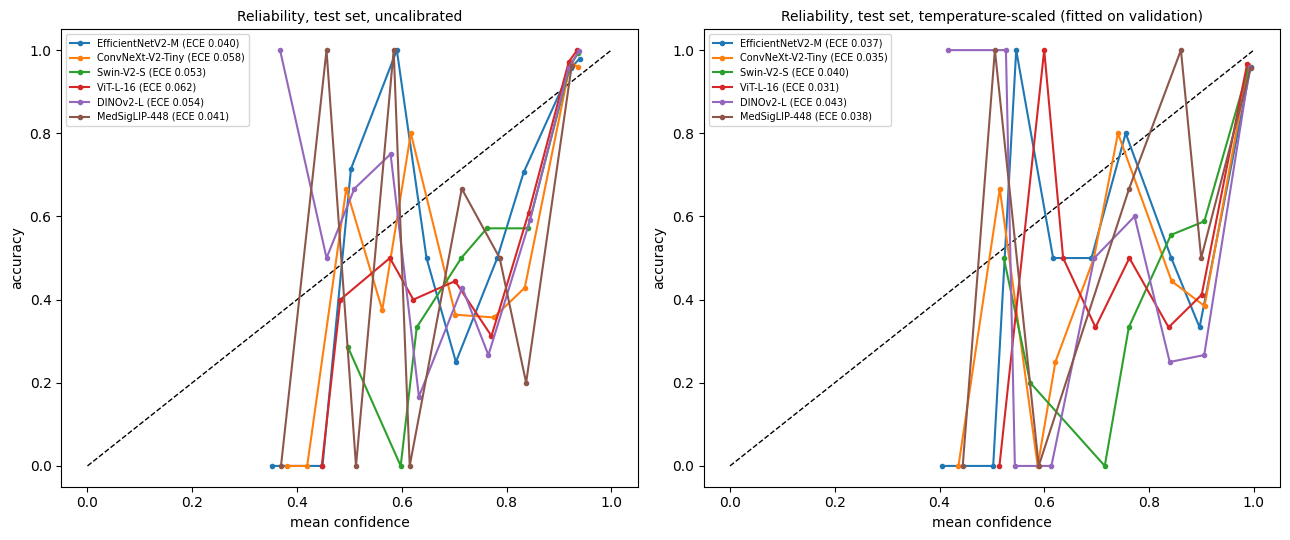}
\caption{Reliability diagrams on the test set before (left) and after (right) temperature scaling fitted on the validation set.}
\label{fig:7}
\end{figure}

All fitted temperatures were below one (T = 0.58--0.68; Table~\ref{tab:10}). This indicates that the label-smoothed models were under-confident on validation data, so temperature scaling sharpened their probabilities. Scaling reduced the equal-width ECE for eight of the nine runs; for example, it fell from 0.062 to 0.031 for ViT-L/16 and from 0.058 to 0.035 for ConvNeXt-V2-Tiny. After scaling, the ECE of all runs lay between 0.030 and 0.043, with overlapping intervals. Scaling slightly increased, or left unchanged, the Brier score and NLL on the test set, however. The temperatures were fitted on validation data on which the models were 98--99\% accurate, and they over-sharpened predictions on the harder test set. This is a second sign of the shift between the validation and test partitions. The reliability diagrams are shown in Figure~\ref{fig:7}.

\begin{table}[!htbp]
\centering
\scriptsize
\setlength{\tabcolsep}{2.0pt}
\renewcommand{\arraystretch}{1.12}
\caption{Calibration on the test set before and after temperature scaling fitted on the validation set (95\% bootstrap confidence intervals in brackets).}
\label{tab:10}
\begin{tabular}{>{\raggedright\arraybackslash}p{0.168\linewidth}>{\raggedright\arraybackslash}p{0.112\linewidth}>{\raggedright\arraybackslash}p{0.119\linewidth}>{\raggedright\arraybackslash}p{0.174\linewidth}>{\raggedright\arraybackslash}p{0.105\linewidth}>{\raggedright\arraybackslash}p{0.105\linewidth}>{\raggedright\arraybackslash}p{0.072\linewidth}>{\raggedright\arraybackslash}p{0.072\linewidth}}
\toprule
\nobreak\hspace{0pt}\textbf{Model} & \nobreak\hspace{0pt}\textbf{Temperature T} & \nobreak\hspace{0pt}\textbf{ECE (15 bins), uncalibrated} & \nobreak\hspace{0pt}\textbf{ECE (15 bins), temperature-scaled} & \nobreak\hspace{0pt}\textbf{Equal-mass ECE, scaled} & \nobreak\hspace{0pt}\textbf{Class-wise ECE, scaled} & \nobreak\hspace{0pt}\textbf{Brier (uncal. $\rightarrow$ scaled)} & \nobreak\hspace{0pt}\textbf{NLL (uncal. $\rightarrow$ scaled)} \\
\midrule
\nobreak\hspace{0pt}EfficientNetV2-M & \nobreak\hspace{0pt}0.578 & \nobreak\hspace{0pt}0.0403 [0.0312, 0.0499] & \nobreak\hspace{0pt}0.0372 [0.0289, 0.0487] & \nobreak\hspace{0pt}0.0347 [0.0261, 0.0460] & \nobreak\hspace{0pt}0.0215 [0.0173, 0.0269] & \nobreak\hspace{0pt}0.0848 $\rightarrow$ 0.0865 & \nobreak\hspace{0pt}0.2278 $\rightarrow$ 0.2624 \\
\nobreak\hspace{0pt}ConvNeXt-V2-Tiny & \nobreak\hspace{0pt}0.626 & \nobreak\hspace{0pt}0.0579 [0.0498, 0.0676] & \nobreak\hspace{0pt}0.0349 [0.0265, 0.0456] & \nobreak\hspace{0pt}0.0370 [0.0305, 0.0492] & \nobreak\hspace{0pt}0.0221 [0.0177, 0.0276] & \nobreak\hspace{0pt}0.0888 $\rightarrow$ 0.0915 & \nobreak\hspace{0pt}0.2331 $\rightarrow$ 0.2442 \\
\nobreak\hspace{0pt}Swin-V2-S & \nobreak\hspace{0pt}0.622 & \nobreak\hspace{0pt}0.0526 [0.0422, 0.0622] & \nobreak\hspace{0pt}0.0402 [0.0309, 0.0517] & \nobreak\hspace{0pt}0.0414 [0.0346, 0.0546] & \nobreak\hspace{0pt}0.0243 [0.0200, 0.0302] & \nobreak\hspace{0pt}0.0982 $\rightarrow$ 0.1020 & \nobreak\hspace{0pt}0.2525 $\rightarrow$ 0.2769 \\
\nobreak\hspace{0pt}ConvNeXt-V2-Tiny (448 px) & \nobreak\hspace{0pt}0.594 & \nobreak\hspace{0pt}0.0589 [0.0508, 0.0684] & \nobreak\hspace{0pt}0.0302 [0.0222, 0.0407] & \nobreak\hspace{0pt}0.0326 [0.0267, 0.0449] & \nobreak\hspace{0pt}0.0190 [0.0149, 0.0247] & \nobreak\hspace{0pt}0.0775 $\rightarrow$ 0.0775 & \nobreak\hspace{0pt}0.2194 $\rightarrow$ 0.2316 \\
\nobreak\hspace{0pt}ConvNeXt-V2-Tiny (448 px, matched) & \nobreak\hspace{0pt}0.681 & \nobreak\hspace{0pt}0.0345 [0.0235, 0.0451] & \nobreak\hspace{0pt}0.0397 [0.0308, 0.0507] & \nobreak\hspace{0pt}0.0434 [0.0368, 0.0564] & \nobreak\hspace{0pt}0.0274 [0.0228, 0.0330] & \nobreak\hspace{0pt}0.1088 $\rightarrow$ 0.1111 & \nobreak\hspace{0pt}0.2806 $\rightarrow$ 0.3035 \\
\nobreak\hspace{0pt}ViT-L/16 & \nobreak\hspace{0pt}0.640 & \nobreak\hspace{0pt}0.0621 [0.0547, 0.0723] & \nobreak\hspace{0pt}0.0307 [0.0221, 0.0410] & \nobreak\hspace{0pt}0.0369 [0.0304, 0.0488] & \nobreak\hspace{0pt}0.0204 [0.0164, 0.0258] & \nobreak\hspace{0pt}0.0860 $\rightarrow$ 0.0877 & \nobreak\hspace{0pt}0.2321 $\rightarrow$ 0.2343 \\
\nobreak\hspace{0pt}DINOv2-L & \nobreak\hspace{0pt}0.590 & \nobreak\hspace{0pt}0.0544 [0.0449, 0.0657] & \nobreak\hspace{0pt}0.0431 [0.0341, 0.0538] & \nobreak\hspace{0pt}0.0425 [0.0345, 0.0544] & \nobreak\hspace{0pt}0.0234 [0.0193, 0.0291] & \nobreak\hspace{0pt}0.0954 $\rightarrow$ 0.0996 & \nobreak\hspace{0pt}0.2450 $\rightarrow$ 0.2757 \\
\nobreak\hspace{0pt}MedSigLIP (224 px) & \nobreak\hspace{0pt}0.640 & \nobreak\hspace{0pt}0.0508 [0.0420, 0.0617] & \nobreak\hspace{0pt}0.0338 [0.0244, 0.0438] & \nobreak\hspace{0pt}0.0388 [0.0317, 0.0497] & \nobreak\hspace{0pt}0.0212 [0.0166, 0.0264] & \nobreak\hspace{0pt}0.0868 $\rightarrow$ 0.0885 & \nobreak\hspace{0pt}0.2330 $\rightarrow$ 0.2433 \\
\nobreak\hspace{0pt}MedSigLIP-448 & \nobreak\hspace{0pt}0.578 & \nobreak\hspace{0pt}0.0412 [0.0314, 0.0504] & \nobreak\hspace{0pt}0.0385 [0.0294, 0.0491] & \nobreak\hspace{0pt}0.0422 [0.0343, 0.0534] & \nobreak\hspace{0pt}0.0222 [0.0178, 0.0273] & \nobreak\hspace{0pt}0.0865 $\rightarrow$ 0.0881 & \nobreak\hspace{0pt}0.2344 $\rightarrow$ 0.2702 \\
\bottomrule
\end{tabular}
\end{table}

\subsection{Ensembles and Ensemble Diversity}

\textbf{Validation-fitted ensembles:} The validation-fitted weights concentrated on MedSigLIP-448 (0.713) and DINOv2-L (0.226), with ConvNeXt-V2-Tiny receiving 0.046, EfficientNetV2-M 0.015, and Swin-V2-S and ViT-L/16 zero. No validation-fitted ensemble was significantly more accurate than the best single model (Table~\ref{tab:11}). Accuracies ranged from 95.44\% (simple average and majority vote) to 95.75\% (stacking; 9 versus 6 discordant images, p = 0.61), and 97.8--98.4\% of ensemble predictions were identical to those of the best single model. The simple average had the best calibration (ECE 0.019) and the highest AUC (0.993) of all methods.

\begin{figure}[!htbp]
\centering
\includegraphics[width=0.88\linewidth,height=0.82\textheight,keepaspectratio]{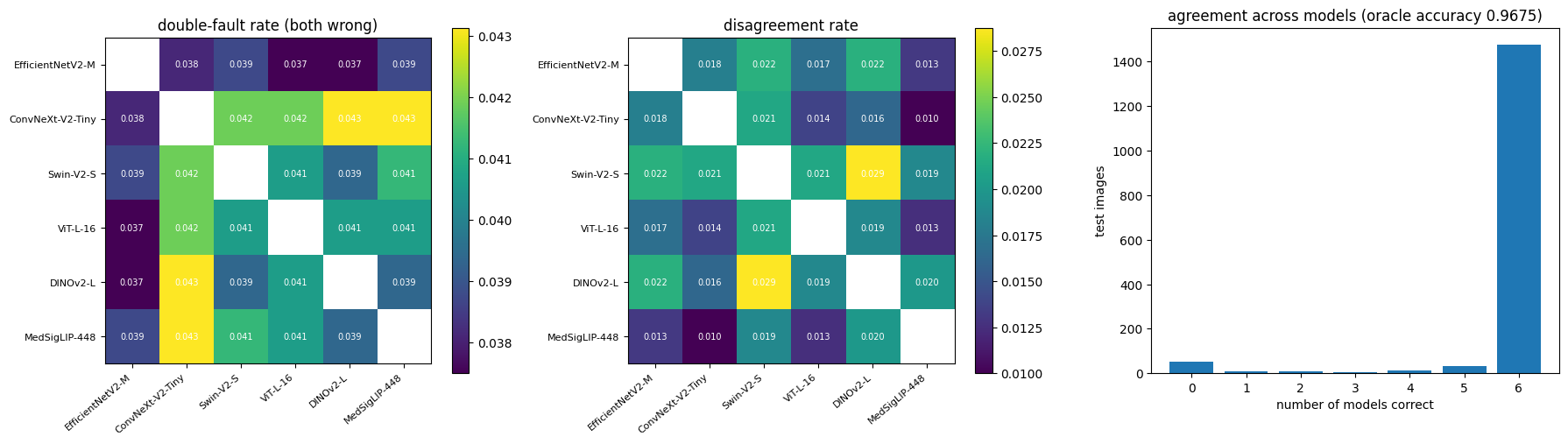}
\caption{Ensemble diversity on the test set: pairwise double-fault rates (left), pairwise disagreement rates (centre) and the number of models that classify each image correctly (right; oracle accuracy 0.9675).}
\label{fig:8}
\end{figure}

\textbf{Ensemble diversity:} The architectures made largely the same errors (Figure~\ref{fig:8}). Pairwise prediction agreement was 95.4--97.8\% (kappa 0.939--0.970). The oracle accuracy was 96.75\%: 52 test images (3.25\%) were misclassified by all six models, and most of the remainder were classified correctly by all six. Ensembles therefore have little complementary information to exploit.

\textbf{Cross-fitted sensitivity analysis:} In the cross-fitted analysis on the test set (test-set groups: 1,210, largest 10; folds of 314--325 images), the weighted average reached 95.25\%. Stacking reached 97.50\% [96.69, 98.19], significantly above the best single model (42 versus 11 discordant images, p \textless{} 0.0001). Because no image was predicted with parameters fitted on itself, its duplicates or its patient, this gain is not leakage in the narrow sense. It shows instead that a stacker fitted on test-distribution data can adapt to how the Testing folder differs from the Training folder. A stacker whose meta-learner is trained on part of the test set is therefore optimistically biased. When the stacker is fitted on validation data, as a deployed model would be, it confers no significant benefit.

\begin{table}[!htbp]
\centering
\scriptsize
\setlength{\tabcolsep}{3.0pt}
\renewcommand{\arraystretch}{1.12}
\caption{Ensembles evaluated on the same 1,600 test images. Validation-fitted rows are the primary analysis; cross-fitted rows are a sensitivity analysis. Comparisons are with the best single model on test (EfficientNetV2-M); the last column is the share of predictions identical to that model.}
\label{tab:11}
\begin{tabular}{>{\raggedright\arraybackslash}p{0.175\linewidth}>{\raggedright\arraybackslash}p{0.134\linewidth}>{\raggedright\arraybackslash}p{0.094\linewidth}>{\raggedright\arraybackslash}p{0.092\linewidth}>{\raggedright\arraybackslash}p{0.064\linewidth}>{\raggedright\arraybackslash}p{0.087\linewidth}>{\raggedright\arraybackslash}p{0.074\linewidth}>{\raggedright\arraybackslash}p{0.065\linewidth}>{\raggedright\arraybackslash}p{0.095\linewidth}}
\toprule
\nobreak\hspace{0pt}\textbf{Method} & \nobreak\hspace{0pt}\textbf{Parameters fitted on} & \nobreak\hspace{0pt}\textbf{Accuracy [95\% CI]} & \nobreak\hspace{0pt}\textbf{Macro-F1} & \nobreak\hspace{0pt}\textbf{Macro AUC} & \nobreak\hspace{0pt}\textbf{Log-loss} & \nobreak\hspace{0pt}\textbf{vs best single, b / c} & \nobreak\hspace{0pt}\textbf{Exact p} & \nobreak\hspace{0pt}\textbf{Identical to best single} \\
\midrule
\nobreak\hspace{0pt}Simple average & \nobreak\hspace{0pt}none & \nobreak\hspace{0pt}0.9544 [0.9437, 0.9644] & \nobreak\hspace{0pt}0.9534 & \nobreak\hspace{0pt}0.9929 & \nobreak\hspace{0pt}0.2072 & \nobreak\hspace{0pt}8 / 10 & \nobreak\hspace{0pt}0.8145 & \nobreak\hspace{0pt}0.984 \\
\nobreak\hspace{0pt}Weighted average & \nobreak\hspace{0pt}validation & \nobreak\hspace{0pt}0.9550 [0.9450, 0.9644] & \nobreak\hspace{0pt}0.9541 & \nobreak\hspace{0pt}0.9892 & \nobreak\hspace{0pt}0.2344 & \nobreak\hspace{0pt}9 / 10 & \nobreak\hspace{0pt}1.0000 & \nobreak\hspace{0pt}0.978 \\
\nobreak\hspace{0pt}Majority vote & \nobreak\hspace{0pt}none & \nobreak\hspace{0pt}0.9544 [0.9437, 0.9644] & \nobreak\hspace{0pt}0.9534 & \nobreak\hspace{0pt}0.9929 & \nobreak\hspace{0pt}0.4219 & \nobreak\hspace{0pt}8 / 10 & \nobreak\hspace{0pt}0.8145 & \nobreak\hspace{0pt}0.984 \\
\nobreak\hspace{0pt}Stacking & \nobreak\hspace{0pt}validation & \nobreak\hspace{0pt}0.9575 [0.9475, 0.9669] & \nobreak\hspace{0pt}0.9569 & \nobreak\hspace{0pt}0.9893 & \nobreak\hspace{0pt}0.2364 & \nobreak\hspace{0pt}9 / 6 & \nobreak\hspace{0pt}0.6072 & \nobreak\hspace{0pt}0.984 \\
\nobreak\hspace{0pt}Best single on validation (MedSigLIP-448) & \nobreak\hspace{0pt}validation & \nobreak\hspace{0pt}0.9538 [0.9431, 0.9637] & \nobreak\hspace{0pt}0.9527 & \nobreak\hspace{0pt}0.9853 & \nobreak\hspace{0pt}0.2702 & \nobreak\hspace{0pt}9 / 12 & \nobreak\hspace{0pt}0.6636 & \nobreak\hspace{0pt}0.976 \\
\nobreak\hspace{0pt}Weighted average & \nobreak\hspace{0pt}test, 5-fold cross-fitted (sensitivity) & \nobreak\hspace{0pt}0.9525 [0.9419, 0.9625] & \nobreak\hspace{0pt}0.9516 & \nobreak\hspace{0pt}0.9916 & \nobreak\hspace{0pt}0.2077 & \nobreak\hspace{0pt}7 / 12 & \nobreak\hspace{0pt}0.3593 & \nobreak\hspace{0pt}-- \\
\nobreak\hspace{0pt}Stacking & \nobreak\hspace{0pt}test, 5-fold cross-fitted (sensitivity) & \nobreak\hspace{0pt}0.9750 [0.9669, 0.9819] & \nobreak\hspace{0pt}0.9749 & \nobreak\hspace{0pt}0.9972 & \nobreak\hspace{0pt}0.0842 & \nobreak\hspace{0pt}42 / 11 & \nobreak\hspace{0pt}\textless{} 0.0001 & \nobreak\hspace{0pt}-- \\
\bottomrule
\end{tabular}
\end{table}

\subsection{Thresholds and the Multiclass Decision Rule}

\begin{figure}[!htbp]
\centering
\includegraphics[width=0.88\linewidth,height=0.82\textheight,keepaspectratio]{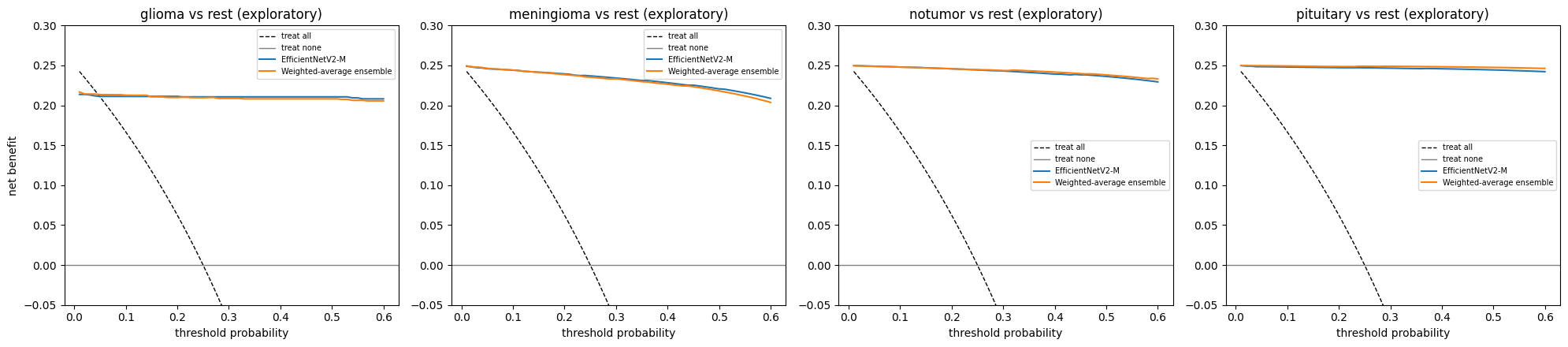}
\caption{Exploratory one-versus-rest decision curves on the test set for the best single model and the weighted-average ensemble.}
\label{fig:9}
\end{figure}

Applying validation-fitted one-versus-rest thresholds changed accuracy by $-$0.94 to +0.37 percentage points relative to argmax (Table~\ref{tab:12}). The scaled-argmax rule helped only MedSigLIP-448 (95.75\% versus 95.38\%), and the one-flag rule was within 0.7 percentage points of argmax for every model. Several validation thresholds were extreme (for example 0.006 or 0.99; Table~\ref{tab:13}), because the validation classes are almost perfectly separated. This makes threshold selection unstable. The exploratory one-versus-rest decision curves are shown in Figure~\ref{fig:9}.

\begin{table}[!htbp]
\centering
\small
\setlength{\tabcolsep}{3.0pt}
\renewcommand{\arraystretch}{1.12}
\caption{Test accuracy under the three multiclass decision rules, with thresholds fitted on the validation set.}
\label{tab:12}
\begin{tabular}{>{\raggedright\arraybackslash}p{0.434\linewidth}>{\raggedright\arraybackslash}p{0.176\linewidth}>{\raggedright\arraybackslash}p{0.176\linewidth}>{\raggedright\arraybackslash}p{0.159\linewidth}}
\toprule
\nobreak\hspace{0pt}\textbf{Model} & \nobreak\hspace{0pt}\textbf{Argmax (no threshold)} & \nobreak\hspace{0pt}\textbf{Scaled argmax, argmax(p\_k / t\_k)} & \nobreak\hspace{0pt}\textbf{One-flag rule} \\
\midrule
\nobreak\hspace{0pt}EfficientNetV2-M & \nobreak\hspace{0pt}0.9556 & \nobreak\hspace{0pt}0.9519 & \nobreak\hspace{0pt}0.9512 \\
\nobreak\hspace{0pt}ConvNeXt-V2-Tiny & \nobreak\hspace{0pt}0.9500 & \nobreak\hspace{0pt}0.9450 & \nobreak\hspace{0pt}0.9488 \\
\nobreak\hspace{0pt}Swin-V2-S & \nobreak\hspace{0pt}0.9450 & \nobreak\hspace{0pt}0.9438 & \nobreak\hspace{0pt}0.9450 \\
\nobreak\hspace{0pt}ViT-L/16 & \nobreak\hspace{0pt}0.9525 & \nobreak\hspace{0pt}0.9500 & \nobreak\hspace{0pt}0.9519 \\
\nobreak\hspace{0pt}DINOv2-L & \nobreak\hspace{0pt}0.9475 & \nobreak\hspace{0pt}0.9438 & \nobreak\hspace{0pt}0.9469 \\
\nobreak\hspace{0pt}MedSigLIP-448 & \nobreak\hspace{0pt}0.9538 & \nobreak\hspace{0pt}0.9575 & \nobreak\hspace{0pt}0.9569 \\
\nobreak\hspace{0pt}Weighted-average ensemble & \nobreak\hspace{0pt}0.9550 & \nobreak\hspace{0pt}0.9456 & \nobreak\hspace{0pt}0.9481 \\
\bottomrule
\end{tabular}
\end{table}

\begin{table}[!htbp]
\centering
\small
\setlength{\tabcolsep}{3.0pt}
\renewcommand{\arraystretch}{1.12}
\caption{Validation-fitted one-versus-rest Youden thresholds, with the resulting test sensitivity / specificity in brackets.}
\label{tab:13}
\begin{tabular}{>{\raggedright\arraybackslash}p{0.256\linewidth}>{\raggedright\arraybackslash}p{0.150\linewidth}>{\raggedright\arraybackslash}p{0.196\linewidth}>{\raggedright\arraybackslash}p{0.144\linewidth}>{\raggedright\arraybackslash}p{0.185\linewidth}}
\toprule
\nobreak\hspace{0pt}\textbf{Model} & \nobreak\hspace{0pt}\textbf{Glioma} & \nobreak\hspace{0pt}\textbf{Meningioma} & \nobreak\hspace{0pt}\textbf{No tumor} & \nobreak\hspace{0pt}\textbf{Pituitary} \\
\midrule
\nobreak\hspace{0pt}EfficientNetV2-M & \nobreak\hspace{0pt}0.792 (0.825 / 1.000) & \nobreak\hspace{0pt}0.117 (0.990 / 0.959) & \nobreak\hspace{0pt}0.367 (1.000 / 0.978) & \nobreak\hspace{0pt}0.706 (0.995 / 0.995) \\
\nobreak\hspace{0pt}ConvNeXt-V2-Tiny & \nobreak\hspace{0pt}0.750 (0.815 / 0.998) & \nobreak\hspace{0pt}0.026 (0.990 / 0.938) & \nobreak\hspace{0pt}0.123 (0.998 / 0.982) & \nobreak\hspace{0pt}0.024 (0.998 / 0.992) \\
\nobreak\hspace{0pt}Swin-V2-S & \nobreak\hspace{0pt}0.084 (0.825 / 0.996) & \nobreak\hspace{0pt}0.006 (0.990 / 0.948) & \nobreak\hspace{0pt}0.265 (1.000 / 0.956) & \nobreak\hspace{0pt}0.931 (0.978 / 0.997) \\
\nobreak\hspace{0pt}ViT-L/16 & \nobreak\hspace{0pt}0.827 (0.838 / 0.998) & \nobreak\hspace{0pt}0.024 (0.990 / 0.943) & \nobreak\hspace{0pt}0.007 (0.998 / 0.965) & \nobreak\hspace{0pt}0.799 (1.000 / 0.990) \\
\nobreak\hspace{0pt}DINOv2-L & \nobreak\hspace{0pt}0.075 (0.833 / 0.991) & \nobreak\hspace{0pt}0.102 (0.993 / 0.942) & \nobreak\hspace{0pt}0.037 (0.995 / 0.983) & \nobreak\hspace{0pt}0.716 (0.973 / 0.997) \\
\nobreak\hspace{0pt}MedSigLIP-448 & \nobreak\hspace{0pt}0.021 (0.843 / 1.000) & \nobreak\hspace{0pt}0.844 (0.988 / 0.962) & \nobreak\hspace{0pt}0.965 (1.000 / 0.985) & \nobreak\hspace{0pt}0.993 (0.998 / 0.999) \\
\bottomrule
\end{tabular}
\end{table}

\subsection{Resolution- and Preprocessing-Controlled Ablation}

The ablations separate the effects of resolution, preprocessing and architecture (Table~\ref{tab:14}). MedSigLIP performed the same at 448 and 224 pixels (95.38\% versus 95.31\%; p = 1.0), so its performance is not attributable to its higher input resolution. Resolution did help the CNN: the accuracy of ConvNeXt-V2-Tiny rose from 95.00\% to 95.88\% at 448 pixels (p = 0.003). At equal resolution, MedSigLIP did not differ from ConvNeXt-V2-Tiny, either at 224 pixels (p = 0.47) or at 448 pixels (p = 0.12). When ConvNeXt-V2-Tiny was trained at 448 pixels with MedSigLIP's batch size and normalisation, its accuracy fell to 94.19\%; MedSigLIP-448 was then 1.19 percentage points better (p = 0.0003), but the matched ConvNeXt was itself 1.69 points worse than the same network with its own settings (p \textless{} 0.0001).

Batch size and normalisation therefore change accuracy as much as the choice between a medical foundation model and a CNN. The internal test set provides no evidence that MedSigLIP's medical pretraining confers an advantage.

\begin{table}[!htbp]
\centering
\footnotesize
\setlength{\tabcolsep}{3.0pt}
\renewcommand{\arraystretch}{1.12}
\caption{Resolution- and preprocessing-controlled ablation on the test set (exact McNemar tests).}
\label{tab:14}
\begin{tabular}{>{\raggedright\arraybackslash}p{0.313\linewidth}>{\raggedright\arraybackslash}p{0.109\linewidth}>{\raggedright\arraybackslash}p{0.109\linewidth}>{\raggedright\arraybackslash}p{0.166\linewidth}>{\raggedright\arraybackslash}p{0.133\linewidth}>{\raggedright\arraybackslash}p{0.089\linewidth}}
\toprule
\nobreak\hspace{0pt}\textbf{Comparison (only the stated factor differs)} & \nobreak\hspace{0pt}\textbf{Accuracy A} & \nobreak\hspace{0pt}\textbf{Accuracy B} & \nobreak\hspace{0pt}\textbf{Difference A $-$ B (pp) [95\% CI]} & \nobreak\hspace{0pt}\textbf{Discordant b / c} & \nobreak\hspace{0pt}\textbf{Exact p} \\
\midrule
\nobreak\hspace{0pt}MedSigLIP-448 vs MedSigLIP (224 px) & \nobreak\hspace{0pt}0.9538 & \nobreak\hspace{0pt}0.9531 & \nobreak\hspace{0pt}+0.06 [-0.56, +0.63] & \nobreak\hspace{0pt}13 / 12 & \nobreak\hspace{0pt}1.0000 \\
\nobreak\hspace{0pt}ConvNeXt-V2-Tiny (448 px) vs ConvNeXt-V2-Tiny & \nobreak\hspace{0pt}0.9588 & \nobreak\hspace{0pt}0.9500 & \nobreak\hspace{0pt}+0.88 [+0.37, +1.44] & \nobreak\hspace{0pt}17 / 3 & \nobreak\hspace{0pt}0.0026 \\
\nobreak\hspace{0pt}MedSigLIP (224 px) vs ConvNeXt-V2-Tiny & \nobreak\hspace{0pt}0.9531 & \nobreak\hspace{0pt}0.9500 & \nobreak\hspace{0pt}+0.31 [-0.37, +1.00] & \nobreak\hspace{0pt}18 / 13 & \nobreak\hspace{0pt}0.4731 \\
\nobreak\hspace{0pt}MedSigLIP-448 vs ConvNeXt-V2-Tiny (448 px) & \nobreak\hspace{0pt}0.9538 & \nobreak\hspace{0pt}0.9588 & \nobreak\hspace{0pt}-0.50 [-1.06, +0.00] & \nobreak\hspace{0pt}6 / 14 & \nobreak\hspace{0pt}0.1153 \\
\nobreak\hspace{0pt}MedSigLIP-448 vs ConvNeXt-V2-Tiny (448 px, matched) & \nobreak\hspace{0pt}0.9538 & \nobreak\hspace{0pt}0.9419 & \nobreak\hspace{0pt}+1.19 [+0.56, +1.81] & \nobreak\hspace{0pt}23 / 4 & \nobreak\hspace{0pt}0.0003 \\
\nobreak\hspace{0pt}ConvNeXt-V2-Tiny (448 px, matched) vs ConvNeXt-V2-Tiny (448 px) & \nobreak\hspace{0pt}0.9419 & \nobreak\hspace{0pt}0.9588 & \nobreak\hspace{0pt}-1.69 [-2.50, -1.00] & \nobreak\hspace{0pt}6 / 33 & \nobreak\hspace{0pt}\textless{} 0.0001 \\
\bottomrule
\end{tabular}
\end{table}

\subsection{Shortcut Tests and Feature Space}

\begin{figure}[!htbp]
\centering
\includegraphics[width=0.75\linewidth,height=0.82\textheight,keepaspectratio]{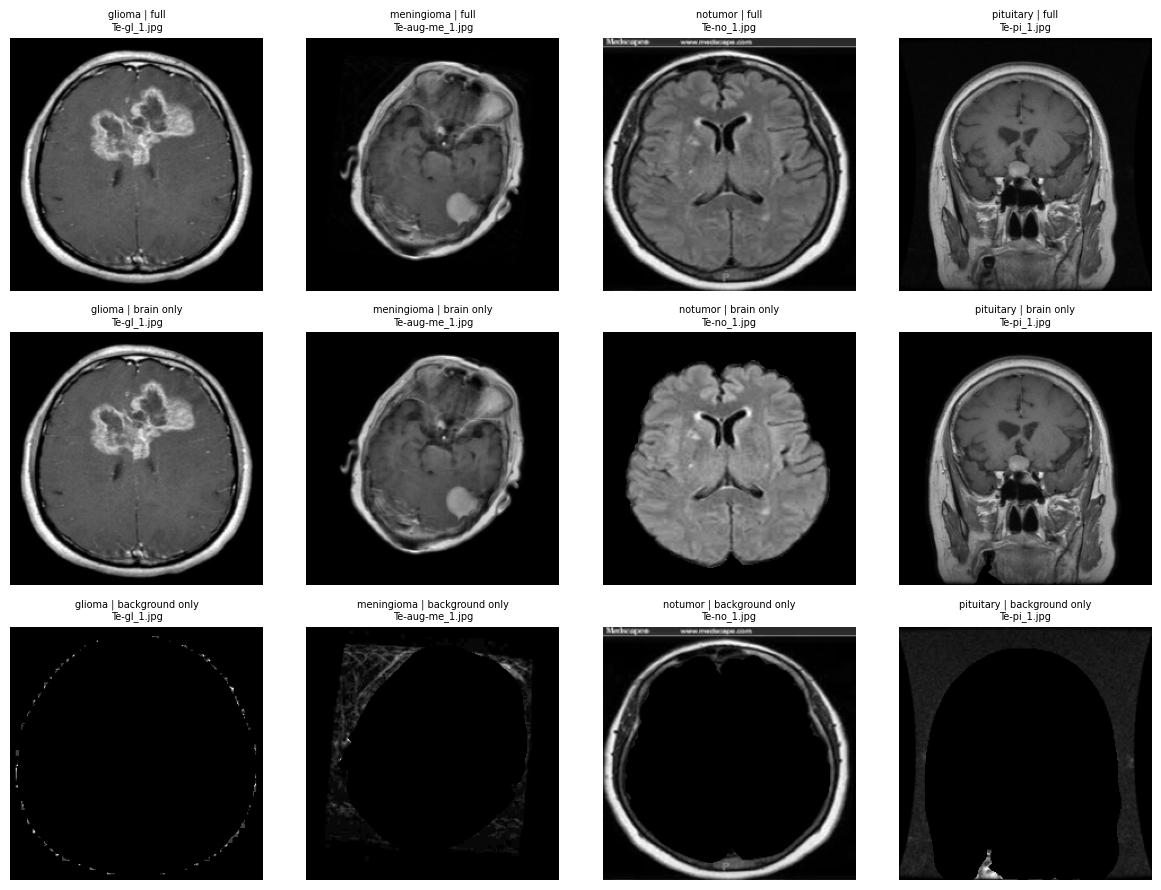}
\caption{The brain masks used for the shortcut tests: full image (top), brain only (middle) and background only (bottom), for one test image per class.}
\label{fig:10}
\end{figure}

\textbf{Masked inputs:} The fine-tuned models relied mainly on the brain region (Table~\ref{tab:15}). With the background removed (Figure~\ref{fig:10}), accuracy fell by only 0.4--1.3 percentage points (significant for eight of nine runs). With the brain removed, accuracy fell to 25.8--40.3\%, close to chance (25\%).

\textbf{Linear probes and source prediction:} The background nevertheless carries strong class information. Linear probes on frozen pretrained features (Table~\ref{tab:16}) reached 84.1--88.5\% accuracy on background-only images, compared with 91.5--93.8\% on full images. The source dataset of a tumor image could also be predicted from the fine-tuned embeddings with 82.3--86.4\% accuracy, compared with a majority-class baseline of 71.4\%; MedSigLIP-448 gave the highest figure. Source-specific information is therefore present in the learned representations. Because class and source are confounded in this collection (Table~\ref{tab:3}), part of the class separability can be explained by source rather than by tumor appearance.

\begin{figure}[!htbp]
\centering
\includegraphics[width=0.82\linewidth,height=0.82\textheight,keepaspectratio]{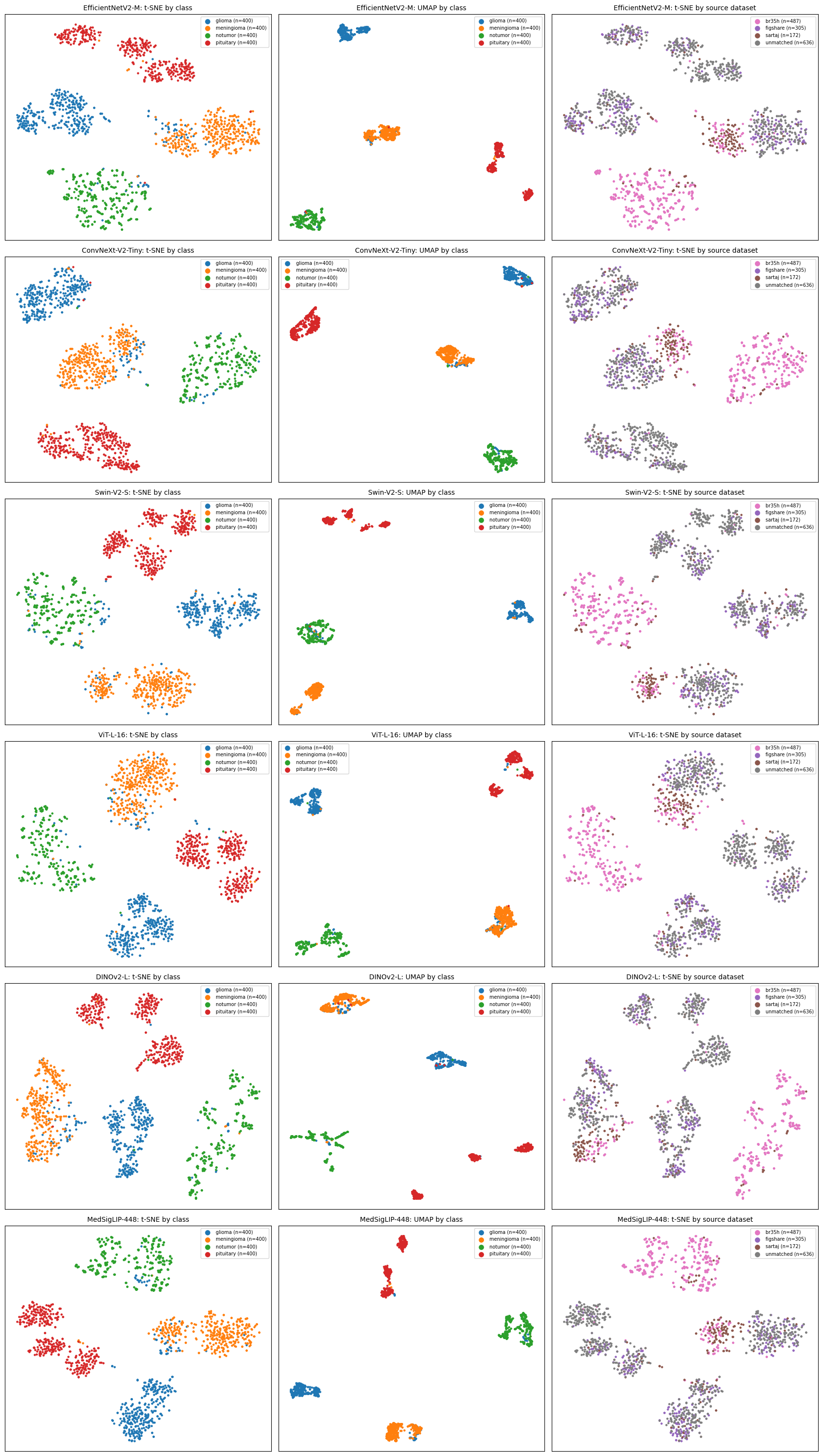}
\caption{t-SNE (left), UMAP (centre) and source-coloured t-SNE (right) of the fine-tuned embeddings of all 1,600 test images (400 per class) for each primary model. Coordinates, labels, sample identifiers and parameters are available from the corresponding author on reasonable request.}
\label{fig:11}
\end{figure}

\textbf{Embedding projections:} The corrected t-SNE and UMAP projections (Figure~\ref{fig:11}) show all four classes, each with 400 points. Colouring by source shows sub-clusters that follow the source dataset within the tumor classes, and a no-tumor cluster composed almost entirely of Br35H images.

\begin{table}[!htbp]
\centering
\footnotesize
\setlength{\tabcolsep}{3.0pt}
\renewcommand{\arraystretch}{1.12}
\caption{Fine-tuned models evaluated on masked test images, and source prediction from their embeddings. $^{\dagger}$ Five-fold cross-validated accuracy for tumor images of Figshare or SARTAJ origin in the validation and test sets (n = 860); majority-class baseline 0.714.}
\label{tab:15}
\begin{tabular}{>{\raggedright\arraybackslash}p{0.288\linewidth}>{\raggedright\arraybackslash}p{0.087\linewidth}>{\raggedright\arraybackslash}p{0.087\linewidth}>{\raggedright\arraybackslash}p{0.088\linewidth}>{\raggedright\arraybackslash}p{0.087\linewidth}>{\raggedright\arraybackslash}p{0.133\linewidth}>{\raggedright\arraybackslash}p{0.137\linewidth}}
\toprule
\nobreak\hspace{0pt}\textbf{Fine-tuned model} & \nobreak\hspace{0pt}\textbf{Full image} & \nobreak\hspace{0pt}\textbf{Brain only} & \nobreak\hspace{0pt}\textbf{Change (pp)} & \nobreak\hspace{0pt}\textbf{p (brain vs full)} & \nobreak\hspace{0pt}\textbf{Background only} & \nobreak\hspace{0pt}\textbf{Source prediction (Figshare vs SARTAJ)$^{\dagger}$} \\
\midrule
\nobreak\hspace{0pt}EfficientNetV2-M & \nobreak\hspace{0pt}0.9556 & \nobreak\hspace{0pt}0.9444 & \nobreak\hspace{0pt}-1.12 & \nobreak\hspace{0pt}0.0003 & \nobreak\hspace{0pt}0.2581 & \nobreak\hspace{0pt}0.8302 \\
\nobreak\hspace{0pt}ConvNeXt-V2-Tiny & \nobreak\hspace{0pt}0.9500 & \nobreak\hspace{0pt}0.9406 & \nobreak\hspace{0pt}-0.94 & \nobreak\hspace{0pt}0.0015 & \nobreak\hspace{0pt}0.2769 & \nobreak\hspace{0pt}0.8477 \\
\nobreak\hspace{0pt}Swin-V2-S & \nobreak\hspace{0pt}0.9450 & \nobreak\hspace{0pt}0.9319 & \nobreak\hspace{0pt}-1.31 & \nobreak\hspace{0pt}0.0000 & \nobreak\hspace{0pt}0.3988 & \nobreak\hspace{0pt}0.8372 \\
\nobreak\hspace{0pt}ConvNeXt-V2-Tiny (448 px) & \nobreak\hspace{0pt}0.9588 & \nobreak\hspace{0pt}0.9456 & \nobreak\hspace{0pt}-1.32 & \nobreak\hspace{0pt}0.0000 & \nobreak\hspace{0pt}0.4031 & \nobreak\hspace{0pt}-- \\
\nobreak\hspace{0pt}ConvNeXt-V2-Tiny (448 px, matched) & \nobreak\hspace{0pt}0.9419 & \nobreak\hspace{0pt}0.9338 & \nobreak\hspace{0pt}-0.81 & \nobreak\hspace{0pt}0.0072 & \nobreak\hspace{0pt}0.3188 & \nobreak\hspace{0pt}-- \\
\nobreak\hspace{0pt}ViT-L/16 & \nobreak\hspace{0pt}0.9525 & \nobreak\hspace{0pt}0.9462 & \nobreak\hspace{0pt}-0.63 & \nobreak\hspace{0pt}0.0063 & \nobreak\hspace{0pt}0.2744 & \nobreak\hspace{0pt}0.8233 \\
\nobreak\hspace{0pt}DINOv2-L & \nobreak\hspace{0pt}0.9475 & \nobreak\hspace{0pt}0.9438 & \nobreak\hspace{0pt}-0.37 & \nobreak\hspace{0pt}0.2101 & \nobreak\hspace{0pt}0.3238 & \nobreak\hspace{0pt}0.8419 \\
\nobreak\hspace{0pt}MedSigLIP (224 px) & \nobreak\hspace{0pt}0.9531 & \nobreak\hspace{0pt}0.9431 & \nobreak\hspace{0pt}-1.00 & \nobreak\hspace{0pt}0.0015 & \nobreak\hspace{0pt}0.2581 & \nobreak\hspace{0pt}-- \\
\nobreak\hspace{0pt}MedSigLIP-448 & \nobreak\hspace{0pt}0.9538 & \nobreak\hspace{0pt}0.9406 & \nobreak\hspace{0pt}-1.32 & \nobreak\hspace{0pt}0.0000 & \nobreak\hspace{0pt}0.2588 & \nobreak\hspace{0pt}0.8640 \\
\bottomrule
\end{tabular}
\end{table}

\begin{table}[!htbp]
\centering
\small
\setlength{\tabcolsep}{3.0pt}
\renewcommand{\arraystretch}{1.12}
\caption{Frozen-feature linear probes (pretrained, not fine-tuned backbones): test accuracy on full, brain-only and background-only images (chance 0.25).}
\label{tab:16}
\begin{tabular}{>{\raggedright\arraybackslash}p{0.422\linewidth}>{\raggedright\arraybackslash}p{0.157\linewidth}>{\raggedright\arraybackslash}p{0.157\linewidth}>{\raggedright\arraybackslash}p{0.209\linewidth}}
\toprule
\nobreak\hspace{0pt}\textbf{Frozen pretrained backbone + logistic regression} & \nobreak\hspace{0pt}\textbf{Full image} & \nobreak\hspace{0pt}\textbf{Brain only} & \nobreak\hspace{0pt}\textbf{Background only} \\
\midrule
\nobreak\hspace{0pt}ConvNeXt-V2-Tiny & \nobreak\hspace{0pt}0.9150 & \nobreak\hspace{0pt}0.9081 & \nobreak\hspace{0pt}0.8406 \\
\nobreak\hspace{0pt}DINOv2-L & \nobreak\hspace{0pt}0.9300 & \nobreak\hspace{0pt}0.9338 & \nobreak\hspace{0pt}0.8850 \\
\nobreak\hspace{0pt}MedSigLIP-448 & \nobreak\hspace{0pt}0.9381 & \nobreak\hspace{0pt}0.9438 & \nobreak\hspace{0pt}0.8531 \\
\bottomrule
\end{tabular}
\end{table}

\subsection{Explainability: Quantitative Evaluation}

\textbf{Localisation against expert masks:} Attribution maps localised tumors poorly (Table~\ref{tab:17}). On the 165 test images with expert masks, the tumor occupied on average 1.6\% of the image. Grad-CAM-family maps placed between 0.25 and 6.2 times this chance level of their energy inside the tumor. The pointing-game hit rate was 25.5\% at best (ConvNeXt-V2-Tiny Grad-CAM) and below 10\% for many model--method pairs. Several combinations were at or below chance: DINOv2-L Grad-CAM and Grad-CAM++, Swin-V2-S Grad-CAM++, and MedSigLIP-448 EigenCAM. Integrated Gradients and LIME reached 1.4--3.0 times chance.

\textbf{Border attribution:} The border analysis (Table~\ref{tab:18}) shows systematic differences between families. The three ViT-based models placed 33--42\% of their Grad-CAM energy in the outer 10\% of the image, the region containing the image border and background. The CNNs and Swin-V2-S placed less than 4\% there and 92--99\% inside the brain.

\textbf{Randomisation and stability:} In the model-randomisation test (Table~\ref{tab:19}), maps changed once deeper layers were re-initialised: Spearman correlations with the original maps fell to 0.19 or lower, and were negative for several models. The maps therefore depend on the learned weights. For ConvNeXt-V2-Tiny (0.55) and Swin-V2-S (0.43), maps were partly preserved after re-initialising the classifier head alone. Under small Gaussian noise, maps were stable for most models (Spearman 0.85--0.98), but much less so for MedSigLIP-448 (0.59).

\textbf{Ensemble explanation:} The occlusion-based explanation of the weighted-average ensemble placed 2.8 times chance of its energy inside the tumor, with a pointing-game hit rate of 7.5\%.

\begin{figure}[!htbp]
\centering
\includegraphics[width=0.88\linewidth,height=0.82\textheight,keepaspectratio]{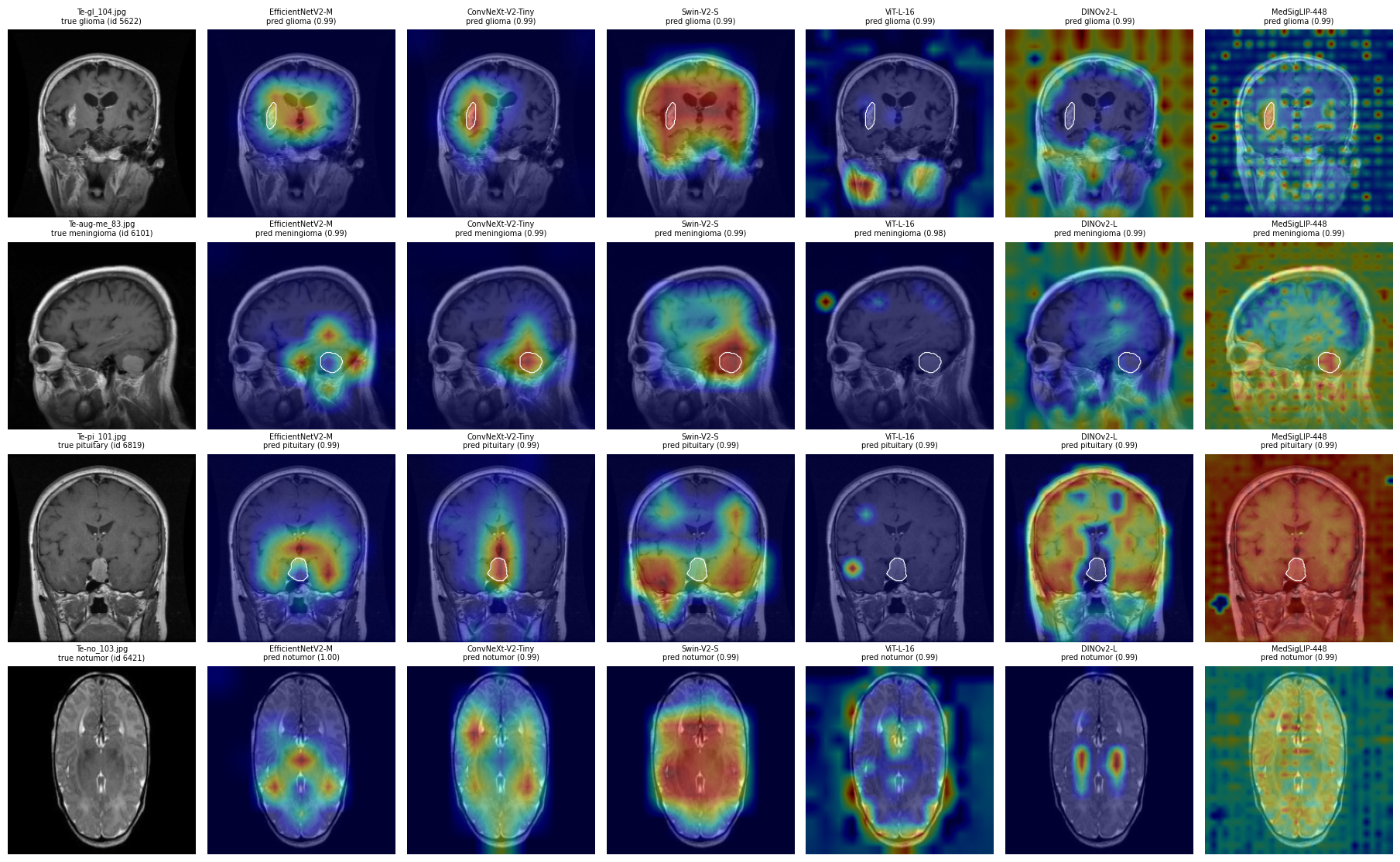}
\caption{Grad-CAM maps of the six primary models for four identified test images (one per class), with the expert tumor outline in white where available, and each model's predicted class and probability.}
\label{fig:12}
\end{figure}

\begin{figure}[!htbp]
\centering
\includegraphics[width=0.85\linewidth,height=0.82\textheight,keepaspectratio]{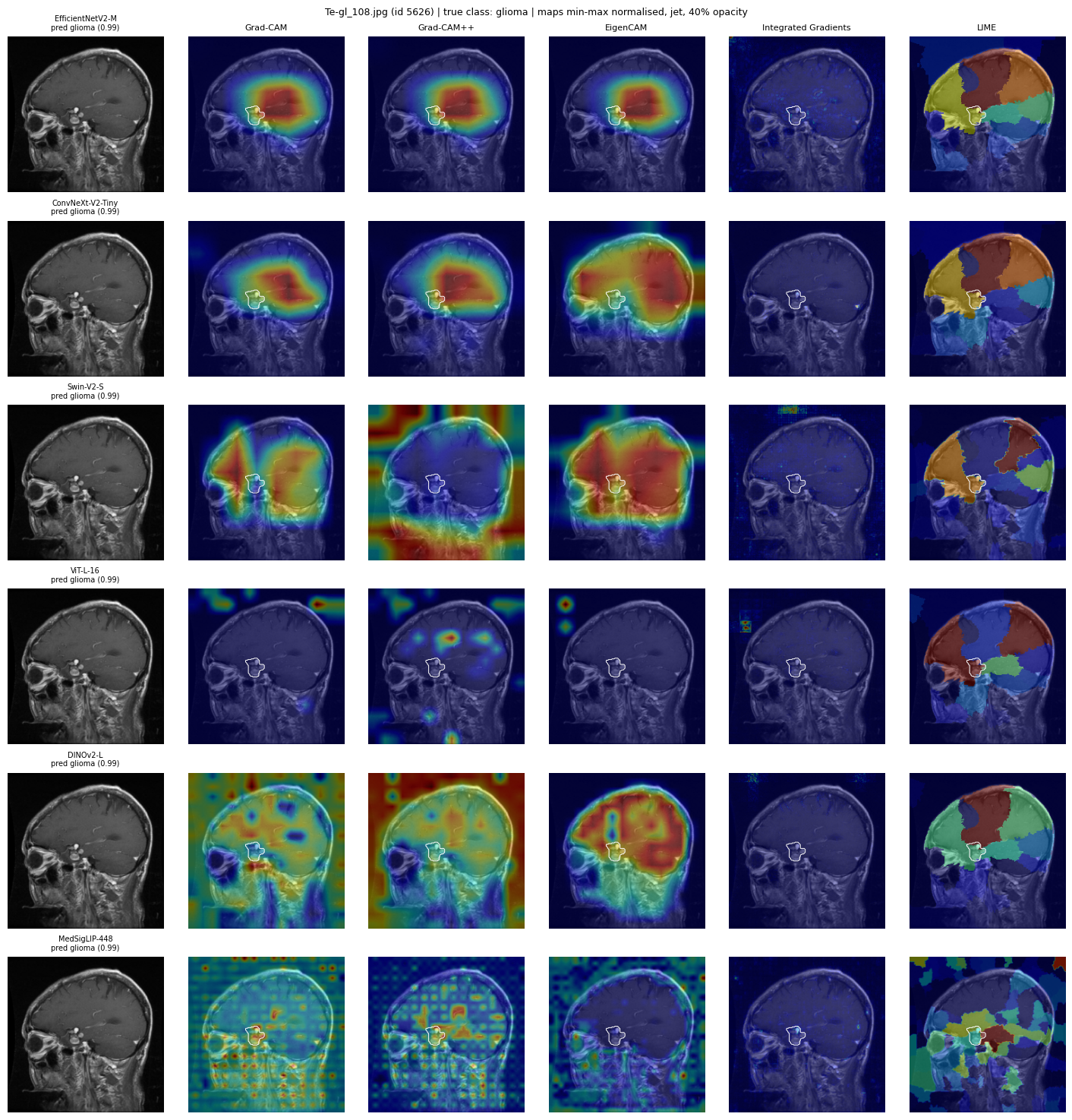}
\caption{Five attribution methods (columns) for every model (rows) on an identified glioma test image (Te-gl\_259). The expert tumor outline is shown in white. Maps are min--max normalised and overlaid at 40\% opacity.}
\label{fig:13}
\end{figure}

\begin{figure}[!htbp]
\centering
\includegraphics[width=0.85\linewidth,height=0.82\textheight,keepaspectratio]{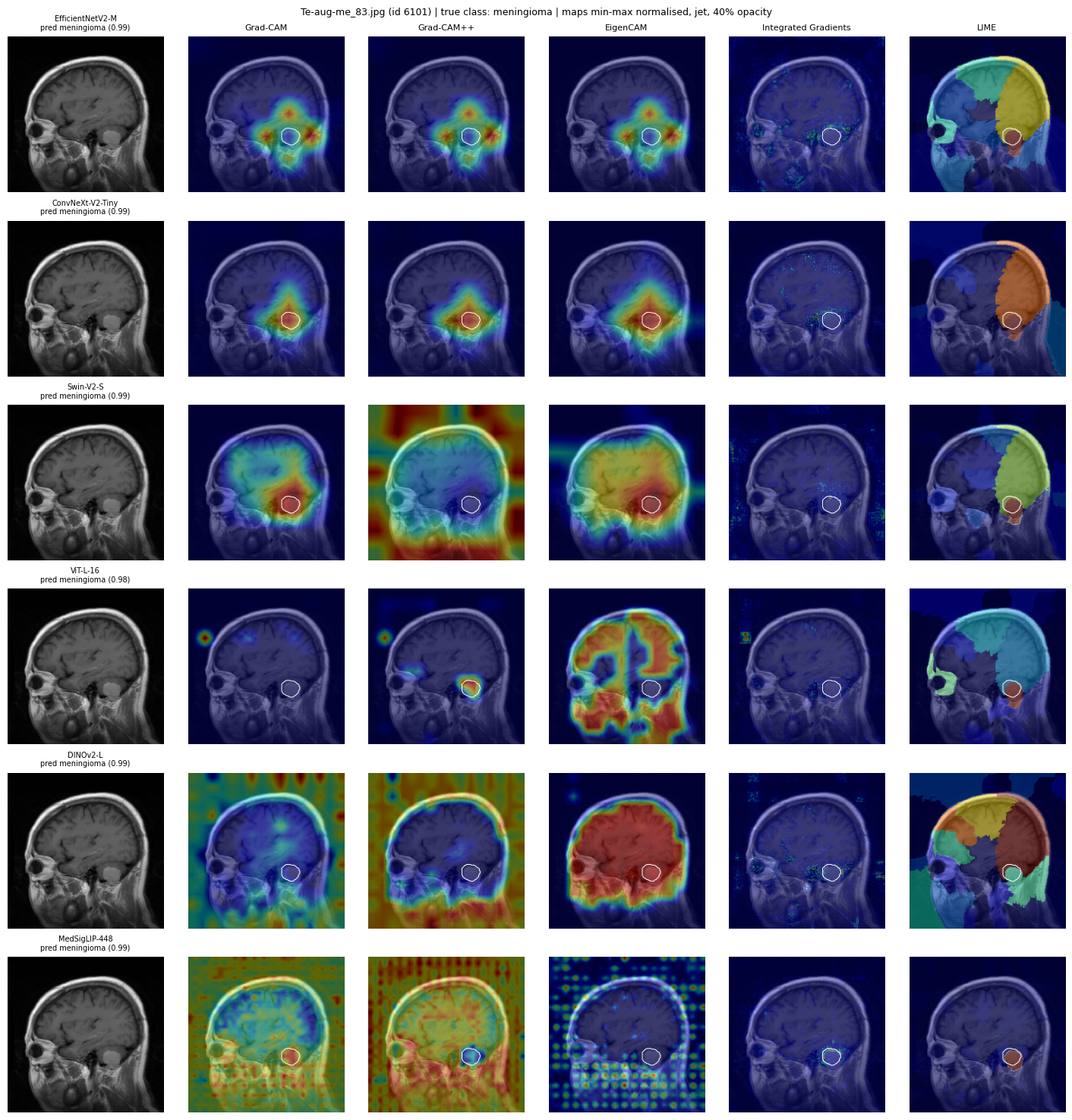}
\caption{As Figure~\ref{fig:13}, for an identified meningioma test image.}
\label{fig:14}
\end{figure}

\begin{figure}[!htbp]
\centering
\includegraphics[width=0.85\linewidth,height=0.82\textheight,keepaspectratio]{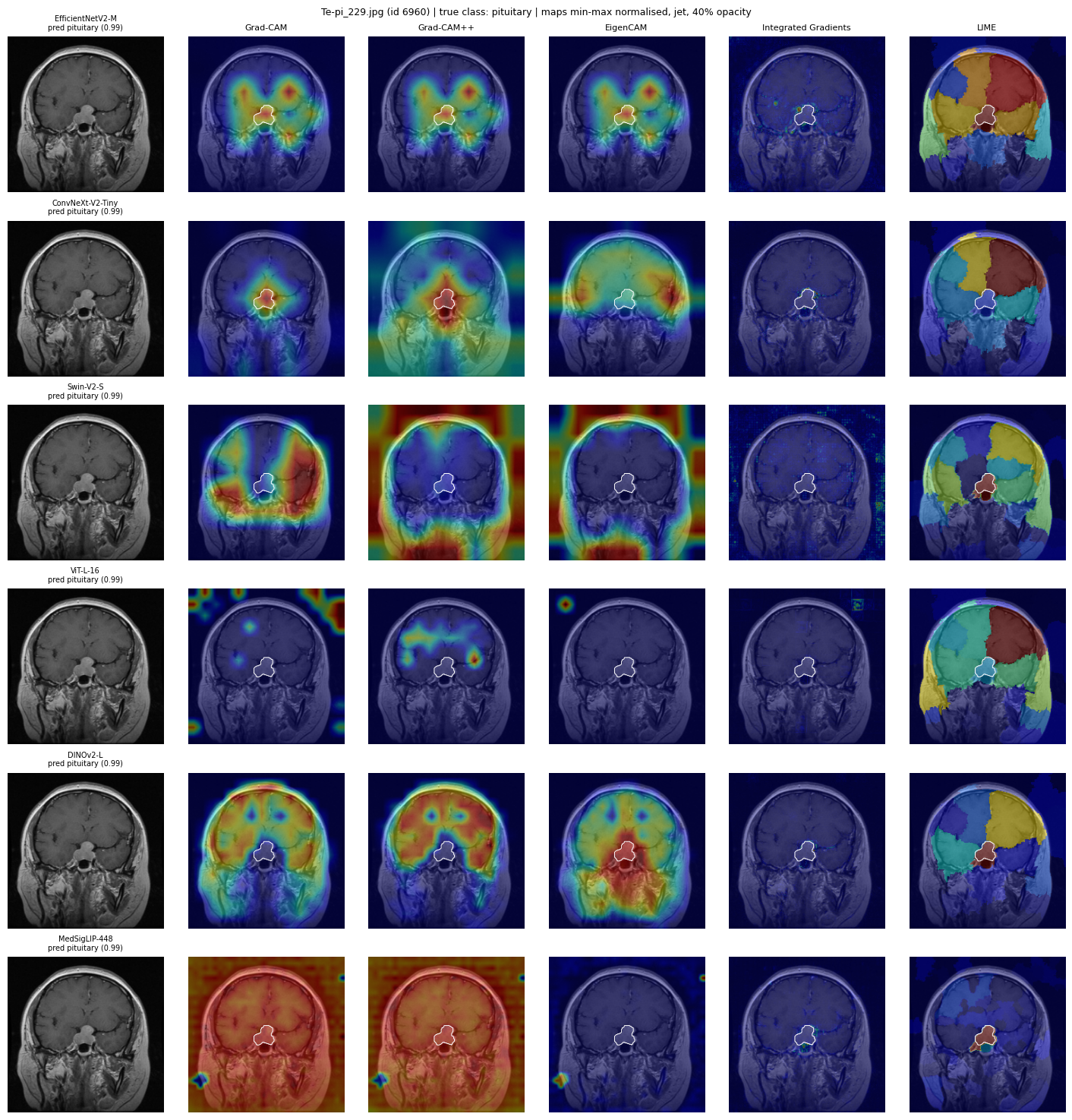}
\caption{As Figure~\ref{fig:13}, for an identified pituitary-tumor test image.}
\label{fig:15}
\end{figure}

\begin{figure}[!htbp]
\centering
\includegraphics[width=0.88\linewidth,height=0.82\textheight,keepaspectratio]{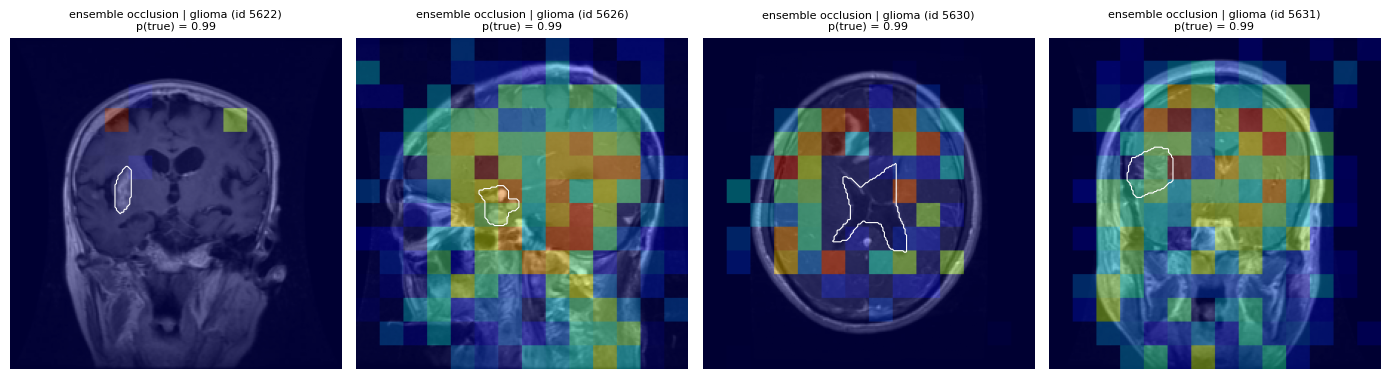}
\caption{Occlusion-sensitivity explanations of the weighted-average ensemble itself (14 $\times$ 14 patch grid; each occluded image passed through all six models) for four tumor test images, with the expert outline and the ensemble's probability for the true class.}
\label{fig:16}
\end{figure}

\begin{figure}[!htbp]
\centering
\includegraphics[width=0.78\linewidth,height=0.82\textheight,keepaspectratio]{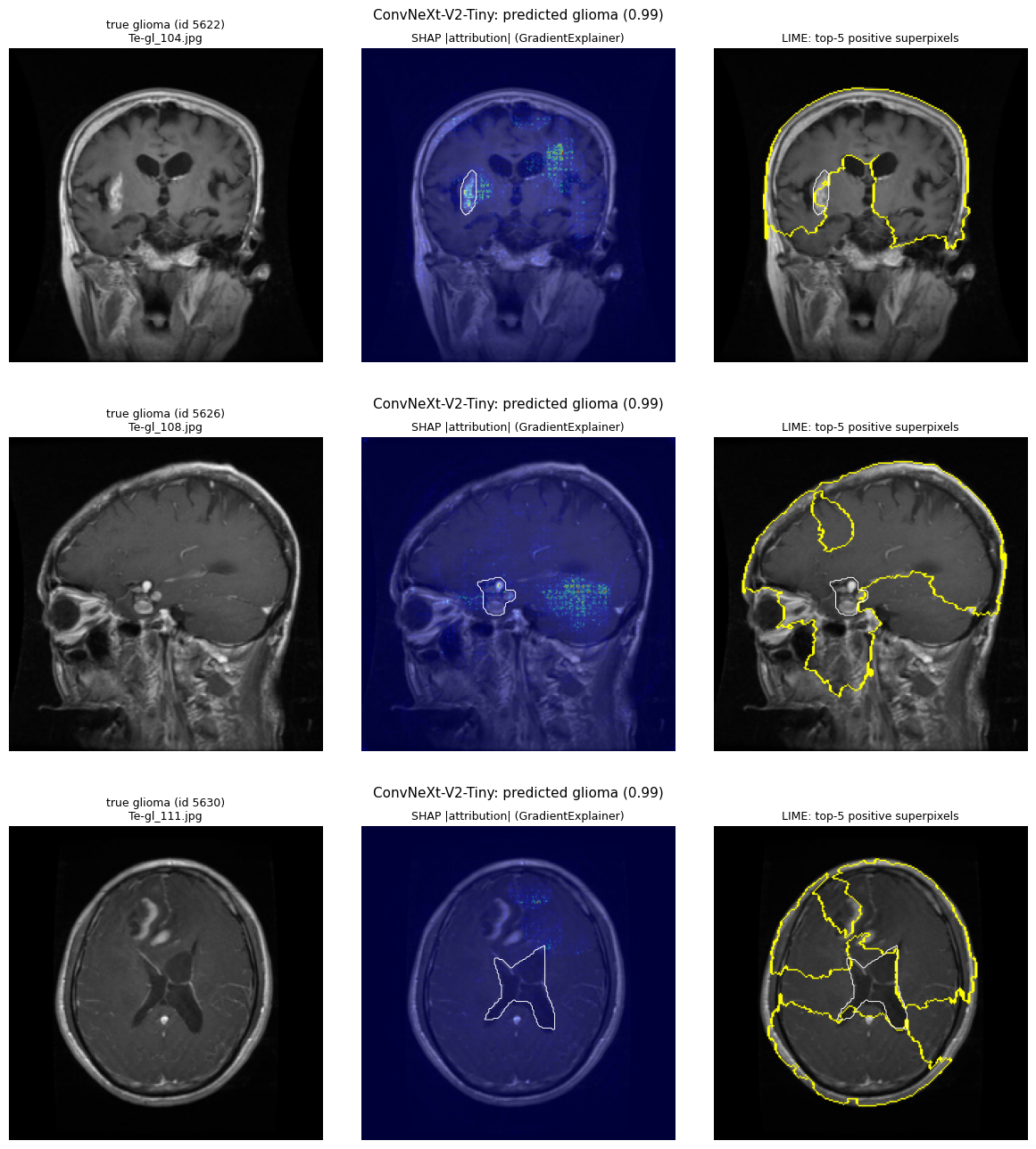}
\caption{SHAP (GradientExplainer) and LIME explanations of ConvNeXt-V2-Tiny for three identified test images with expert tumor outlines.}
\label{fig:17}
\end{figure}

\begin{figure}[!htbp]
\centering
\includegraphics[width=0.78\linewidth,height=0.82\textheight,keepaspectratio]{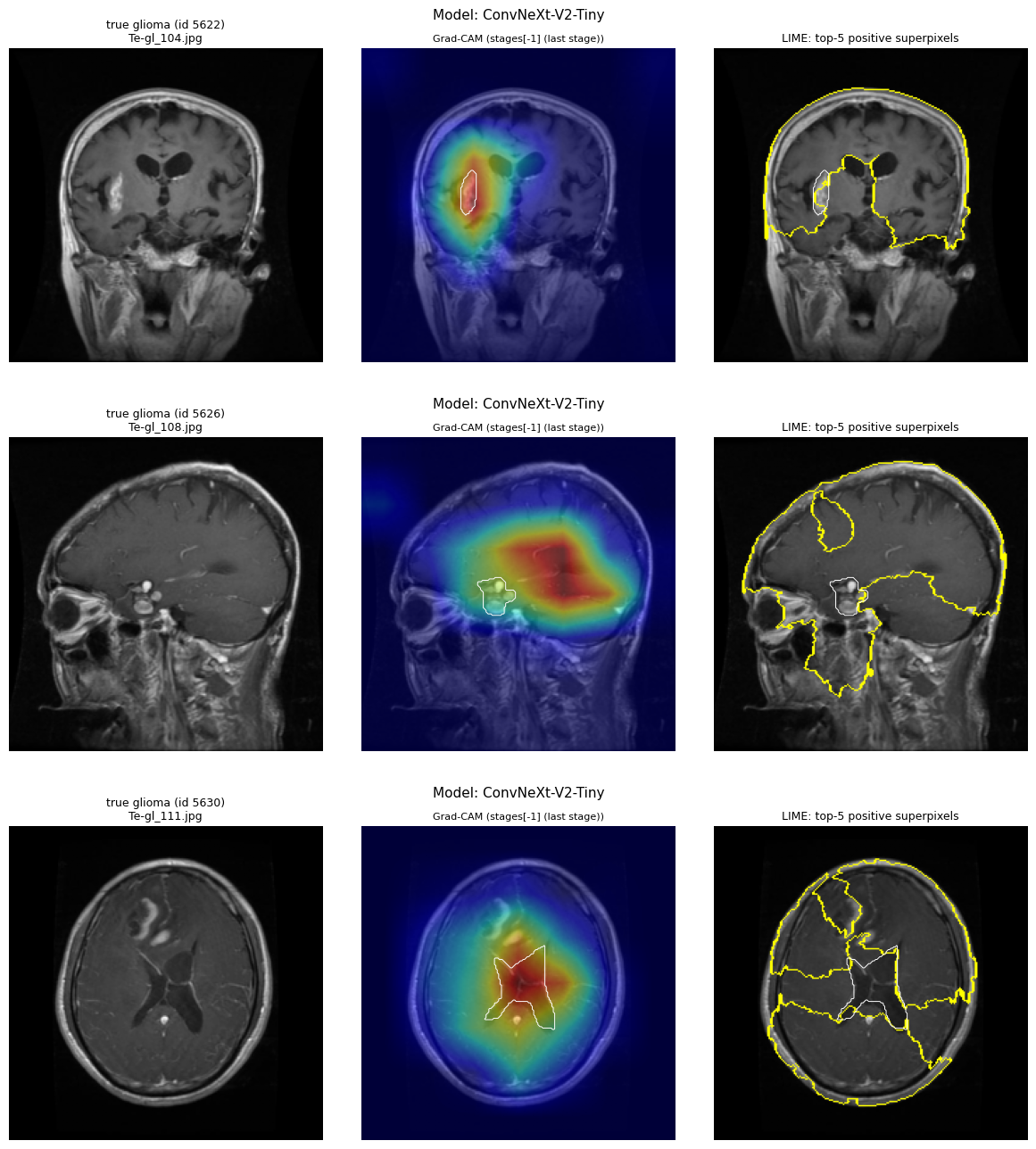}
\caption{Grad-CAM and LIME explanations of ConvNeXt-V2-Tiny for the same three test images.}
\label{fig:18}
\end{figure}

\textbf{Qualitative figures:} Figures~\ref{fig:12}--\ref{fig:18} show the qualitative maps for identified test images: Grad-CAM for all models, the five-method comparison, the ensemble occlusion maps, and the SHAP, LIME and Grad-CAM panels. They illustrate the same pattern as the quantitative results: maps are often diffuse, frequently miss small tumors, and for the transformer-based models often highlight the image border. These maps should therefore not be read as evidence that the models base their decisions on tumor tissue.

\begin{table}[!htbp]
\centering
\scriptsize
\setlength{\tabcolsep}{3.0pt}
\renewcommand{\arraystretch}{1.12}
\caption{Localisation of attribution maps (true class) against expert Figshare tumor masks on test images. The chance level of energy in the mask equals the mask area fraction.}
\label{tab:17}
\begin{tabular}{>{\raggedright\arraybackslash}p{0.209\linewidth}>{\raggedright\arraybackslash}p{0.165\linewidth}>{\raggedright\arraybackslash}p{0.043\linewidth}>{\raggedright\arraybackslash}p{0.144\linewidth}>{\raggedright\arraybackslash}p{0.080\linewidth}>{\raggedright\arraybackslash}p{0.080\linewidth}>{\raggedright\arraybackslash}p{0.077\linewidth}>{\raggedright\arraybackslash}p{0.096\linewidth}}
\toprule
\nobreak\hspace{0pt}\textbf{Model} & \nobreak\hspace{0pt}\textbf{Method} & \nobreak\hspace{0pt}\textbf{n} & \nobreak\hspace{0pt}\textbf{Pointing-game hit rate} & \nobreak\hspace{0pt}\textbf{Energy in mask} & \nobreak\hspace{0pt}\textbf{Chance (mask area)} & \nobreak\hspace{0pt}\textbf{Energy / chance} & \nobreak\hspace{0pt}\textbf{IoU (top-20\% map)} \\
\midrule
\nobreak\hspace{0pt}EfficientNetV2-M & \nobreak\hspace{0pt}Grad-CAM & \nobreak\hspace{0pt}165 & \nobreak\hspace{0pt}0.139 & \nobreak\hspace{0pt}0.085 & \nobreak\hspace{0pt}0.016 & \nobreak\hspace{0pt}5.25 & \nobreak\hspace{0pt}0.076 \\
 & \nobreak\hspace{0pt}Grad-CAM++ & \nobreak\hspace{0pt}165 & \nobreak\hspace{0pt}0.145 & \nobreak\hspace{0pt}0.092 & \nobreak\hspace{0pt}0.016 & \nobreak\hspace{0pt}5.63 & \nobreak\hspace{0pt}0.076 \\
 & \nobreak\hspace{0pt}EigenCAM & \nobreak\hspace{0pt}165 & \nobreak\hspace{0pt}0.139 & \nobreak\hspace{0pt}0.094 & \nobreak\hspace{0pt}0.016 & \nobreak\hspace{0pt}5.77 & \nobreak\hspace{0pt}0.076 \\
 & \nobreak\hspace{0pt}Integrated Gradients & \nobreak\hspace{0pt}40 & \nobreak\hspace{0pt}0.000 & \nobreak\hspace{0pt}0.045 & \nobreak\hspace{0pt}0.021 & \nobreak\hspace{0pt}2.12 & \nobreak\hspace{0pt}0.055 \\
 & \nobreak\hspace{0pt}LIME & \nobreak\hspace{0pt}20 & \nobreak\hspace{0pt}0.050 & \nobreak\hspace{0pt}0.048 & \nobreak\hspace{0pt}0.020 & \nobreak\hspace{0pt}2.37 & \nobreak\hspace{0pt}0.055 \\
\nobreak\hspace{0pt}ConvNeXt-V2-Tiny & \nobreak\hspace{0pt}Grad-CAM & \nobreak\hspace{0pt}165 & \nobreak\hspace{0pt}0.255 & \nobreak\hspace{0pt}0.095 & \nobreak\hspace{0pt}0.016 & \nobreak\hspace{0pt}5.82 & \nobreak\hspace{0pt}0.073 \\
 & \nobreak\hspace{0pt}Grad-CAM++ & \nobreak\hspace{0pt}165 & \nobreak\hspace{0pt}0.230 & \nobreak\hspace{0pt}0.079 & \nobreak\hspace{0pt}0.016 & \nobreak\hspace{0pt}4.85 & \nobreak\hspace{0pt}0.072 \\
 & \nobreak\hspace{0pt}EigenCAM & \nobreak\hspace{0pt}165 & \nobreak\hspace{0pt}0.206 & \nobreak\hspace{0pt}0.058 & \nobreak\hspace{0pt}0.016 & \nobreak\hspace{0pt}3.58 & \nobreak\hspace{0pt}0.068 \\
 & \nobreak\hspace{0pt}Integrated Gradients & \nobreak\hspace{0pt}40 & \nobreak\hspace{0pt}0.200 & \nobreak\hspace{0pt}0.049 & \nobreak\hspace{0pt}0.021 & \nobreak\hspace{0pt}2.31 & \nobreak\hspace{0pt}0.054 \\
 & \nobreak\hspace{0pt}LIME & \nobreak\hspace{0pt}20 & \nobreak\hspace{0pt}0.000 & \nobreak\hspace{0pt}0.036 & \nobreak\hspace{0pt}0.020 & \nobreak\hspace{0pt}1.79 & \nobreak\hspace{0pt}0.036 \\
\nobreak\hspace{0pt}Swin-V2-S & \nobreak\hspace{0pt}Grad-CAM & \nobreak\hspace{0pt}165 & \nobreak\hspace{0pt}0.085 & \nobreak\hspace{0pt}0.055 & \nobreak\hspace{0pt}0.016 & \nobreak\hspace{0pt}3.40 & \nobreak\hspace{0pt}0.062 \\
 & \nobreak\hspace{0pt}Grad-CAM++ & \nobreak\hspace{0pt}165 & \nobreak\hspace{0pt}0.006 & \nobreak\hspace{0pt}0.005 & \nobreak\hspace{0pt}0.016 & \nobreak\hspace{0pt}0.33 & \nobreak\hspace{0pt}0.004 \\
 & \nobreak\hspace{0pt}EigenCAM & \nobreak\hspace{0pt}165 & \nobreak\hspace{0pt}0.127 & \nobreak\hspace{0pt}0.040 & \nobreak\hspace{0pt}0.016 & \nobreak\hspace{0pt}2.48 & \nobreak\hspace{0pt}0.068 \\
 & \nobreak\hspace{0pt}Integrated Gradients & \nobreak\hspace{0pt}40 & \nobreak\hspace{0pt}0.075 & \nobreak\hspace{0pt}0.030 & \nobreak\hspace{0pt}0.021 & \nobreak\hspace{0pt}1.41 & \nobreak\hspace{0pt}0.034 \\
 & \nobreak\hspace{0pt}LIME & \nobreak\hspace{0pt}20 & \nobreak\hspace{0pt}0.000 & \nobreak\hspace{0pt}0.034 & \nobreak\hspace{0pt}0.020 & \nobreak\hspace{0pt}1.67 & \nobreak\hspace{0pt}0.035 \\
\nobreak\hspace{0pt}ViT-L/16 & \nobreak\hspace{0pt}Grad-CAM & \nobreak\hspace{0pt}165 & \nobreak\hspace{0pt}0.176 & \nobreak\hspace{0pt}0.060 & \nobreak\hspace{0pt}0.016 & \nobreak\hspace{0pt}3.66 & \nobreak\hspace{0pt}0.025 \\
 & \nobreak\hspace{0pt}Grad-CAM++ & \nobreak\hspace{0pt}165 & \nobreak\hspace{0pt}0.212 & \nobreak\hspace{0pt}0.100 & \nobreak\hspace{0pt}0.016 & \nobreak\hspace{0pt}6.15 & \nobreak\hspace{0pt}0.019 \\
 & \nobreak\hspace{0pt}EigenCAM & \nobreak\hspace{0pt}165 & \nobreak\hspace{0pt}0.012 & \nobreak\hspace{0pt}0.020 & \nobreak\hspace{0pt}0.016 & \nobreak\hspace{0pt}1.24 & \nobreak\hspace{0pt}0.014 \\
 & \nobreak\hspace{0pt}Integrated Gradients & \nobreak\hspace{0pt}40 & \nobreak\hspace{0pt}0.000 & \nobreak\hspace{0pt}0.040 & \nobreak\hspace{0pt}0.021 & \nobreak\hspace{0pt}1.87 & \nobreak\hspace{0pt}0.049 \\
 & \nobreak\hspace{0pt}LIME & \nobreak\hspace{0pt}20 & \nobreak\hspace{0pt}0.000 & \nobreak\hspace{0pt}0.040 & \nobreak\hspace{0pt}0.020 & \nobreak\hspace{0pt}1.98 & \nobreak\hspace{0pt}0.037 \\
\nobreak\hspace{0pt}DINOv2-L & \nobreak\hspace{0pt}Grad-CAM & \nobreak\hspace{0pt}165 & \nobreak\hspace{0pt}0.006 & \nobreak\hspace{0pt}0.008 & \nobreak\hspace{0pt}0.016 & \nobreak\hspace{0pt}0.48 & \nobreak\hspace{0pt}0.004 \\
 & \nobreak\hspace{0pt}Grad-CAM++ & \nobreak\hspace{0pt}165 & \nobreak\hspace{0pt}0.000 & \nobreak\hspace{0pt}0.008 & \nobreak\hspace{0pt}0.016 & \nobreak\hspace{0pt}0.51 & \nobreak\hspace{0pt}0.000 \\
 & \nobreak\hspace{0pt}EigenCAM & \nobreak\hspace{0pt}165 & \nobreak\hspace{0pt}0.030 & \nobreak\hspace{0pt}0.043 & \nobreak\hspace{0pt}0.016 & \nobreak\hspace{0pt}2.65 & \nobreak\hspace{0pt}0.058 \\
 & \nobreak\hspace{0pt}Integrated Gradients & \nobreak\hspace{0pt}40 & \nobreak\hspace{0pt}0.075 & \nobreak\hspace{0pt}0.053 & \nobreak\hspace{0pt}0.021 & \nobreak\hspace{0pt}2.50 & \nobreak\hspace{0pt}0.056 \\
 & \nobreak\hspace{0pt}LIME & \nobreak\hspace{0pt}20 & \nobreak\hspace{0pt}0.000 & \nobreak\hspace{0pt}0.047 & \nobreak\hspace{0pt}0.020 & \nobreak\hspace{0pt}2.33 & \nobreak\hspace{0pt}0.041 \\
\nobreak\hspace{0pt}MedSigLIP-448 & \nobreak\hspace{0pt}Grad-CAM & \nobreak\hspace{0pt}165 & \nobreak\hspace{0pt}0.073 & \nobreak\hspace{0pt}0.020 & \nobreak\hspace{0pt}0.016 & \nobreak\hspace{0pt}1.20 & \nobreak\hspace{0pt}0.038 \\
 & \nobreak\hspace{0pt}Grad-CAM++ & \nobreak\hspace{0pt}165 & \nobreak\hspace{0pt}0.152 & \nobreak\hspace{0pt}0.020 & \nobreak\hspace{0pt}0.016 & \nobreak\hspace{0pt}1.25 & \nobreak\hspace{0pt}0.039 \\
 & \nobreak\hspace{0pt}EigenCAM & \nobreak\hspace{0pt}165 & \nobreak\hspace{0pt}0.061 & \nobreak\hspace{0pt}0.004 & \nobreak\hspace{0pt}0.016 & \nobreak\hspace{0pt}0.25 & \nobreak\hspace{0pt}0.003 \\
 & \nobreak\hspace{0pt}Integrated Gradients & \nobreak\hspace{0pt}40 & \nobreak\hspace{0pt}0.075 & \nobreak\hspace{0pt}0.054 & \nobreak\hspace{0pt}0.021 & \nobreak\hspace{0pt}2.56 & \nobreak\hspace{0pt}0.066 \\
 & \nobreak\hspace{0pt}LIME & \nobreak\hspace{0pt}20 & \nobreak\hspace{0pt}0.200 & \nobreak\hspace{0pt}0.061 & \nobreak\hspace{0pt}0.020 & \nobreak\hspace{0pt}2.99 & \nobreak\hspace{0pt}0.070 \\
\bottomrule
\end{tabular}
\end{table}

\begin{table}[!htbp]
\centering
\scriptsize
\setlength{\tabcolsep}{3.0pt}
\renewcommand{\arraystretch}{1.12}
\caption{Occlusion-based explanation of the weighted-average ensemble and of each model (40 images), and the distribution of Grad-CAM energy between the brain, the outer 10\% border of the image and the tumor mask for the same images.}
\label{tab:18}
\begin{tabular}{>{\raggedright\arraybackslash}p{0.189\linewidth}>{\raggedright\arraybackslash}p{0.109\linewidth}>{\raggedright\arraybackslash}p{0.098\linewidth}>{\raggedright\arraybackslash}p{0.098\linewidth}>{\raggedright\arraybackslash}p{0.097\linewidth}>{\raggedright\arraybackslash}p{0.101\linewidth}>{\raggedright\arraybackslash}p{0.101\linewidth}>{\raggedright\arraybackslash}p{0.101\linewidth}}
\toprule
\nobreak\hspace{0pt}\textbf{Model} & \nobreak\hspace{0pt}\textbf{n (occlusion)} & \nobreak\hspace{0pt}\textbf{Occlusion pointing} & \nobreak\hspace{0pt}\textbf{Occlusion energy in mask} & \nobreak\hspace{0pt}\textbf{Occlusion energy / chance} & \nobreak\hspace{0pt}\textbf{Grad-CAM energy in brain (\%)} & \nobreak\hspace{0pt}\textbf{Grad-CAM energy in outer 10\% border (\%)} & \nobreak\hspace{0pt}\textbf{Grad-CAM energy in tumor mask (\%)} \\
\midrule
\nobreak\hspace{0pt}ConvNeXt-V2-Tiny & \nobreak\hspace{0pt}40 & \nobreak\hspace{0pt}0.050 & \nobreak\hspace{0pt}0.038 & \nobreak\hspace{0pt}1.77 & \nobreak\hspace{0pt}91.6 & \nobreak\hspace{0pt}3.9 & \nobreak\hspace{0pt}9.6 \\
\nobreak\hspace{0pt}DINOv2-L & \nobreak\hspace{0pt}40 & \nobreak\hspace{0pt}0.275 & \nobreak\hspace{0pt}0.085 & \nobreak\hspace{0pt}3.97 & \nobreak\hspace{0pt}40.4 & \nobreak\hspace{0pt}41.5 & \nobreak\hspace{0pt}1.3 \\
\nobreak\hspace{0pt}EfficientNetV2-M & \nobreak\hspace{0pt}40 & \nobreak\hspace{0pt}0.150 & \nobreak\hspace{0pt}0.065 & \nobreak\hspace{0pt}3.03 & \nobreak\hspace{0pt}98.8 & \nobreak\hspace{0pt}0.5 & \nobreak\hspace{0pt}11.0 \\
\nobreak\hspace{0pt}Ensemble (weighted average) & \nobreak\hspace{0pt}40 & \nobreak\hspace{0pt}0.075 & \nobreak\hspace{0pt}0.060 & \nobreak\hspace{0pt}2.80 & \nobreak\hspace{0pt}-- & \nobreak\hspace{0pt}-- & \nobreak\hspace{0pt}-- \\
\nobreak\hspace{0pt}MedSigLIP-448 & \nobreak\hspace{0pt}40 & \nobreak\hspace{0pt}0.025 & \nobreak\hspace{0pt}0.040 & \nobreak\hspace{0pt}1.88 & \nobreak\hspace{0pt}49.8 & \nobreak\hspace{0pt}35.2 & \nobreak\hspace{0pt}2.6 \\
\nobreak\hspace{0pt}Swin-V2-S & \nobreak\hspace{0pt}40 & \nobreak\hspace{0pt}0.100 & \nobreak\hspace{0pt}0.027 & \nobreak\hspace{0pt}1.25 & \nobreak\hspace{0pt}96.6 & \nobreak\hspace{0pt}0.8 & \nobreak\hspace{0pt}6.4 \\
\nobreak\hspace{0pt}ViT-L/16 & \nobreak\hspace{0pt}40 & \nobreak\hspace{0pt}0.125 & \nobreak\hspace{0pt}0.053 & \nobreak\hspace{0pt}2.50 & \nobreak\hspace{0pt}51.7 & \nobreak\hspace{0pt}32.9 & \nobreak\hspace{0pt}6.1 \\
\bottomrule
\end{tabular}
\end{table}

\begin{table}[!htbp]
\centering
\footnotesize
\setlength{\tabcolsep}{3.0pt}
\renewcommand{\arraystretch}{1.12}
\caption{Model-randomisation sanity check (Spearman correlation between Grad-CAM maps of the trained and progressively re-initialised model; 20 images) and stability under perturbation (Spearman correlation; 40 images).}
\label{tab:19}
\begin{tabular}{>{\raggedright\arraybackslash}p{0.241\linewidth}>{\raggedright\arraybackslash}p{0.131\linewidth}>{\raggedright\arraybackslash}p{0.089\linewidth}>{\raggedright\arraybackslash}p{0.090\linewidth}>{\raggedright\arraybackslash}p{0.090\linewidth}>{\raggedright\arraybackslash}p{0.133\linewidth}>{\raggedright\arraybackslash}p{0.133\linewidth}}
\toprule
\nobreak\hspace{0pt}\textbf{Model} & \nobreak\hspace{0pt}\textbf{Head randomised} & \nobreak\hspace{0pt}\textbf{+ last stage / block} & \nobreak\hspace{0pt}\textbf{+ 2 stages / 4 blocks} & \nobreak\hspace{0pt}\textbf{+ 3 stages / 12 blocks} & \nobreak\hspace{0pt}\textbf{Stability: Gaussian noise} & \nobreak\hspace{0pt}\textbf{Stability: horizontal flip} \\
\midrule
\nobreak\hspace{0pt}EfficientNetV2-M & \nobreak\hspace{0pt}-0.696 & \nobreak\hspace{0pt}-0.014 & \nobreak\hspace{0pt}-0.207 & \nobreak\hspace{0pt}-0.192 & \nobreak\hspace{0pt}0.956 & \nobreak\hspace{0pt}0.853 \\
\nobreak\hspace{0pt}ConvNeXt-V2-Tiny & \nobreak\hspace{0pt}0.546 & \nobreak\hspace{0pt}0.174 & \nobreak\hspace{0pt}-0.049 & \nobreak\hspace{0pt}-0.123 & \nobreak\hspace{0pt}0.918 & \nobreak\hspace{0pt}0.921 \\
\nobreak\hspace{0pt}Swin-V2-S & \nobreak\hspace{0pt}0.434 & \nobreak\hspace{0pt}-0.116 & \nobreak\hspace{0pt}-0.172 & \nobreak\hspace{0pt}-0.125 & \nobreak\hspace{0pt}0.978 & \nobreak\hspace{0pt}0.966 \\
\nobreak\hspace{0pt}ViT-L/16 & \nobreak\hspace{0pt}-0.126 & \nobreak\hspace{0pt}0.090 & \nobreak\hspace{0pt}0.133 & \nobreak\hspace{0pt}0.034 & \nobreak\hspace{0pt}0.847 & \nobreak\hspace{0pt}0.720 \\
\nobreak\hspace{0pt}DINOv2-L & \nobreak\hspace{0pt}-0.349 & \nobreak\hspace{0pt}-0.237 & \nobreak\hspace{0pt}-0.434 & \nobreak\hspace{0pt}-0.307 & \nobreak\hspace{0pt}0.978 & \nobreak\hspace{0pt}0.768 \\
\nobreak\hspace{0pt}MedSigLIP-448 & \nobreak\hspace{0pt}0.043 & \nobreak\hspace{0pt}0.120 & \nobreak\hspace{0pt}0.151 & \nobreak\hspace{0pt}0.189 & \nobreak\hspace{0pt}0.592 & \nobreak\hspace{0pt}0.689 \\
\bottomrule
\end{tabular}
\end{table}

\subsection{Decision-Analytic Evaluation and Diagnostic Accuracy (Exploratory)}

\begin{figure}[!htbp]
\centering
\includegraphics[width=0.88\linewidth,height=0.82\textheight,keepaspectratio]{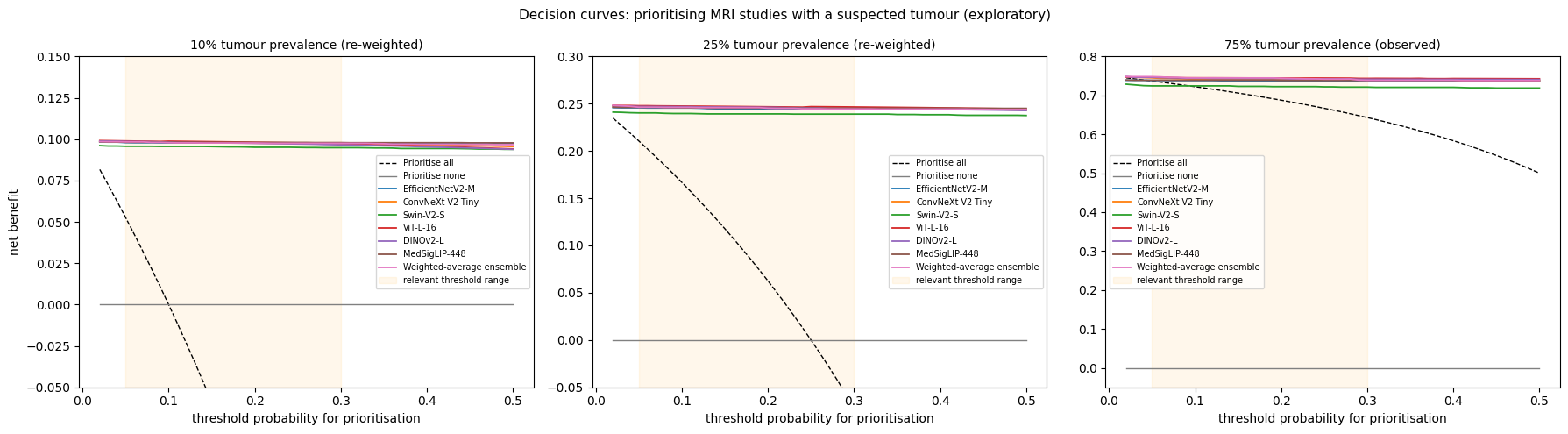}
\caption{Decision curves for prioritising MRI studies with a suspected tumor at 10\% and 25\% re-weighted tumor prevalence and at the observed prevalence (exploratory; shaded band: threshold range of interest, 5--30\%).}
\label{fig:19}
\end{figure}

\textbf{Prioritization decision:} For the hypothetical decision of prioritising studies with a suspected tumor, all models separated tumor from no-tumor images almost perfectly. At a re-weighted tumor prevalence of 10\% and a threshold probability of 0.10, net benefit was 0.0956--0.0987 for the five models in Table~\ref{tab:20}, against a maximum of 0.100 (Table~\ref{tab:20}, Figure~\ref{fig:19}). This corresponds to avoiding 86--89 of every 100 unnecessary prioritisations, compared with prioritising every study. Per 1,000 studies, the models missed 0.7--4.4 tumors and generated 0--4.5 unnecessary prioritisations. Results at 25\% prevalence and at the observed prevalence followed the same pattern (Figure~\ref{fig:19}).

\begin{figure}[!htbp]
\centering
\includegraphics[width=0.88\linewidth,height=0.82\textheight,keepaspectratio]{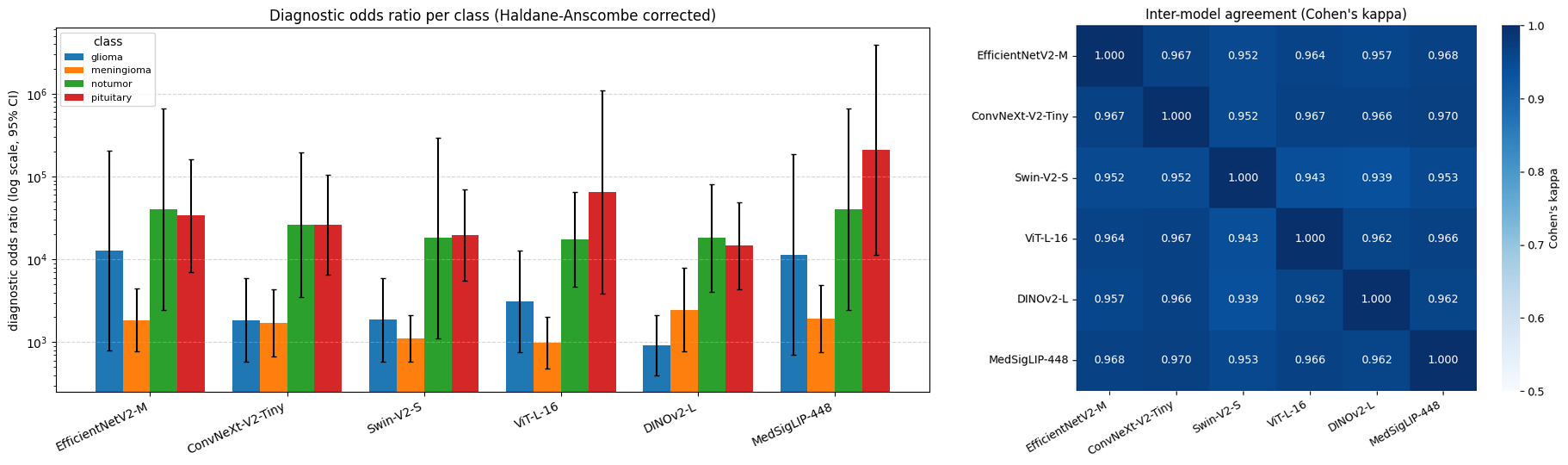}
\caption{Per-class diagnostic odds ratios with 95\% confidence intervals (log scale; left) and pairwise inter-model agreement (Cohen's kappa; right).}
\label{fig:20}
\end{figure}

\begin{figure}[!htbp]
\centering
\includegraphics[width=0.88\linewidth,height=0.82\textheight,keepaspectratio]{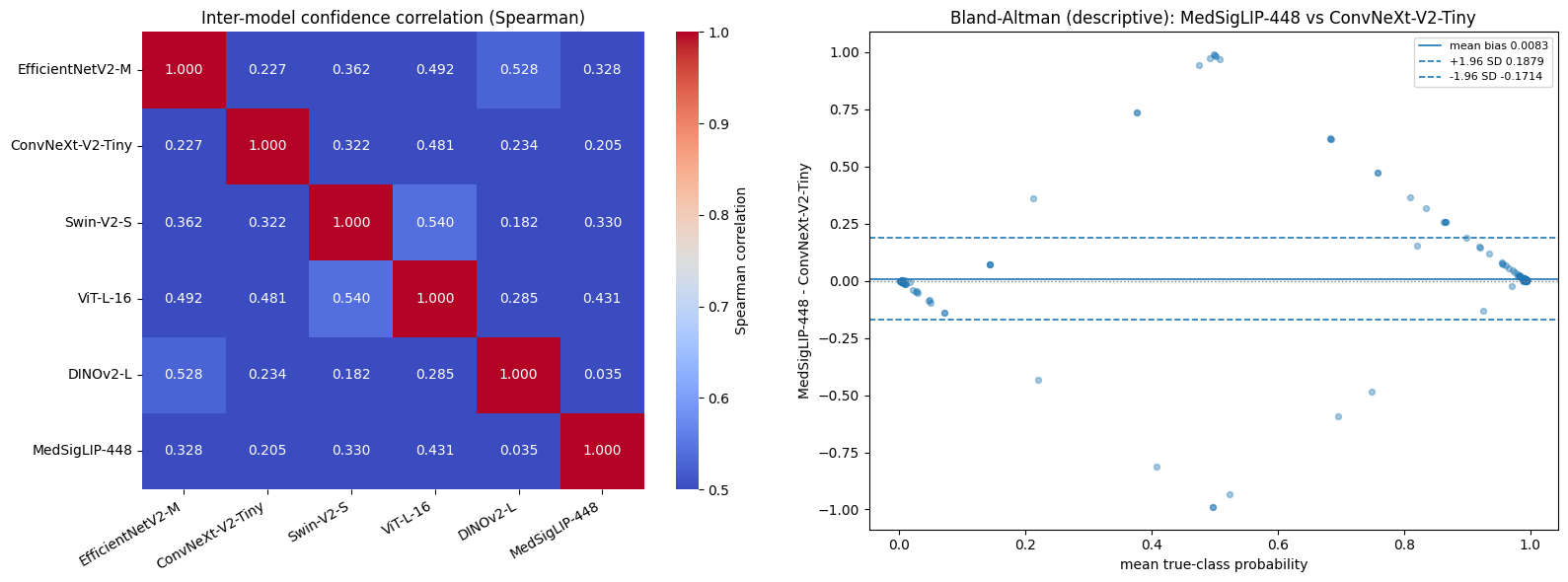}
\caption{Spearman correlation of model confidences (left) and a descriptive Bland--Altman comparison of true-class probabilities between MedSigLIP-448 and ConvNeXt-V2-Tiny (right; mean bias 0.008, limits of agreement $-$0.171 to +0.188).}
\label{fig:21}
\end{figure}

\begin{figure}[!htbp]
\centering
\includegraphics[width=0.88\linewidth,height=0.82\textheight,keepaspectratio]{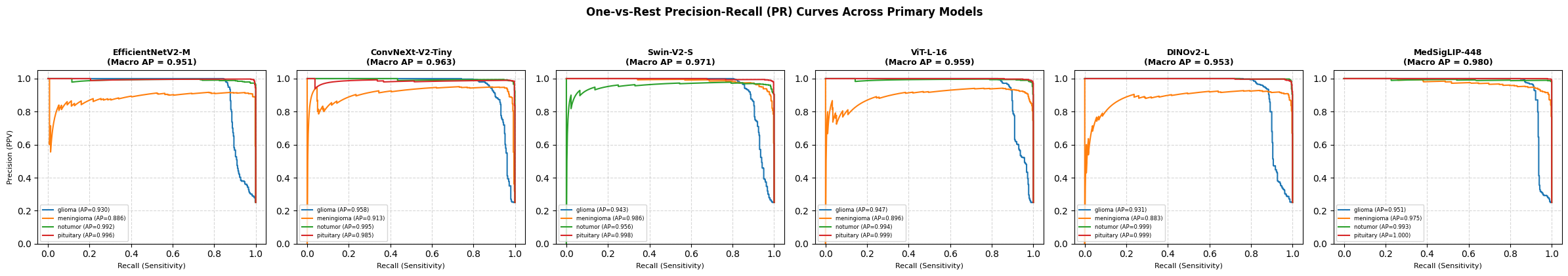}
\caption{One-versus-rest precision--recall curves per class on the test set, with average precision.}
\label{fig:22}
\end{figure}

\begin{figure}[!htbp]
\centering
\includegraphics[width=0.63\linewidth,height=0.82\textheight,keepaspectratio]{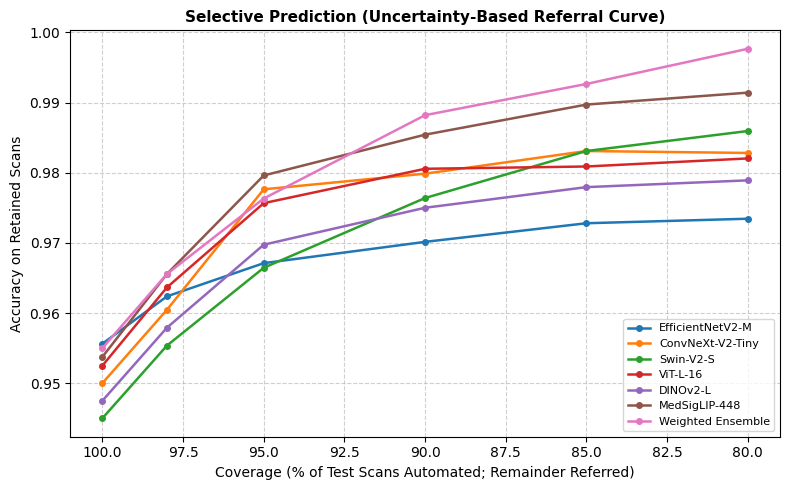}
\caption{Selective prediction: accuracy on the retained test images when the least-confident images are referred for human review.}
\label{fig:23}
\end{figure}

\textbf{Diagnostic accuracy:} The per-class diagnostic odds ratios were very large (Table~\ref{tab:21}, Figure~\ref{fig:20}), reflecting near-perfect separation of the no-tumor and pituitary classes, and the zero cells make them unstable. Pairwise inter-model agreement was high (Cohen's kappa 0.939--0.970; Figure~\ref{fig:20}). Because model confidences cluster near one, their rank correlations were nevertheless low (Spearman 0.04--0.54; Figure~\ref{fig:21}). The descriptive Bland--Altman comparison of MedSigLIP-448 and ConvNeXt-V2-Tiny (Figure~\ref{fig:21}) shows a negligible mean bias with wide limits of agreement, driven by a small number of images on which the two models disagree strongly. Macro-averaged precision--recall AUC was 0.951--0.980 (Table~\ref{tab:22}, Figure~\ref{fig:22}). Referring the 10\% least-confident images would raise accuracy on the retained images to 97.0--98.5\% for the individual models (Figure~\ref{fig:23}).

\textbf{Caveat:} These results show what the models' operating characteristics would imply for one explicitly defined decision under stated assumptions. They are not evidence that using the models would improve patient outcomes or workflow.

\begin{table}[!htbp]
\centering
\scriptsize
\setlength{\tabcolsep}{2.0pt}
\renewcommand{\arraystretch}{1.12}
\caption{Exploratory decision-analytic evaluation of prioritising MRI studies with a suspected tumor at a re-weighted tumor prevalence of 10\%. Rows are shown for the models whose values were tabulated; Figure~\ref{fig:19} shows the complete curves for all models.}
\label{tab:20}
\begin{tabular}{>{\raggedright\arraybackslash}p{0.167\linewidth}>{\raggedright\arraybackslash}p{0.094\linewidth}>{\raggedright\arraybackslash}p{0.085\linewidth}>{\raggedright\arraybackslash}p{0.145\linewidth}>{\raggedright\arraybackslash}p{0.113\linewidth}>{\raggedright\arraybackslash}p{0.113\linewidth}>{\raggedright\arraybackslash}p{0.066\linewidth}>{\raggedright\arraybackslash}p{0.145\linewidth}}
\toprule
\nobreak\hspace{0pt}\textbf{Model} & \nobreak\hspace{0pt}\textbf{Threshold p\_t} & \nobreak\hspace{0pt}\textbf{Net benefit [95\% CI] (maximum 0.100)} & \nobreak\hspace{0pt}\textbf{Unnecessary prioritisations avoided per 100 vs prioritise-all} & \nobreak\hspace{0pt}\textbf{Sensitivity} & \nobreak\hspace{0pt}\textbf{Specificity} & \nobreak\hspace{0pt}\textbf{Missed tumors per 1,000} & \nobreak\hspace{0pt}\textbf{Unnecessary prioritisations per 1,000} \\
\midrule
\nobreak\hspace{0pt}EfficientNetV2-M & \nobreak\hspace{0pt}0.05 & \nobreak\hspace{0pt}0.0979 [0.0970, 0.0987] & \nobreak\hspace{0pt}86.0 & \nobreak\hspace{0pt}0.979 & \nobreak\hspace{0pt}1.000 & \nobreak\hspace{0pt}2.1 & \nobreak\hspace{0pt}0.0 \\
 & \nobreak\hspace{0pt}0.10 & \nobreak\hspace{0pt}0.0977 [0.0968, 0.0985] & \nobreak\hspace{0pt}88.0 & \nobreak\hspace{0pt}0.978 & \nobreak\hspace{0pt}1.000 & \nobreak\hspace{0pt}2.2 & \nobreak\hspace{0pt}0.0 \\
 & \nobreak\hspace{0pt}0.20 & \nobreak\hspace{0pt}0.0977 [0.0968, 0.0985] & \nobreak\hspace{0pt}89.1 & \nobreak\hspace{0pt}0.978 & \nobreak\hspace{0pt}1.000 & \nobreak\hspace{0pt}2.2 & \nobreak\hspace{0pt}0.0 \\
\nobreak\hspace{0pt}ConvNeXt-V2-Tiny & \nobreak\hspace{0pt}0.05 & \nobreak\hspace{0pt}0.0984 [0.0976, 0.0991] & \nobreak\hspace{0pt}86.9 & \nobreak\hspace{0pt}0.985 & \nobreak\hspace{0pt}0.998 & \nobreak\hspace{0pt}1.5 & \nobreak\hspace{0pt}2.2 \\
 & \nobreak\hspace{0pt}0.10 & \nobreak\hspace{0pt}0.0982 [0.0972, 0.0990] & \nobreak\hspace{0pt}88.3 & \nobreak\hspace{0pt}0.984 & \nobreak\hspace{0pt}0.998 & \nobreak\hspace{0pt}1.6 & \nobreak\hspace{0pt}2.2 \\
 & \nobreak\hspace{0pt}0.20 & \nobreak\hspace{0pt}0.0976 [0.0961, 0.0987] & \nobreak\hspace{0pt}89.0 & \nobreak\hspace{0pt}0.982 & \nobreak\hspace{0pt}0.998 & \nobreak\hspace{0pt}1.8 & \nobreak\hspace{0pt}2.2 \\
\nobreak\hspace{0pt}Swin-V2-S & \nobreak\hspace{0pt}0.05 & \nobreak\hspace{0pt}0.0957 [0.0945, 0.0967] & \nobreak\hspace{0pt}81.8 & \nobreak\hspace{0pt}0.957 & \nobreak\hspace{0pt}1.000 & \nobreak\hspace{0pt}4.3 & \nobreak\hspace{0pt}0.0 \\
 & \nobreak\hspace{0pt}0.10 & \nobreak\hspace{0pt}0.0956 [0.0944, 0.0966] & \nobreak\hspace{0pt}86.0 & \nobreak\hspace{0pt}0.956 & \nobreak\hspace{0pt}1.000 & \nobreak\hspace{0pt}4.4 & \nobreak\hspace{0pt}0.0 \\
 & \nobreak\hspace{0pt}0.20 & \nobreak\hspace{0pt}0.0951 [0.0938, 0.0962] & \nobreak\hspace{0pt}88.0 & \nobreak\hspace{0pt}0.951 & \nobreak\hspace{0pt}1.000 & \nobreak\hspace{0pt}4.9 & \nobreak\hspace{0pt}0.0 \\
\nobreak\hspace{0pt}ViT-L/16 & \nobreak\hspace{0pt}0.05 & \nobreak\hspace{0pt}0.0989 [0.0982, 0.0995] & \nobreak\hspace{0pt}87.9 & \nobreak\hspace{0pt}0.993 & \nobreak\hspace{0pt}0.993 & \nobreak\hspace{0pt}0.7 & \nobreak\hspace{0pt}6.7 \\
 & \nobreak\hspace{0pt}0.10 & \nobreak\hspace{0pt}0.0987 [0.0978, 0.0994] & \nobreak\hspace{0pt}88.9 & \nobreak\hspace{0pt}0.993 & \nobreak\hspace{0pt}0.995 & \nobreak\hspace{0pt}0.7 & \nobreak\hspace{0pt}4.5 \\
 & \nobreak\hspace{0pt}0.20 & \nobreak\hspace{0pt}0.0981 [0.0963, 0.0994] & \nobreak\hspace{0pt}89.2 & \nobreak\hspace{0pt}0.993 & \nobreak\hspace{0pt}0.995 & \nobreak\hspace{0pt}0.7 & \nobreak\hspace{0pt}4.5 \\
\nobreak\hspace{0pt}DINOv2-L & \nobreak\hspace{0pt}0.05 & \nobreak\hspace{0pt}0.0986 [0.0979, 0.0992] & \nobreak\hspace{0pt}87.3 & \nobreak\hspace{0pt}0.988 & \nobreak\hspace{0pt}0.995 & \nobreak\hspace{0pt}1.2 & \nobreak\hspace{0pt}4.5 \\
 & \nobreak\hspace{0pt}0.10 & \nobreak\hspace{0pt}0.0982 [0.0973, 0.0991] & \nobreak\hspace{0pt}88.4 & \nobreak\hspace{0pt}0.988 & \nobreak\hspace{0pt}0.995 & \nobreak\hspace{0pt}1.2 & \nobreak\hspace{0pt}4.5 \\
\bottomrule
\end{tabular}
\end{table}

\begin{table}[!htbp]
\centering
\small
\setlength{\tabcolsep}{3.0pt}
\renewcommand{\arraystretch}{1.12}
\caption{Per-class diagnostic odds ratios on the test set (Haldane--Anscombe corrected when a cell is zero). Confidence intervals are shown in Figure~\ref{fig:20}; the very large values reflect near-perfect separation.}
\label{tab:21}
\begin{tabular}{>{\raggedright\arraybackslash}p{0.336\linewidth}>{\raggedright\arraybackslash}p{0.130\linewidth}>{\raggedright\arraybackslash}p{0.168\linewidth}>{\raggedright\arraybackslash}p{0.125\linewidth}>{\raggedright\arraybackslash}p{0.173\linewidth}}
\toprule
\nobreak\hspace{0pt}\textbf{Model} & \nobreak\hspace{0pt}\textbf{Glioma} & \nobreak\hspace{0pt}\textbf{Meningioma} & \nobreak\hspace{0pt}\textbf{No tumor} & \nobreak\hspace{0pt}\textbf{Pituitary} \\
\midrule
\nobreak\hspace{0pt}EfficientNetV2-M & \nobreak\hspace{0pt}12,761 & \nobreak\hspace{0pt}1,856 & \nobreak\hspace{0pt}40,135 & \nobreak\hspace{0pt}33,915 \\
\nobreak\hspace{0pt}ConvNeXt-V2-Tiny & \nobreak\hspace{0pt}1,849 & \nobreak\hspace{0pt}1,710 & \nobreak\hspace{0pt}26,201 & \nobreak\hspace{0pt}26,334 \\
\nobreak\hspace{0pt}Swin-V2-S & \nobreak\hspace{0pt}1,881 & \nobreak\hspace{0pt}1,113 & \nobreak\hspace{0pt}18,248 & \nobreak\hspace{0pt}19,701 \\
\nobreak\hspace{0pt}ViT-L/16 & \nobreak\hspace{0pt}3,145 & \nobreak\hspace{0pt}979 & \nobreak\hspace{0pt}17,512 & \nobreak\hspace{0pt}65,544 \\
\nobreak\hspace{0pt}DINOv2-L & \nobreak\hspace{0pt}922 & \nobreak\hspace{0pt}2,471 & \nobreak\hspace{0pt}18,170 & \nobreak\hspace{0pt}14,651 \\
\nobreak\hspace{0pt}MedSigLIP-448 & \nobreak\hspace{0pt}11,452 & \nobreak\hspace{0pt}1,938 & \nobreak\hspace{0pt}40,135 & \nobreak\hspace{0pt}212,977 \\
\bottomrule
\end{tabular}
\end{table}

\begin{table}[!htbp]
\centering
\scriptsize
\setlength{\tabcolsep}{3.0pt}
\renewcommand{\arraystretch}{1.12}
\caption{Area under the precision--recall curve (average precision) per class, and accuracy under selective prediction at decreasing coverage (the least-confident images are referred).}
\label{tab:22}
\begin{tabular}{>{\raggedright\arraybackslash}p{0.173\linewidth}>{\raggedright\arraybackslash}p{0.076\linewidth}>{\raggedright\arraybackslash}p{0.100\linewidth}>{\raggedright\arraybackslash}p{0.076\linewidth}>{\raggedright\arraybackslash}p{0.091\linewidth}>{\raggedright\arraybackslash}p{0.076\linewidth}>{\raggedright\arraybackslash}p{0.086\linewidth}>{\raggedright\arraybackslash}p{0.064\linewidth}>{\raggedright\arraybackslash}p{0.064\linewidth}>{\raggedright\arraybackslash}p{0.064\linewidth}}
\toprule
\nobreak\hspace{0pt}\textbf{Model} & \nobreak\hspace{0pt}\textbf{AUPRC glioma} & \nobreak\hspace{0pt}\textbf{AUPRC meningioma} & \nobreak\hspace{0pt}\textbf{AUPRC no tumor} & \nobreak\hspace{0pt}\textbf{AUPRC pituitary} & \nobreak\hspace{0pt}\textbf{Macro AUPRC} & \nobreak\hspace{0pt}\textbf{Accuracy at 80\% coverage} & \nobreak\hspace{0pt}\textbf{at 90\%} & \nobreak\hspace{0pt}\textbf{at 95\%} & \nobreak\hspace{0pt}\textbf{at 100\%} \\
\midrule
\nobreak\hspace{0pt}EfficientNetV2-M & \nobreak\hspace{0pt}0.930 & \nobreak\hspace{0pt}0.886 & \nobreak\hspace{0pt}0.992 & \nobreak\hspace{0pt}0.996 & \nobreak\hspace{0pt}0.951 & \nobreak\hspace{0pt}0.9734 & \nobreak\hspace{0pt}0.9701 & \nobreak\hspace{0pt}0.9671 & \nobreak\hspace{0pt}0.9556 \\
\nobreak\hspace{0pt}ConvNeXt-V2-Tiny & \nobreak\hspace{0pt}0.958 & \nobreak\hspace{0pt}0.912 & \nobreak\hspace{0pt}0.995 & \nobreak\hspace{0pt}0.985 & \nobreak\hspace{0pt}0.963 & \nobreak\hspace{0pt}0.9828 & \nobreak\hspace{0pt}0.9799 & \nobreak\hspace{0pt}0.9776 & \nobreak\hspace{0pt}0.9500 \\
\nobreak\hspace{0pt}Swin-V2-S & \nobreak\hspace{0pt}0.943 & \nobreak\hspace{0pt}0.986 & \nobreak\hspace{0pt}0.956 & \nobreak\hspace{0pt}0.998 & \nobreak\hspace{0pt}0.971 & \nobreak\hspace{0pt}0.9859 & \nobreak\hspace{0pt}0.9764 & \nobreak\hspace{0pt}0.9664 & \nobreak\hspace{0pt}0.9450 \\
\nobreak\hspace{0pt}ViT-L/16 & \nobreak\hspace{0pt}0.947 & \nobreak\hspace{0pt}0.896 & \nobreak\hspace{0pt}0.994 & \nobreak\hspace{0pt}0.999 & \nobreak\hspace{0pt}0.959 & \nobreak\hspace{0pt}0.9820 & \nobreak\hspace{0pt}0.9806 & \nobreak\hspace{0pt}0.9757 & \nobreak\hspace{0pt}0.9525 \\
\nobreak\hspace{0pt}DINOv2-L & \nobreak\hspace{0pt}0.931 & \nobreak\hspace{0pt}0.883 & \nobreak\hspace{0pt}0.999 & \nobreak\hspace{0pt}0.999 & \nobreak\hspace{0pt}0.953 & \nobreak\hspace{0pt}0.9789 & \nobreak\hspace{0pt}0.9750 & \nobreak\hspace{0pt}0.9697 & \nobreak\hspace{0pt}0.9475 \\
\nobreak\hspace{0pt}MedSigLIP-448 & \nobreak\hspace{0pt}0.951 & \nobreak\hspace{0pt}0.975 & \nobreak\hspace{0pt}0.993 & \nobreak\hspace{0pt}1.000 & \nobreak\hspace{0pt}0.980 & \nobreak\hspace{0pt}0.9914 & \nobreak\hspace{0pt}0.9854 & \nobreak\hspace{0pt}0.9796 & \nobreak\hspace{0pt}0.9538 \\
\nobreak\hspace{0pt}Weighted-average ensemble & \nobreak\hspace{0pt}-- & \nobreak\hspace{0pt}-- & \nobreak\hspace{0pt}-- & \nobreak\hspace{0pt}-- & \nobreak\hspace{0pt}-- & \nobreak\hspace{0pt}0.9977 & \nobreak\hspace{0pt}0.9882 & \nobreak\hspace{0pt}0.9763 & \nobreak\hspace{0pt}0.9550 \\
\bottomrule
\end{tabular}
\end{table}

\subsection{External Validation on Duplicate-Filtered BDNeuro-MRI}

\textbf{Duplicate screening:} BDNeuro-MRI was not independent of the training data. Of its 5,941 images, 4,456 (75.0\%) were hash candidates of Kaggle images, and 4,266 (71.8\%) were pixel-verified near-duplicates. No image was repeated within the dataset. The primary external set therefore comprises 1,675 images (862 glioma, 306 meningioma, 222 no tumor and 285 pituitary tumor), and the strict set, which also excludes hash candidates, comprises 1,485 images.

\begin{figure}[!htbp]
\centering
\includegraphics[width=0.88\linewidth,height=0.82\textheight,keepaspectratio]{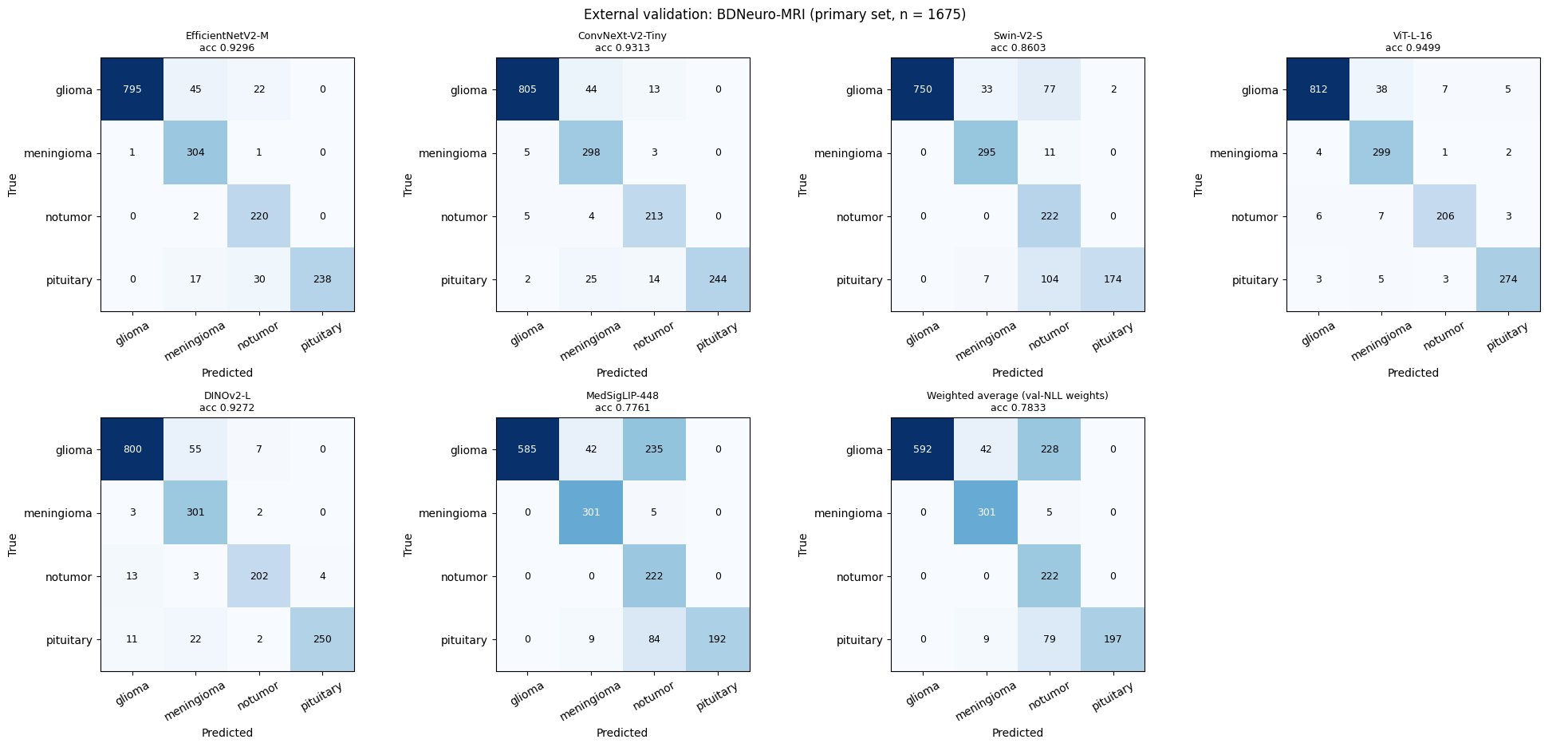}
\caption{Confusion matrices on the 1,675 non-duplicate BDNeuro-MRI images for the six primary models and the validation-weighted ensemble.}
\label{fig:24}
\end{figure}

\textbf{Performance: }On the primary external set the architectures differed more widely (Table~\ref{tab:23}, Figure~\ref{fig:24}). ViT-L/16 was the most accurate model (94.99\% [93.97, 96.06]; $-$0.26 percentage points relative to its internal accuracy), followed by ConvNeXt-V2-Tiny (93.13\%), EfficientNetV2-M (92.96\%) and DINOv2-L (92.72\%). Swin-V2-S fell to 86.03\% ($-$8.47 points). MedSigLIP-448, the best model on the validation set, fell to 77.61\% ($-$17.76 points). Twelve of the 15 pairwise differences were significant after Holm correction, against none on the internal test set.

\textbf{MedSigLIP's errors:} MedSigLIP-448's errors were concentrated in two classes (Table~\ref{tab:24}): glioma sensitivity was 0.679 and pituitary sensitivity 0.674, and its no-tumor specificity fell to 0.777. Its calibration error rose to 0.205, so its predictions were wrong with high confidence.

\textbf{Ensembles:} The validation-fitted weighted ensemble, which weighted MedSigLIP-448 at 0.71, collapsed with it (78.33\%). The simple average and majority vote, which fit no parameters, matched the best model (94.99\%) and had the highest balanced accuracy (95.93--95.99\%).

\textbf{Strict set:} All conclusions held on the strict set, with accuracies 0.3--0.9 percentage points lower.

\begin{table}[!htbp]
\centering
\scriptsize
\setlength{\tabcolsep}{3.0pt}
\renewcommand{\arraystretch}{1.12}
\caption{External validation on the 1,675 BDNeuro-MRI images that are not near-duplicates of the Kaggle data (primary set), and accuracy on the strict set. All parameters were fitted on the internal validation set.}
\label{tab:23}
\begin{tabular}{>{\raggedright\arraybackslash}p{0.167\linewidth}>{\raggedright\arraybackslash}p{0.087\linewidth}>{\raggedright\arraybackslash}p{0.087\linewidth}>{\raggedright\arraybackslash}p{0.092\linewidth}>{\raggedright\arraybackslash}p{0.081\linewidth}>{\raggedright\arraybackslash}p{0.062\linewidth}>{\raggedright\arraybackslash}p{0.062\linewidth}>{\raggedright\arraybackslash}p{0.085\linewidth}>{\raggedright\arraybackslash}p{0.066\linewidth}>{\raggedright\arraybackslash}p{0.080\linewidth}}
\toprule
\nobreak\hspace{0pt}\textbf{Model / ensemble} & \nobreak\hspace{0pt}\textbf{Accuracy [95\% CI]} & \nobreak\hspace{0pt}\textbf{Balanced accuracy [95\% CI]} & \nobreak\hspace{0pt}\textbf{Macro-F1 [95\% CI]} & \nobreak\hspace{0pt}\textbf{Macro AUC [95\% CI]} & \nobreak\hspace{0pt}\textbf{MCC} & \nobreak\hspace{0pt}\textbf{ECE} & \nobreak\hspace{0pt}\textbf{Internal test accuracy} & \nobreak\hspace{0pt}\textbf{Change (pp)} & \nobreak\hspace{0pt}\textbf{Strict set accuracy (n = 1,485)} \\
\midrule
\nobreak\hspace{0pt}EfficientNetV2-M & \nobreak\hspace{0pt}0.9296 [0.9164, 0.9415] & \nobreak\hspace{0pt}0.9355 [0.9227, 0.9471] & \nobreak\hspace{0pt}0.9150 [0.8994, 0.9295] & \nobreak\hspace{0pt}0.9865 [0.9822, 0.9904] & \nobreak\hspace{0pt}0.8980 & \nobreak\hspace{0pt}0.0545 & \nobreak\hspace{0pt}0.9556 & \nobreak\hspace{0pt}-2.61 & \nobreak\hspace{0pt}0.9239 \\
\nobreak\hspace{0pt}ConvNeXt-V2-Tiny & \nobreak\hspace{0pt}0.9313 [0.9194, 0.9439] & \nobreak\hspace{0pt}0.9308 [0.9173, 0.9444] & \nobreak\hspace{0pt}0.9195 [0.9044, 0.9340] & \nobreak\hspace{0pt}0.9917 [0.9886, 0.9946] & \nobreak\hspace{0pt}0.8988 & \nobreak\hspace{0pt}0.0472 & \nobreak\hspace{0pt}0.9500 & \nobreak\hspace{0pt}-1.87 & \nobreak\hspace{0pt}0.9259 \\
\nobreak\hspace{0pt}Swin-V2-S & \nobreak\hspace{0pt}0.8603 [0.8430, 0.8764] & \nobreak\hspace{0pt}0.8612 [0.8440, 0.8776] & \nobreak\hspace{0pt}0.8260 [0.8044, 0.8461] & \nobreak\hspace{0pt}0.9867 [0.9831, 0.9902] & \nobreak\hspace{0pt}0.8086 & \nobreak\hspace{0pt}0.1001 & \nobreak\hspace{0pt}0.9450 & \nobreak\hspace{0pt}-8.47 & \nobreak\hspace{0pt}0.8572 \\
\nobreak\hspace{0pt}ViT-L/16 & \nobreak\hspace{0pt}0.9499 [0.9397, 0.9606] & \nobreak\hspace{0pt}0.9521 [0.9406, 0.9633] & \nobreak\hspace{0pt}0.9443 [0.9325, 0.9560] & \nobreak\hspace{0pt}0.9917 [0.9887, 0.9945] & \nobreak\hspace{0pt}0.9251 & \nobreak\hspace{0pt}0.0306 & \nobreak\hspace{0pt}0.9525 & \nobreak\hspace{0pt}-0.26 & \nobreak\hspace{0pt}0.9461 \\
\nobreak\hspace{0pt}DINOv2-L & \nobreak\hspace{0pt}0.9272 [0.9146, 0.9397] & \nobreak\hspace{0pt}0.9247 [0.9098, 0.9387] & \nobreak\hspace{0pt}0.9200 [0.9057, 0.9335] & \nobreak\hspace{0pt}0.9878 [0.9841, 0.9913] & \nobreak\hspace{0pt}0.8918 & \nobreak\hspace{0pt}0.0556 & \nobreak\hspace{0pt}0.9475 & \nobreak\hspace{0pt}-2.03 & \nobreak\hspace{0pt}0.9239 \\
\nobreak\hspace{0pt}MedSigLIP-448 & \nobreak\hspace{0pt}0.7761 [0.7564, 0.7952] & \nobreak\hspace{0pt}0.8340 [0.8175, 0.8497] & \nobreak\hspace{0pt}0.7767 [0.7563, 0.7960] & \nobreak\hspace{0pt}0.9809 [0.9766, 0.9850] & \nobreak\hspace{0pt}0.7239 & \nobreak\hspace{0pt}0.2054 & \nobreak\hspace{0pt}0.9538 & \nobreak\hspace{0pt}-17.76 & \nobreak\hspace{0pt}0.7690 \\
\nobreak\hspace{0pt}Simple average & \nobreak\hspace{0pt}0.9499 [0.9391, 0.9600] & \nobreak\hspace{0pt}0.9593 [0.9496, 0.9689] & \nobreak\hspace{0pt}0.9423 [0.9292, 0.9544] & \nobreak\hspace{0pt}0.9946 [0.9923, 0.9966] & \nobreak\hspace{0pt}0.9268 & \nobreak\hspace{0pt}0.0426 & \nobreak\hspace{0pt}0.9544 & \nobreak\hspace{0pt}-0.45 & \nobreak\hspace{0pt}0.9448 \\
\nobreak\hspace{0pt}Weighted average (val-NLL weights) & \nobreak\hspace{0pt}0.7833 [0.7636, 0.8024] & \nobreak\hspace{0pt}0.8404 [0.8241, 0.8557] & \nobreak\hspace{0pt}0.7835 [0.7629, 0.8025] & \nobreak\hspace{0pt}0.9940 [0.9919, 0.9960] & \nobreak\hspace{0pt}0.7313 & \nobreak\hspace{0pt}0.1452 & \nobreak\hspace{0pt}0.9550 & \nobreak\hspace{0pt}-17.17 & \nobreak\hspace{0pt}0.7744 \\
\nobreak\hspace{0pt}Majority vote & \nobreak\hspace{0pt}0.9499 [0.9391, 0.9600] & \nobreak\hspace{0pt}0.9599 [0.9503, 0.9695] & \nobreak\hspace{0pt}0.9426 [0.9296, 0.9549] & \nobreak\hspace{0pt}0.9946 [0.9923, 0.9966] & \nobreak\hspace{0pt}0.9268 & \nobreak\hspace{0pt}0.0498 & \nobreak\hspace{0pt}0.9544 & \nobreak\hspace{0pt}-0.45 & \nobreak\hspace{0pt}0.9448 \\
\nobreak\hspace{0pt}Stacking (LR on validation) & \nobreak\hspace{0pt}0.9254 [0.9122, 0.9379] & \nobreak\hspace{0pt}0.9313 [0.9182, 0.9437] & \nobreak\hspace{0pt}0.9115 [0.8954, 0.9265] & \nobreak\hspace{0pt}0.9869 [0.9818, 0.9913] & \nobreak\hspace{0pt}0.8926 & \nobreak\hspace{0pt}0.0482 & \nobreak\hspace{0pt}0.9575 & \nobreak\hspace{0pt}-3.21 & \nobreak\hspace{0pt}0.9212 \\
\bottomrule
\end{tabular}
\end{table}

\begin{table}[!htbp]
\centering
\small
\setlength{\tabcolsep}{3.0pt}
\renewcommand{\arraystretch}{1.12}
\caption{External per-class sensitivity / specificity on the primary set.}
\label{tab:24}
\begin{tabular}{>{\raggedright\arraybackslash}p{0.339\linewidth}>{\raggedright\arraybackslash}p{0.132\linewidth}>{\raggedright\arraybackslash}p{0.178\linewidth}>{\raggedright\arraybackslash}p{0.115\linewidth}>{\raggedright\arraybackslash}p{0.167\linewidth}}
\toprule
\nobreak\hspace{0pt}\textbf{Model / ensemble} & \nobreak\hspace{0pt}\textbf{Glioma (n = 862)} & \nobreak\hspace{0pt}\textbf{Meningioma (n = 306)} & \nobreak\hspace{0pt}\textbf{No tumor (n = 222)} & \nobreak\hspace{0pt}\textbf{Pituitary (n = 285)} \\
\midrule
\nobreak\hspace{0pt}EfficientNetV2-M & \nobreak\hspace{0pt}0.922 / 0.999 & \nobreak\hspace{0pt}0.994 / 0.953 & \nobreak\hspace{0pt}0.991 / 0.964 & \nobreak\hspace{0pt}0.835 / 1.000 \\
\nobreak\hspace{0pt}ConvNeXt-V2-Tiny & \nobreak\hspace{0pt}0.934 / 0.985 & \nobreak\hspace{0pt}0.974 / 0.947 & \nobreak\hspace{0pt}0.960 / 0.979 & \nobreak\hspace{0pt}0.856 / 1.000 \\
\nobreak\hspace{0pt}Swin-V2-S & \nobreak\hspace{0pt}0.870 / 1.000 & \nobreak\hspace{0pt}0.964 / 0.971 & \nobreak\hspace{0pt}1.000 / 0.868 & \nobreak\hspace{0pt}0.611 / 0.999 \\
\nobreak\hspace{0pt}ViT-L/16 & \nobreak\hspace{0pt}0.942 / 0.984 & \nobreak\hspace{0pt}0.977 / 0.964 & \nobreak\hspace{0pt}0.928 / 0.992 & \nobreak\hspace{0pt}0.961 / 0.993 \\
\nobreak\hspace{0pt}DINOv2-L & \nobreak\hspace{0pt}0.928 / 0.967 & \nobreak\hspace{0pt}0.984 / 0.942 & \nobreak\hspace{0pt}0.910 / 0.992 & \nobreak\hspace{0pt}0.877 / 0.997 \\
\nobreak\hspace{0pt}MedSigLIP-448 & \nobreak\hspace{0pt}0.679 / 1.000 & \nobreak\hspace{0pt}0.984 / 0.963 & \nobreak\hspace{0pt}1.000 / 0.777 & \nobreak\hspace{0pt}0.674 / 1.000 \\
\nobreak\hspace{0pt}Simple average & \nobreak\hspace{0pt}0.935 / 1.000 & \nobreak\hspace{0pt}0.994 / 0.963 & \nobreak\hspace{0pt}1.000 / 0.977 & \nobreak\hspace{0pt}0.909 / 1.000 \\
\nobreak\hspace{0pt}Weighted average (val-NLL weights) & \nobreak\hspace{0pt}0.687 / 1.000 & \nobreak\hspace{0pt}0.984 / 0.963 & \nobreak\hspace{0pt}1.000 / 0.785 & \nobreak\hspace{0pt}0.691 / 1.000 \\
\nobreak\hspace{0pt}Majority vote & \nobreak\hspace{0pt}0.934 / 1.000 & \nobreak\hspace{0pt}0.994 / 0.963 & \nobreak\hspace{0pt}1.000 / 0.977 & \nobreak\hspace{0pt}0.912 / 1.000 \\
\nobreak\hspace{0pt}Stacking (LR on validation) & \nobreak\hspace{0pt}0.919 / 1.000 & \nobreak\hspace{0pt}0.997 / 0.939 & \nobreak\hspace{0pt}0.996 / 0.972 & \nobreak\hspace{0pt}0.814 / 1.000 \\
\bottomrule
\end{tabular}
\end{table}

\subsection{Computational Cost}

Parameter counts ranged from 27.9 M (ConvNeXt-V2-Tiny) to 428.6 M (MedSigLIP-448), and median single-image GPU latency from 12 ms (ConvNeXt-V2-Tiny) to 40 ms (Swin-V2-S) (Table~\ref{tab:25}). Training MedSigLIP-448 took 126 s per epoch, against 11 s for ConvNeXt-V2-Tiny. Batch throughput measurements were not reproducible across identical architectures, because no warm-up was used for batched inference, and are not reported.

\begin{table}[!htbp]
\centering
\footnotesize
\setlength{\tabcolsep}{3.0pt}
\renewcommand{\arraystretch}{1.12}
\caption{Computational cost on an NVIDIA A100-SXM4-80GB GPU with bfloat16 inference.}
\label{tab:25}
\begin{tabular}{>{\raggedright\arraybackslash}p{0.349\linewidth}>{\raggedright\arraybackslash}p{0.144\linewidth}>{\raggedright\arraybackslash}p{0.077\linewidth}>{\raggedright\arraybackslash}p{0.124\linewidth}>{\raggedright\arraybackslash}p{0.118\linewidth}>{\raggedright\arraybackslash}p{0.108\linewidth}}
\toprule
\nobreak\hspace{0pt}\textbf{Model} & \nobreak\hspace{0pt}\textbf{Parameters (M)} & \nobreak\hspace{0pt}\textbf{Input (px)} & \nobreak\hspace{0pt}\textbf{Inference latency, batch 1 (ms, median)} & \nobreak\hspace{0pt}\textbf{Training time per epoch (s)} & \nobreak\hspace{0pt}\textbf{Peak training GPU memory (GB)} \\
\midrule
\nobreak\hspace{0pt}EfficientNetV2-M & \nobreak\hspace{0pt}52.86 & \nobreak\hspace{0pt}224 & \nobreak\hspace{0pt}37.4 & \nobreak\hspace{0pt}24.6 & \nobreak\hspace{0pt}5.1 \\
\nobreak\hspace{0pt}ConvNeXt-V2-Tiny & \nobreak\hspace{0pt}27.87 & \nobreak\hspace{0pt}224 & \nobreak\hspace{0pt}12.0 & \nobreak\hspace{0pt}10.9 & \nobreak\hspace{0pt}4.2 \\
\nobreak\hspace{0pt}Swin-V2-S & \nobreak\hspace{0pt}48.96 & \nobreak\hspace{0pt}256 & \nobreak\hspace{0pt}39.8 & \nobreak\hspace{0pt}23.9 & \nobreak\hspace{0pt}6.9 \\
\nobreak\hspace{0pt}ConvNeXt-V2-Tiny (448 px) & \nobreak\hspace{0pt}27.87 & \nobreak\hspace{0pt}448 & \nobreak\hspace{0pt}12.3 & \nobreak\hspace{0pt}29.8 & \nobreak\hspace{0pt}15.1 \\
\nobreak\hspace{0pt}ConvNeXt-V2-Tiny (448 px, matched) & \nobreak\hspace{0pt}27.87 & \nobreak\hspace{0pt}448 & \nobreak\hspace{0pt}12.2 & \nobreak\hspace{0pt}38.0 & \nobreak\hspace{0pt}5.2 \\
\nobreak\hspace{0pt}ViT-L/16 & \nobreak\hspace{0pt}303.31 & \nobreak\hspace{0pt}224 & \nobreak\hspace{0pt}16.8 & \nobreak\hspace{0pt}16.0 & \nobreak\hspace{0pt}11.3 \\
\nobreak\hspace{0pt}DINOv2-L & \nobreak\hspace{0pt}304.38 & \nobreak\hspace{0pt}224 & \nobreak\hspace{0pt}18.4 & \nobreak\hspace{0pt}22.6 & \nobreak\hspace{0pt}13.9 \\
\nobreak\hspace{0pt}MedSigLIP (224 px) & \nobreak\hspace{0pt}428.57 & \nobreak\hspace{0pt}224 & \nobreak\hspace{0pt}20.8 & \nobreak\hspace{0pt}59.9 & \nobreak\hspace{0pt}10.0 \\
\nobreak\hspace{0pt}MedSigLIP-448 & \nobreak\hspace{0pt}428.57 & \nobreak\hspace{0pt}448 & \nobreak\hspace{0pt}21.1 & \nobreak\hspace{0pt}126.2 & \nobreak\hspace{0pt}18.1 \\
\bottomrule
\end{tabular}
\end{table}

\FloatBarrier

\section{Discussion}

This study benchmarked six architectures from three families for explainable brain tumor MRI classification and evaluated them for accuracy, calibration, explainability, ensembles and decision-analytic performance. The main finding is that the architectures could not be distinguished by their internal test accuracy, whereas they differed on duplicate-filtered external data. Internally, accuracies spanned one percentage point and no pairwise difference survived multiple-comparison correction. Externally, accuracy spanned 17 percentage points and 12 of 15 differences were significant. The model that would have been selected on validation accuracy, MedSigLIP-448, generalised worst, while ViT-L/16 lost almost no accuracy. Internal comparisons on this collection may therefore be a poor guide to how the architectures behave on new data.

The data audit explains why the internal comparison is uninformative. More than a third of the test images have a verified near-duplicate in the training folder, almost all Figshare test patients also contribute training images, and class membership is confounded with source. All three properties allow memorisation of acquisition- and source-specific appearance to pass as diagnostic performance~\cite{r42,r43,r44}. Accuracy fell by about 4 percentage points on the clean test subset. Background-only images retained most of the class information for frozen features. Source membership could be predicted from the learned representations. We therefore interpret the 94.5--95.6\% internal accuracies as upper bounds that partly reflect this leakage, not as estimates of performance on new patients. Accuracies above 97\% reported on the same collection~\cite{r35,r36,r37} should be read in this light, particularly when duplicates and patients are not controlled.

The external evaluation also shows that a separately published dataset is not automatically independent. 72\% of BDNeuro-MRI duplicates the Kaggle composite, although the dataset is described as originating from a single hospital. Had the screening been omitted, the external accuracies would have been dominated by images that the models had effectively seen. Duplicate screening against the training data should therefore precede any external evaluation on public MRI collections. More broadly, the provenance of public medical-imaging datasets deserves the same scrutiny as their labels.

The ablations temper the expectation that medical-domain pretraining confers an advantage. MedSigLIP's internal performance did not depend on its resolution. It was matched by a 28-million-parameter CNN trained with its own settings, and it was the least stable model under both input perturbation and data shift. One plausible explanation is that its representations encode more source-specific information, which helps on in-distribution data and hurts under shift. Its embeddings gave the highest source-prediction accuracy (86.4\%), and its explanation maps placed a third of their energy on the image border. This interpretation is supported by these consistent signals but not proven by them. ViT-L/16 and the convolutional models degraded least. The practical implication is that architecture choice should rest on robustness under realistic shift, not on validation accuracy from a single source mix.

Ensembles offered little. The models make largely the same errors (oracle accuracy 96.75\%). Validation-fitted weighting transferred the failure of MedSigLIP-448 to the ensemble, while unweighted averaging held its accuracy externally. Stacking gave no gain once its meta-learner was fitted on validation data. The cross-fitted analysis suggests that it reflected adaptation to the test distribution rather than complementary information.

The quantitative explainability analysis gives a sober picture. Attribution maps rarely peaked on the tumor, placed only a few times the chance level of energy within it, and for transformer-based models concentrated substantially on the image border. Maps also differed widely between methods applied to the same model. The maps did depend on the learned weights, so they are not trivially uninformative~\cite{r64}. But they do not support the view that foundation models produce more anatomically coherent attention, and their qualitative appeal should not be mistaken for evidence of pathology-based reasoning.

The decision-analytic results should be read narrowly. Tumor versus no-tumor separation is nearly perfect on this collection, which is partly a consequence of the no-tumor class coming from a single source, so the favourable decision curves are an upper bound. They describe one defined decision under re-weighted prevalence, not a demonstrated improvement in patient care. Establishing clinical utility would require a prospective study with consecutive patients, volumetric examinations, comparison with neuroradiologists and outcome or workflow end points.

\subsection{Limitations}

This study has several limitations. The external evaluation rests on the 1,675 non-duplicate images of a single dataset whose provenance cannot be fully verified, given its overlap with public data, so it is a partially independent test rather than a multicentre validation. Patient identifiers exist only for the Figshare images; for the SARTAJ, Br35H and unmatched images, duplicate screening reduces but cannot remove the chance of correlated images, such as adjacent slices, in different partitions. The predefined test folder was kept for comparability although it overlaps the training data, and the clean subset that reduces this overlap is small for some sources. All analyses use two-dimensional slices of mixed planes and contrasts rather than complete volumetric examinations, and no demographic, scanner or acquisition metadata were available.

Expert tumor masks exist only for Figshare tumor images, only 6 of which belong to the clean subset, so the localisation analysis could not be stratified by leakage status, and the blinded rating of attribution maps by neuroradiologists has been prepared but not yet carried out. The validation partition, drawn from the Training folder, is distributed differently from the test folder, which limits how well validation-fitted temperatures and thresholds transfer to the test set. Each architecture was trained with one hyperparameter configuration and one seed, and the matched-preprocessing ablation shows that such choices can change accuracy by more than one percentage point. Finally, the decision analysis is retrospective, uses illustrative prevalences and does not measure patient outcomes.

\section{Conclusion}

This benchmark of six convolutional, vision-transformer and foundation-model architectures for explainable brain tumor MRI classification found statistically indistinguishable internal accuracy (94.5--95.6\%) under a leakage-aware protocol. Validation-fitted temperature scaling reduced the expected calibration error of all six primary models, and ensembling gave no benefit. Internal accuracy fell by about 4 percentage points once duplicates and shared patients were removed: a third of the public test set is not independent of the training data, class and source are confounded, and a dataset proposed for external validation largely duplicated the training composite. On the remaining external images the architectures differed widely. The medical foundation model that performed best on validation data generalised worst, while a large vision transformer and two CNNs degraded least. Attribution maps passed a sanity check but localised tumors poorly.

The study does not identify a model ready for clinical use. Its conclusions concern evaluation practice: duplicates should be screened with pixel verification, splits should be grouped by patient and by duplicate, ensembles and thresholds should be fitted on validation data only, external data should be checked against the training data before use, and explanations should be validated quantitatively.

Future work should prioritise prospective, multicentre, patient-level validation on volumetric MRI with realistic prevalence; completion of the blinded expert rating of attribution maps; training strategies that suppress source-specific shortcuts; and evaluation of model robustness across scanners and institutions.

\section*{Acknowledgements}

The author thanks the creators and maintainers of the Figshare, SARTAJ, Br35H, Kaggle Brain Tumor MRI and BDNeuro-MRI datasets for making their data openly accessible to the research community.

\section*{Funding}

This research received no specific grant from any funding agency in the public, commercial, or not-for-profit sectors.

\section*{Ethics Statement}

This study used only publicly available, de-identified MRI datasets released by their original custodians under open-access terms: the Kaggle Brain Tumor MRI Dataset~\cite{r38}, the Figshare brain tumor dataset~\cite{r39}, SARTAJ~\cite{r40}, Br35H~\cite{r41} and BDNeuro-MRI~\cite{r92}. No new patient data were collected, no human participants were recruited and no clinical interventions were performed, so institutional ethics approval was not required. All data use complied with the terms of the original dataset licences.

\section*{Data Availability}

All datasets analysed in this study are publicly available: the Kaggle Brain Tumor MRI Dataset (\url{https://www.kaggle.com/datasets/masoudnickparvar/brain-tumor-mri-dataset}; version accessed in 2026, 7,216 images), the Figshare brain tumor dataset (\url{https://doi.org/10.6084/m9.figshare.1512427}), SARTAJ (\url{https://www.kaggle.com/datasets/sartajbhuvaji/brain-tumor-classification-mri}), Br35H (\url{https://www.kaggle.com/datasets/ahmedhamada0/brain-tumor-detection}) and BDNeuro-MRI (\url{https://doi.org/10.17632/zwr4ntf94j.7}). The exact training, validation and test assignment of every image, the duplicate pairs with their pixel correlations, the source attributions and Figshare patient identifiers, the clean-test flags, the external duplicate screening and all prediction files are available from the corresponding author on reasonable request.

\section*{Author Contributions}

A.A.: conceptualisation, methodology, software, formal analysis, data curation, investigation, visualisation, and writing (original draft, review and editing).

\section*{Competing Interests}

The author declares no competing interests.

\end{document}